\documentclass[12pt]{article}

\PassOptionsToPackage{numbers,sort&compress}{natbib}

\IfFileExists{neurips_2027.sty}{%
  \usepackage[preprint]{neurips_2027}%
}{%
  \IfFileExists{neurips_2026.sty}{%
    \usepackage[preprint]{neurips_2026}%
  }{%
    \usepackage[margin=1in]{geometry}%
    \usepackage{times}%
    \usepackage{parskip}% no paragraph indentation; space between paragraphs instead
    \usepackage{natbib}% neurips_202x.sty normally loads this; load it here too so \cite sorting/compression (see line 3) still works without the official style file%
  }%
}

\makeatletter
\ifdefined\@noticestring\renewcommand{\@noticestring}{}\fi
\makeatother

\usepackage[utf8]{inputenc}
\usepackage[T1]{fontenc}
\usepackage{amsmath,amssymb,amsthm,amsfonts}
\usepackage{upgreek}% upright \uppi for the number 3.14159; \pi is the policy
\usepackage{mathtools}
\usepackage{algorithm}
\usepackage{algorithmic}
\usepackage{booktabs}
\usepackage{array}
\newcolumntype{L}[1]{>{\raggedright\arraybackslash}p{#1}}
\usepackage{hyperref}
\usepackage{url}

\newtheoremstyle{dmplain}%
  {9.9pt plus 1pt minus 2pt}% space above
  {\topsep}% space below
  {\itshape}% body font
  {0pt}% indent
  {\bfseries}% head font
  {.}% punctuation after head
  {5pt plus 1pt minus 1pt}% space after head
  {}% head spec
\newtheoremstyle{dmdefinition}%
  {9.9pt plus 1pt minus 2pt}%
  {\topsep}%
  {\normalfont}%
  {0pt}%
  {\bfseries}%
  {.}%
  {5pt plus 1pt minus 1pt}%
  {}%
\theoremstyle{dmplain}
\newtheorem{theorem}{Theorem}[section]
\newtheorem{lemma}[theorem]{Lemma}
\newtheorem{proposition}[theorem]{Proposition}
\newtheorem{corollary}[theorem]{Corollary}
\theoremstyle{dmdefinition}
\newtheorem{definition}[theorem]{Definition}

\newtheorem{example}[theorem]{Example}
\newtheorem{remark}[theorem]{Remark}% uses the definition style, so the heading is bold like Examples

\usepackage{etoolbox}
\BeforeBeginEnvironment{proof}{\par\addvspace{4pt plus 1pt}}

\DeclareMathOperator{\Tr}{Tr}
\DeclareMathOperator{\Var}{Var}
\DeclareMathOperator{\Cov}{Cov}
\DeclareMathOperator{\re}{Re}

\DeclareMathOperator{\diag}{diag}
\DeclareMathOperator{\supp}{supp}
\newcommand{\C}{\mathbb{C}}
\newcommand{\R}{\mathbb{R}}
\newcommand{\E}{\mathbb{E}}
\newcommand{\A}{\mathcal{A}}
\newcommand{\Sset}{\mathcal{S}}
\newcommand{\eb}{\mathbf{e}}% standard basis vector e_i (bold, to distinguish from exp)

\title{Density-Matrix Reinforcement Learning}

\author{%
  Chiuman Ho\thanks{Email: \texttt{chiuman100@gmail.com}}\\[1em]
  TetraMem Inc.\thanks{This research was conducted independently by the author in a personal capacity, outside the scope of their employment and without the use of company resources. It is unrelated to the author’s work at the company.}
}

\date{}

\renewcommand{\footnoterule}{\kern 2\baselineskip\hrule width 0.2\columnwidth\kern 2.6pt}% shorter rule above footnote (default is 0.4\columnwidth)

\begin{document}

\maketitle

\begin{abstract}\noindent
What information does a reinforcement learning policy omit? At each state, a standard policy stores one probability per action. It has no place to record how actions relate to each other, such as which pairs go together and which conflict. We ask when this missing structure matters, and we give exact answers. We store the policy at each state as a density matrix, a positive semidefinite, unit-trace matrix borrowed from quantum information theory. We use it here as a purely classical data structure, computed on ordinary classical hardware. The diagonal holds the usual action probabilities. The off-diagonal entries, called coherences, are new parameters. This gives density-matrix reinforcement learning (DMRL). Coherences change outcomes in two settings. First, suppose a policy is regularized toward a coherent reference policy using matrix relative entropy. The optimum then has coherences that no diagonal reference policy produces, and its action probabilities can depend on the reference policy's coherences. This is relevant to fine-tuning of large language models (LLMs) with a Kullback--Leibler penalty. Second, suppose the agent chooses a unit-vector action and the reward is a quadratic function of that vector. This could arise when an LLM judge scores a weighted consolidation of reasoning paths, or when coupled robotic sensors each respond to a mix of actions. Then some coherent policies earn strictly more expected reward than every classical policy with the same action probabilities, and we give the margin in closed form. In a two-action example, both policies assign probability $0.5$ to each action, yet the classical one earns expected reward $0$ while the coherent one earns $1$, the maximum. Outside these two settings, DMRL matches or transfers the classical results. This is a theory paper. All results are proved, and we report no empirical evaluation.
\end{abstract}

\section{Introduction}\label{sec:intro}
At each state, a stochastic policy in reinforcement learning (RL) is a probability vector with one number per action. It records how likely each action is, but not how actions relate to one another, such as two actions that tend to occur together. When an agent selects one action at a time and its reward depends only on that action, this omission does not affect the expected reward. Two settings go beyond that case. First, many methods keep the policy close to a fixed reference policy through a Kullback--Leibler (KL) penalty. Offline RL does this \cite{levine2020offline}, as does the fine-tuning of large language models (LLMs) through reinforcement learning from human feedback (RLHF) and many forms of reinforcement learning with verifiable rewards (RLVR) \cite{ouyang2022training,shao2024deepseekmath,guo2025deepseekr1}. The reference policy is estimated from data, and a single data point may provide information about multiple actions. For example, an LLM response can be scored on several styles. Storing the reference policy as a probability vector would discard how those actions go together. Second, a reward model could, in principle, score a weighted combination of candidate responses instead of a single one. One example is several LLM reasoning paths consolidated into one answer, as in parallel thinking \cite{google2025deepthink,wang2025survey}. A probability vector cannot record how such paths work together. In both settings, the relations between actions that a probability vector does not record could matter. This paper asks when they do.

Answering this requires a policy representation that records those relations. Such a representation should meet three requirements. It should always define a valid policy, whether we sample from it, average several of them, or update it by gradient steps. It should reduce exactly to a probability vector when there is nothing extra to record. And it should have a compressed form when the true structure is simple. One natural object has all three. It is the density matrix, a positive semidefinite (PSD), unit-trace matrix borrowed from quantum information theory \cite{vonneumann1955mathematical,nielsen2010quantum}. We use it here purely as a classical data structure. It is not new, and we do not claim it is the only choice. A classical agent could store action embeddings, a covariance matrix, or a kernel instead. What makes the density matrix natural is that classical RL sits inside it. Its diagonal is the familiar policy, its off-diagonal entries record pairwise structure, and deleting them recovers classical RL exactly.

Concretely, each state gets a density matrix $\rho_s$ in place of a probability vector. For two actions this is just the $2\times2$ table $\rho=\bigl(\begin{smallmatrix}p& c\\ c&1-p\end{smallmatrix}\bigr)$. The diagonal entries $p$ and $1-p$ are ordinary action probabilities. The number $c$ is new. It records how the two actions relate, and the classical case is exactly $c=0$. When $\rho$ is real (real symmetric rather than complex Hermitian), it is the second-moment matrix $\E[\mathbf{x}\mathbf{x}^\top]$ of a distribution over unit vectors $\mathbf{x}$, and $c=\E[x_1x_2]$ is the cross-moment between the two components (Section~\ref{subsec:coupledreward}). Informally it behaves like a correlation, though it is not centered or scaled the way a statistical correlation coefficient is. Throughout the paper, \emph{classical} means this coherence-free, diagonal case, using the term in the quantum-information sense. We call the resulting framework \emph{D}ensity-\emph{M}atrix \emph{R}einforcement \emph{L}earning (DMRL). We keep names inherited from quantum information theory (coherence, observable, phase, von Neumann entropy, purity, Gibbs state, etc.) because they are the established terms for the matrix quantities we use. Every density matrix in this paper is an ordinary PSD matrix, and every computation runs on an ordinary classical computer.

We summarize below where this formalism changes outcomes.

\begin{itemize}
    \item \textbf{Coherent reference policies carry pairwise structure into training.} Many RL methods keep the policy close to a fixed reference policy through a KL penalty. Offline RL builds that reference policy from logged data, that is, interactions recorded earlier by some other policy \cite{levine2020offline}. In LLM fine-tuning, RLHF and KL-regularized RLVR, including the original formulation of group relative policy optimization (GRPO), build it from a supervised or previous-round model \cite{ouyang2022training,shao2024deepseekmath,guo2025deepseekr1}. Suppose the reference policy is a density matrix with coherences, and the penalty is matrix relative entropy. Such a reference policy can be built from data that score each recorded sample on several actions at once, for example an LLM response scored on several styles (Section~\ref{subsec:coherentprior}). Then the trained policy has coherences too (Corollary~\ref{cor:coherentprior}, Proposition~\ref{prop:coherencetransfer}). Its action probabilities can also depend on the reference policy's coherences, which a method that tracks only the reference policy's diagonal never sees (Example~\ref{ex:coherentprior}).
    \item \textbf{Coupled rewards turn coherences into higher reward.} Here the action model changes instead. In a one-step (bandit) setting, suppose the agent chooses a unit vector $\mathbf{x}$ instead of a single action label, and the reward is $\mathbf{x}^\top O\mathbf{x}$. The expected reward is then $\Tr(\rho O)$, where $\rho$ is the second-moment matrix of the chosen vectors. A classical policy is the special case that chooses only standard basis vectors. If $O$ is non-diagonal, some coherent policies earn strictly more expected reward than every classical policy with the same action probabilities (Proposition~\ref{prop:coherentdominance}). We also give the margin over the best classical policy in closed form (Proposition~\ref{prop:gap}). In our worked two-action example, the two actions are equally likely under both policies. Still, the classical policy earns expected reward $0$, whereas a coherent policy earns $1$, the maximum possible. Possible sources include an LLM judge scoring a weighted consolidation of reasoning paths or robotic sensors that each respond to a mix of actions (Section~\ref{subsec:coupledreward}, Example~\ref{ex:llmpaths}).
\end{itemize}

Outside these two settings, DMRL reduces consistently to classical RL. In a standard Markov decision process (MDP), the reward depends only on the chosen action. Actions are sampled from the diagonal of $\rho$, so the coherences affect neither the sampled actions nor the expected return. The entropy, trust-region, and estimation-time results then either match classical methods or transfer from them. These include the Gibbs optimum with its annealing law, the trust-region guarantees, and the control-variate variance reduction. Appendices~\ref{sec:entropy} and~\ref{sec:theory} develop these results. Section~\ref{sec:beyond} studies the two settings where coherences do matter. The price of the DMRL representation is extra compute, which we discuss in Section~\ref{sec:limitations}. Proofs, algorithms, and worked examples are in the appendix.

\section{Background: the density matrix without the physics}\label{sec:background}
Let $\A=\{a_1,\dots,a_k\}$ be a finite action space, and let $\Sset$ denote the state space. A classical stochastic policy is $\pi(\cdot\mid s)\in\Delta_k$ at each state $s$, where $\Delta_k$ is the probability simplex. The values $Q^\pi(s,a)$ and $V^\pi(s)$ are the action and state values \cite{sutton2018reinforcement}. Nothing in this section or in Section~\ref{sec:foundations} is specific to actions. The same facts hold for any finite index set expressed in the same basis.

We identify action $a_i$ with the $i$-th standard basis vector $\eb_i\in\C^k$, so that $\eb_i^\dagger \eb_j=\delta_{ij}$. Here $(\cdot)^\dagger$ is the conjugate transpose, which is the ordinary transpose for real vectors, and $\Tr(\cdot)$ is the trace. For a matrix $B$, the product $\eb_i^\dagger B\eb_j$ is its $(i,j)$ entry. The matrix $\eb_i\eb_i^\dagger$ has a single $1$ in position $(i,i)$ and zeros elsewhere. We use natural logarithms. Bold letters such as $\mathbf{x}$, $\eb_i$, $\mathbf{p}$ and $\mathbf{v}_n$ denote vectors, and their entries, such as $p_i$, are plain. The letters $p$ and $c$ are plain scalars in the two-action example of Appendix~\ref{sec:theory}. At each state $s$, the action values form the diagonal matrix $Q_s=\diag\bigl(Q(s,a_1),\dots,Q(s,a_k)\bigr)$, which we call the action-value operator. We drop the superscript $\pi$ from $Q^\pi$ when the policy is clear.

\begin{definition}[Density matrix]\label{def:dm}
A density matrix is an operator $\rho\in\C^{k\times k}$ with $\rho\succeq0$, $\rho=\rho^\dagger$ (i.e., $\rho$ is Hermitian), and $\Tr(\rho)=1$. We write $\mathcal D_k$ for the set of $k\times k$ density matrices.
\end{definition}

\begin{definition}[Observable]\label{def:observable}
An observable is a Hermitian operator $O\in\C^{k\times k}$, $O=O^\dagger$. For $\rho\in\mathcal D_k$, the expectation of $O$ under $\rho$ is $\E_\rho[O]=\Tr(\rho O)$.
\end{definition}

The expectation $\Tr(\rho O)=\sum_{i,j}\rho_{ij}O_{ji}$ is taken over the whole matrix, not just its diagonal. It reduces to the familiar weighted average $\sum_i\rho_{ii}O_{ii}$ for every $\rho$ exactly when $O$ is diagonal.

The off-diagonal entries $\rho_{ij}$ with $i\neq j$ are the coherences. By Hermiticity, $\rho_{ji}=\overline{\rho_{ij}}$. Write $\rho_{ij}=|\rho_{ij}|\,e^{\mathrm{i}\phi_{ij}}$, where $|\rho_{ij}|$ is the size of the coherence and $\phi_{ij}$ is its phase. If all entries of $\rho$ are real, the phase is only a sign. Section~\ref{sec:foundations} shows when coherences affect an evaluation and when they do not.
Being Hermitian, $\rho$ has a spectral decomposition $\rho=\sum_{n=1}^k\lambda_n\mathbf{v}_n\mathbf{v}_n^\dagger$, with real eigenvalues $\lambda_n\ge0$ and $\sum_n\lambda_n=1$. The eigenvectors $\{\mathbf{v}_n\}$ are orthonormal. They form the natural basis of $\rho$, which is generally different from the standard basis $\{\eb_i\}$. Write $v_{n,i}=\eb_i^\dagger \mathbf{v}_n$ for the overlap of $\mathbf{v}_n$ with the $i$-th basis vector. Because the eigenbasis is unitary, $\sum_n|v_{n,i}|^2=\sum_i|v_{n,i}|^2=1$.

\begin{definition}[Von Neumann entropy]\label{def:vne}
For $\rho\in\mathcal D_k$, the von Neumann entropy is $S(\rho)=-\Tr(\rho\log\rho)$ \cite{vonneumann1955mathematical}. It equals the Shannon entropy of the spectrum, $S(\rho)=-\sum_n\lambda_n\log\lambda_n$. If $\rho$ is diagonal with entries $p_i$, this is the Shannon entropy $-\sum_i p_i\log p_i$.
\end{definition}

The diagonal of $\rho$ is always at least as spread out as its spectrum. Hence $S(\rho)\le S(\diag(\rho))$, with equality exactly when $\rho$ is diagonal. This is the Schur--Horn relation, which we state and prove as Lemma~\ref{lem:schurhorn} in Appendix~\ref{sec:entropy}. It says that at a fixed diagonal, coherences can only lower the entropy.

\begin{definition}[Matrix relative entropy]\label{def:qre}
For $\rho,\rho'\in\mathcal D_k$ with $\supp(\rho)\subseteq\supp(\rho')$, the matrix relative entropy is given by $D(\rho\,\Vert\,\rho')=\Tr\bigl[\,\rho\,(\log\rho-\log\rho')\,\bigr]$, and $D(\rho\,\Vert\,\rho')=+\infty$ otherwise. It generalizes the classical Kullback--Leibler divergence. Its standard properties, including Klein's inequality, are stated in Appendix~\ref{sec:entropy}.
\end{definition}

Table~\ref{tab:dictionary} lists the main correspondences with classical RL. Setting all off-diagonal entries to zero turns each operator concept into its classical analogue.

\begin{table}[t]
\centering
\small
\caption{Operator concepts and their classical analogues. Setting the off-diagonal entries of $\rho$ (and of $\rho'$ and $O$, where relevant) to zero recovers the right-hand column. The diagonal is unchanged by this operation. The coherences have no classical counterpart because they vanish.}
\label{tab:dictionary}
\renewcommand{\arraystretch}{1.3}
\begin{tabular}{ll}
\toprule
Operator concept & Classical analogue (diagonal case) \\
\midrule
Density matrix $\rho$ & Probability vector  \\
Diagonal $(\rho_s)_{ii}$ & Action probabilities $\pi(a_i\mid s)$ \\
Coherence $\rho_{ij}$, $i\neq j$ & None (zero in the classical case) \\
Observable $O$; expectation $\Tr(\rho O)$ & Reward table; expected value $\sum_i \pi_i O_{ii}$ \\
Von Neumann entropy $S(\rho)$ & Shannon entropy $-\sum_i\pi_i\log\pi_i$ \\
Matrix relative entropy $D(\rho\Vert\rho')$ & KL divergence $D_{\mathrm{KL}}(\pi\Vert\pi')$ \\
\bottomrule
\end{tabular}
\end{table}

\section{Foundational principles: what coherences change and what they do not}\label{sec:foundations}

Section~\ref{sec:background} defined $\mathcal D_k$ as a mathematical object. This section presents three general facts about it. The first is how $\mathcal D_k$ contains ordinary probability vectors. The second is when the off-diagonal entries can move an expectation and when they cannot. The third is a parameterization in which every gradient step stays inside $\mathcal D_k$. Nothing here is specific to RL, and every later section builds on these facts. All proofs are in Appendix~\ref{app:proofs-foundations}.

\subsection{Probability vectors as diagonal density matrices}

\begin{proposition}[Probability vectors inside $\mathcal D_k$]\label{prop:inclusion}
The probability simplex $\Delta_k$ is in bijection with the set of density matrices that are diagonal in the basis $\{\eb_i\}$. Hence probability vectors form a strict subset of $\mathcal D_k$ for $k\ge2$. Moreover, $\mathcal D_k$ is a compact convex set of real dimension $k^2-1$, whereas $\Delta_k$ embedded as diagonal density matrices has dimension $k-1$.
\end{proposition}

So the extra $k(k-1)$ dimensions are exactly the coherences. Reading $\{a_i\}$ as actions, a diagonal $\rho_s$ is exactly an ordinary RL policy (Section~\ref{sec:policies}).

\subsection{Phase relevance of non-diagonal evaluations}\label{subsec:phase}

The next two results say exactly when coherence is invisible and when it is not.

\begin{lemma}[Phase irrelevance]\label{lem:phaseirrelevance}
If $O$ is diagonal in the basis $\{\eb_i\}$, then $\Tr(\rho O)=\sum_i O_{ii}\,\rho_{ii}$ depends only on the diagonal of $\rho$. Off-diagonal entries and their phases have no effect on this expectation.
\end{lemma}

\begin{theorem}[Phase relevance]\label{thm:phaserelevance}
Suppose $O=O^\dagger$ has an off-diagonal entry $O_{ij}\ne 0$ for some $i\ne j$. Then there exist density matrices with identical diagonals but different expectations $\Tr(\rho O)$.
\end{theorem}

So if $O$ has a nonzero off-diagonal entry, two density matrices with the same diagonal can still earn different expectations. The off-diagonal entries are exactly what moves such evaluations. Example~\ref{ex:toy} in Appendix~\ref{app:constructions} gives a worked $2\times2$ case. Section~\ref{sec:policies} uses the converse, Lemma~\ref{lem:phaseirrelevance}, to show that an ordinary MDP never sees the off-diagonal entries.

\subsection{Factor parameterization}

We also fix a parameterization of $\mathcal D_k$ that is valid at every rank.

\begin{definition}[Factor parameterization]\label{def:factorparam}
For $1\le r\le k$ and $U\in\C^{k\times r}$ nonzero, the factor parameterization is
\begin{equation}
\rho(U)=\frac{UU^\dagger}{\Tr(UU^\dagger)}.
\end{equation}
\end{definition}

\begin{lemma}[Well-definedness and rank bound]\label{lem:factorwelldef}
For every nonzero $U$, $\rho(U)\succeq0$ and $\Tr\rho(U)=1$. So $\rho(U)\in\mathcal D_k$.
Moreover, $\mathrm{rank}\,\rho(U)\le r$. Taking $r\ll k$ therefore gives a scalable,
automatically valid low-rank object.
\end{lemma}

By Lemma~\ref{lem:factorwelldef}, any nonzero $U$ gives a valid $\rho(U)$. Standard unconstrained optimizers therefore apply directly to $U$, with no projection step. \textit{DMPG} (density-matrix policy gradient; Algorithm~\ref{alg:dmpg} of Appendix~\ref{app:algorithms}) samples actions from the diagonal of $\rho_s=\rho(U_\theta(s))$. It then ascends $J(\theta)$ using the closed-form gradient below. It can optionally add the entropy regularizer $\tau S(\rho_s)$ of Appendix~\ref{sec:entropy}.

\begin{theorem}[Closed-form gradient of a linear functional]\label{thm:gradient}
Let $\Psi=UU^\dagger$, $h=\Tr(\Psi)>0$, and $J(U)=\Tr\bigl(\rho(U)\,O\bigr)=\Tr(\Psi O)/h$ for a fixed Hermitian $O$. Then
\begin{equation}
\frac{\partial J}{\partial \Psi}=\frac{O-J(U)\,I}{h},
\qquad
\nabla_{\overline{U}} J = \frac{1}{h}\bigl(O-J(U)\,I\bigr)U ,
\end{equation}
where $\nabla_{\overline{U}}$ denotes the Wirtinger gradient with respect to $\overline{U}$ (entrywise complex conjugate of $U$). For real $U$ and real symmetric $O$, the ordinary real gradient is $\tfrac{2}{h}(O-J(U)I)U$.
\end{theorem}

The gradient is closed form for any Hermitian $O$, not only for an RL value operator. It needs no differentiation through an eigendecomposition and no projection step. It also never forms the $k\times k$ matrix $\rho$, because it works directly with the $k\times r$ factor $U$. For real $U$ and real symmetric $O$, no complex calculus is needed. Remark~\ref{rem:wirtinger} in Appendix~\ref{app:remarks} explains the Wirtinger convention for the complex case. DMPG sets $O=Q_s$, and Section~\ref{sec:limitations} gives its per-state cost.

\section{Density-matrix policies: decision-time equivalence with classical RL}\label{sec:policies}

We now use the density matrix as an RL policy. The goal here is narrow, which is to check that nothing in ordinary RL breaks. Each state gets one density matrix in place of the usual probability vector, and its diagonal recovers an ordinary action distribution.

\begin{definition}[Density-matrix policy]\label{def:dmpolicy}
A \emph{density-matrix policy} assigns to each state $s\in\Sset$ a density matrix $\rho_s\in\mathcal D_k$. When measured in the action basis, the probability of action $a_i$ is the diagonal element
$\Pr_{\rho}(a_i\mid s)=\Tr\bigl(\rho_s\,\eb_i\eb_i^\dagger\bigr)=(\rho_s)_{ii}$.
\end{definition}

The policy stores a matrix, but drawing an action uses only the diagonal, exactly as in classical RL. We ask what, if anything, the off-diagonal entries do here. All proofs for this section are in Appendix~\ref{app:proofs-policies}.

\begin{proposition}[Decision-time behavioral equivalence]\label{prop:MDPbehavior}
Consider an MDP in which, at each state $s$, an action is sampled from the classical policy $\pi(a_i\mid s)=(\rho_s)_{ii}$ and the reward is just a function of that single realized action, $R(s,a)$. In expectation, this equals $\Tr(\rho_s O_R)$, where $O_R=\diag\bigl(R(s,a_1),\dots,R(s,a_k)\bigr)$ is a \emph{diagonal observable}. It only assigns a number to each action and never couples two different actions. By Lemma~\ref{lem:phaseirrelevance}, the expected return depends on $\{\rho_s\}$ only through $\diag(\rho_s)$. In this case, the achievable behaviors coincide with the classical ones.
\end{proposition}

So in an ordinary MDP rollout, coherences are inert at decision time. Actions are sampled from the diagonal and scored by a diagonal observable, and DMRL reduces consistently to classical RL.

The same holds for the rest of the standard toolkit. For a diagonal value operator, the Gibbs optimum and its annealing law reduce to maximum-entropy RL. The matrix softmax map reduces to softmax policy improvement. The matrix trust region reduces to the KL-constrained update behind trust region policy optimization (TRPO) \cite{schulman2015trust}. Estimation is similar. A diagonal $O$ makes $\Tr(\rho_sO)$ independent of the coherences, so the off-diagonal entries can serve only as control-variate coefficients. Eigenmode averaging likewise matches classical stratified sampling. Appendix~\ref{sec:theory} also turns these variance reductions into sample-efficiency gain factors, written $g_{\mathrm{EA}}$ and $g_{\mathrm{CV}}$. These results are consistency checks. They are not advantages over well-tuned classical methods. We develop them in Appendices~\ref{sec:entropy} and~\ref{sec:theory}. Only the Gibbs optimum (Theorem~\ref{thm:gibbs}) stays in the main text, because Section~\ref{sec:beyond} uses it.

Outside this standard setting, however, coherences can matter in two ways. If the reference policy has coherences, they shape the trained policy through the regularizer (Section~\ref{subsec:coherentprior}). If the reward operator is non-diagonal, coherences act at decision time. There it is Theorem~\ref{thm:phaserelevance}, not the decision-time equivalence just proved, that governs decisions (Section~\ref{subsec:coupledreward}).

\section{Beyond the diagonal: where coherences matter}\label{sec:beyond}

The standard setting rests on two assumptions. The reward operator $Q_s$ is diagonal, and the regularizer is taken relative to the uniform state. Under these assumptions, coherences never move the sampled action. This section relaxes each assumption. The two ways differ in kind.

The first way, Section~\ref{subsec:coherentprior}, changes only the training objective. The action set stays $\{a_1,\dots,a_k\}$, exactly as in an ordinary MDP. The reward operator also stays diagonal. Only the reference policy used for regularization becomes a density matrix. So the comparison to classical RL is on identical ground.

The second way, Section~\ref{subsec:coupledreward}, changes what kind of action the agent chooses. The classical baseline there is restricted to one of $k$ mutually exclusive labels. The DMRL agent chooses a richer object instead, and the environment scores it accordingly. The two methods are then solving different decision problems, and any advantage comes from the larger space of actions available, not from the density matrix alone. We say this plainly again where the result appears.

All proofs for this section are in Appendix~\ref{app:proofs-beyond}.

\subsection{A coherent reference policy}\label{subsec:coherentprior}

We start with the first way, in which the reward operator $Q_s$ stays diagonal and only the reference policy changes.

Many training objectives keep an explicit KL term against a fixed reference policy. Offline and conservative RL keep a new policy close to a reference policy built from logged data \cite{levine2020offline}. In LLM fine-tuning, RLHF keeps it close to a supervised or previous-round reference policy \cite{ouyang2022training}. GRPO does the same, both in its original formulation and in its use for RLVR \cite{shao2024deepseekmath,guo2025deepseekr1}. Not every RLVR variant keeps this term. DAPO and several successors drop it on purpose, so that long chain-of-thought training can drift further from the initial model \cite{yu2025dapo}. Our construction replaces this term with matrix relative entropy toward a coherent reference. It applies wherever the term is present.

Usually the reference policy is a plain probability vector. Here we let it be a genuine density matrix $\rho_s^{\mathrm{ref}}$. Real data can motivate this. Two response styles, such as concise and formal, may tend to be expressed together within the same LLM response. A probability vector records how much each style is used on average, but not how much the styles overlap within a response, and $\rho_s^{\mathrm{ref}}$ can record both. Recorded actions alone cannot supply this information, because sampling reveals only the diagonal (Proposition~\ref{prop:MDPbehavior}). The data must also say how strongly each sample expresses each style. For example, suppose a classifier gives each recorded response style scores $w_1,\dots,w_k\ge0$ that sum to $1$. Setting $x_i=\sqrt{w_i}$ gives a unit vector, and each $\mathbf{x}\mathbf{x}^\top$ is PSD with trace $1$. So the average of $\mathbf{x}\mathbf{x}^\top$ over the data is a valid density matrix. Its diagonal holds the average style scores. Its off-diagonal entry $(i,j)$ is the average of $\sqrt{w_iw_j}$, which measures how much styles $i$ and $j$ overlap within the same response. It summarizes the scores and is not a calibrated estimate of any semantic dependence between the styles. Mixing in a small multiple of $I/k$ makes $\rho_s^{\mathrm{ref}}$ full rank, as Corollary~\ref{cor:coherentprior} below requires. This is one possible construction, and building coherent reference policies from data in general remains an open problem.

Whatever the source of $\rho_s^{\mathrm{ref}}$, our claim about it is narrow. We do not claim that a coherent reference policy improves training. We claim only that if the reference policy is a density matrix and the penalty is matrix relative entropy, the optimal policy has a closed form, and that form carries the reference policy's structure into the trained policy.

To state this result, we first recall the free-energy objective. Entropy-regularized RL replaces the raw value by a free-energy objective \cite{haarnoja2018soft,ziebart2008maximum}. The matrix version adds a von Neumann entropy bonus to the expected value $\Tr(\rho Q_s)$, where $Q_s$ is a Hermitian reward operator. In an ordinary MDP, it is the diagonal action-value operator defined in Section~\ref{sec:background}.

\begin{definition}[Matrix free-energy objective]\label{def:freeenergy}
At temperature $\tau>0$, the matrix free-energy objective is
\begin{equation}
F_\tau(\rho)=\Tr(\rho Q_s) + \tau S(\rho).
\end{equation}
\end{definition}

$F_\tau$ is maximized, not minimized, because $Q_s$ plays the role of a reward and not a physical energy. Large $\tau$ rewards spread-out policies (exploration). Small $\tau$ rewards concentrating on value (exploitation). The maximizer is a Gibbs state, which we define next.

\begin{definition}[Gibbs state]\label{def:gibbs}
For a Hermitian $Q\in\C^{k\times k}$ and a temperature $\tau>0$, the Gibbs state of $Q$ is $\rho_\tau(Q)=e^{Q/\tau}/\Tr\bigl(e^{Q/\tau}\bigr)$.
\end{definition}

The matrix $e^{Q/\tau}$ is Hermitian and positive definite, so $\rho_\tau(Q)$ has full rank. It gives more weight to eigenvectors of $Q$ with larger eigenvalues, because $Q$ is a reward and not a cost. If $Q$ is diagonal, so is $\rho_\tau(Q)$, and its diagonal is the softmax $e^{Q_{ii}/\tau}/\sum_j e^{Q_{jj}/\tau}$.

\begin{theorem}[Gibbs variational principle]\label{thm:gibbs}
For any Hermitian $Q_s$ and $\tau>0$, the unique maximizer of $F_\tau$ over $\mathcal D_k$ is the Gibbs state (Definition~\ref{def:gibbs}) of $Q_s$ at temperature $\tau$,
\begin{equation}
\rho_\tau(Q_s) \;=\; \frac{e^{Q_s/\tau}}{\Tr\, e^{Q_s/\tau}},
\qquad\text{with}\qquad
\max_{\rho\in\mathcal D_k}F_\tau(\rho)\;=\;\tau\log\Tr\, e^{Q_s/\tau}.
\end{equation}
\end{theorem}

The optimal entropy-regularized policy is an exponential reweighting of the value operator, and the optimal objective is the log-partition function. For diagonal $Q_s$, these are exactly the classical softmax policy and the log-sum-exp value.

Let $\rho_s^{\mathrm{ref}}\in\mathcal D_k$ be the reference policy. We assume $\rho_s^{\mathrm{ref}}\succ0$, so that $\log\rho_s^{\mathrm{ref}}$ is well defined. Regularizing toward $\rho_s^{\mathrm{ref}}$ instead of the uniform state gives the objective $F_\tau^{\mathrm{ref}}(\rho)=\Tr(\rho Q_s)-\tau\,D(\rho\Vert\rho_s^{\mathrm{ref}})$. For $\rho_s^{\mathrm{ref}}=I/k$ we have $D(\rho\Vert I/k)=\log k-S(\rho)$. So $F_\tau^{\mathrm{ref}}$ equals $F_\tau$ up to the constant $-\tau\log k$, and its maximizer is the Gibbs state of Theorem~\ref{thm:gibbs}.

For a general reference policy, expand $D(\rho\Vert\rho_s^{\mathrm{ref}})=-S(\rho)-\Tr(\rho\log\rho_s^{\mathrm{ref}})$. This gives $F_\tau^{\mathrm{ref}}(\rho)=\Tr(\rho\,\tilde Q_s)+\tau S(\rho)$, where $\tilde Q_s=Q_s+\tau\log\rho_s^{\mathrm{ref}}$. This has the form of $F_\tau$ from Definition~\ref{def:freeenergy}, with $Q_s$ replaced by $\tilde Q_s$. So Theorem~\ref{thm:gibbs} applies directly, and no new proof is needed. Note that $\tilde Q_s$ itself depends on $\tau$.

\begin{corollary}[Coherent-prior regularization]\label{cor:coherentprior}
For any Hermitian $Q_s$, any $\rho_s^{\mathrm{ref}}\in\mathcal D_k$ with $\rho_s^{\mathrm{ref}}\succ0$, and any $\tau>0$, the unique maximizer of $F_\tau^{\mathrm{ref}}$ over $\mathcal D_k$ is
\begin{equation}
\rho_\tau(\tilde Q_s) \;=\; \frac{\exp\bigl((Q_s+\tau\log\rho_s^{\mathrm{ref}})/\tau\bigr)}{\Tr\,\exp\bigl((Q_s+\tau\log\rho_s^{\mathrm{ref}})/\tau\bigr)}.
\end{equation}
As $\tau\to\infty$, $\rho_\tau(\tilde Q_s)\to\rho_s^{\mathrm{ref}}$. As $\tau\to0^+$, the weight of $\rho_\tau(\tilde Q_s)$ outside the top eigenspace of $Q_s$ tends to zero. If the top eigenvalue is simple, the limit is the projector onto its eigenvector. In that case the limit matches Corollary~\ref{cor:annealing} of Appendix~\ref{sec:entropy} and does not depend on $\rho_s^{\mathrm{ref}}$. If the top eigenvalue is degenerate, the limit inside the top eigenspace can depend on $\rho_s^{\mathrm{ref}}$.
\end{corollary}

The matrix $\log\rho_s^{\mathrm{ref}}$ is Hermitian, and it is non-diagonal exactly when $\rho_s^{\mathrm{ref}}$ has coherences. So $\log\rho_s^{\mathrm{ref}}$ acts as a new non-diagonal observable, and it enters training through $\tilde Q_s$. The next result states what this implies.

\begin{proposition}[Coherence transfer]\label{prop:coherencetransfer}
Let $\rho_\tau(\tilde Q_s)$ be as in Corollary~\ref{cor:coherentprior}, with $Q_s$ diagonal. Then $\rho_\tau(\tilde Q_s)$ is diagonal in the action basis if and only if $\rho_s^{\mathrm{ref}}$ is diagonal in the action basis.
\end{proposition}

So whenever the reference policy has coherences, the trained policy has coherences of its own. This is not a side effect of estimation. It comes directly from the KL term toward the reference policy. A classical method that tracks only $\diag(\rho_s^{\mathrm{ref}})$ cannot represent how the trained policy depends on the reference policy's coherences. Example~\ref{ex:coherentprior} in Appendix~\ref{app:constructions} shows a concrete case. As the coherences of the reference policy grow, the classical action distribution stays fixed, while the DMRL action distribution moves steadily away from it.

This comparison holds the diagonal of the reference policy fixed. It is not a claim about expressivity, since a classical method could match the DMRL action distribution with a different diagonal reference policy. The difference is informational. The DMRL answer depends on the coherences of the reference policy, and a diagonal representation does not store them.

The operator $\tilde Q_s$ also matters beyond the optimum. It is non-diagonal whenever $\rho_s^{\mathrm{ref}}$ has coherences, so Theorem~\ref{thm:phaserelevance} applies to it directly. Two policies with the same diagonal can therefore receive different scores under it. This separating observable is not hypothetical. It is the reward-plus-reference operator that already sits inside $F_\tau^{\mathrm{ref}}$ in any run regularized toward a coherent reference policy.

\subsection{A coupled reward on blended actions}\label{subsec:coupledreward}

We now turn to the second way, in which the reward itself is non-diagonal and the action model changes along with it. Elsewhere we write the reward operator as $Q_s$, because in the diagonal case its entries are the action values $Q(s,a_i)$. A non-diagonal operator has no such interpretation. So in this subsection we write it as $O$, the general observable of Definition~\ref{def:observable}. As in Appendix~\ref{sec:theory}, we take $\rho$ and every observable to be real symmetric.

An ordinary MDP cannot produce a non-diagonal $O$. There the agent chooses one action label, and the environment pays for that label. This is the diagonal case of Section~\ref{sec:policies}. It holds even when a reward model scores each sampled response through a shared internal representation, because the expected reward of each strategy is still one number. A non-diagonal $O$ needs an environment that scores something richer than a label. Table~\ref{tab:actionmodels} contrasts the two action models. The remainder of this subsection focuses on the right-hand column, so every comparison with the classical baseline below involves this change of action model. \\

\begin{table}[h]
\centering
\small
\caption{Two action models. The standard basis vectors $\eb_i$ are unit vectors, and choosing $\mathbf{x}=\pm \eb_i$ earns $\mathbf{x}^\top O\mathbf{x}=O_{ii}$, the reward of action $a_i$. So an agent that chooses only $\pm \eb_1,\dots,\pm \eb_k$ is exactly the left column. This restriction is the classical baseline used throughout this subsection.}
\label{tab:actionmodels}
\begin{tabular}{L{3.0cm}L{4.6cm}L{4.6cm}}
\toprule
 & Ordinary RL & Blended-action RL \\
\midrule
Action & one label $a_i$ & one unit vector $\mathbf{x}\in\R^k$ \\
Reward & $R_i$, one number per label & $R(\mathbf{x})=\mathbf{x}^\top O\mathbf{x}$ \\
Policy statistic & $p_i=\pi(a_i)$ & $\rho=\E_{\pi}[\mathbf{x}\mathbf{x}^\top]$ \\
\bottomrule
\end{tabular}
\end{table} 

\begin{definition}[Blended-action environment]\label{def:structuredaction}
Fix a real symmetric $O\in\R^{k\times k}$. The environment is one-step (a bandit). In each round, the agent chooses a unit vector $\mathbf{x}\in\R^k$ instead of a single action label. The environment returns a reward $R$ with
\begin{equation}
\E[R\mid \mathbf{x}]=\mathbf{x}^\top O\mathbf{x}=\sum_ix_i^2O_{ii}+\sum_{i\ne j}x_ix_jO_{ij}.
\end{equation}
The chosen $\mathbf{x}$ affects nothing but this reward. The off-diagonal entries contribute only when $\mathbf{x}$ is not a standard basis vector, that is, when it has several nonzero components and so mixes actions. As in classical RL, we write $\pi$ for the agent's policy. Here it is a probability distribution over unit vectors instead of action labels. A classical policy is the special case in which $\pi$ puts all its probability on the standard basis vectors $\pm \eb_i$, where $\eb_i$ is the vector for action $a_i$. We define $\rho=\E_{\mathbf{x}\sim\pi}[\mathbf{x}\mathbf{x}^\top]$ as the average of the matrix $\mathbf{x}\mathbf{x}^\top$ over the random choice of $\mathbf{x}$, and call it the second-moment matrix of $\pi$. If the rounds come with states, the states are drawn independently of past actions (a contextual bandit), and everything below applies at each state separately.
\end{definition}

\begin{proposition}[Second-moment representation]\label{prop:secondmoment}
For every distribution $\pi$ over unit vectors, its second-moment matrix $\rho=\E_\pi[\mathbf{x}\mathbf{x}^\top]$ lies in $\mathcal D_k$ and $\E_\pi[R]=\Tr(\rho\, O)$. Conversely, every real symmetric $\rho\in\mathcal D_k$ is the second-moment matrix of some $\pi$. One such $\pi$ chooses the eigenvector $\mathbf{v}_n$ of $\rho$ with probability $\lambda_n$. This is a valid choice because the eigenvalues of $\rho$ are non-negative and sum to $1$.
\end{proposition}

The entry $\rho_{ii}=\E[x_i^2]$ is the average weight the chosen vector puts on action $i$. The entry $\rho_{ij}=\E[x_ix_j]$ is the cross-moment between components $i$ and $j$. So the coherences are ordinary statistics of $\mathbf{x}$, not a new kind of object. For example, with $k=2$, a policy that always chooses $\mathbf{x}=(\eb_1+\eb_2)/\sqrt2$ has $\rho=\mathbf{x}\mathbf{x}^\top$, with diagonal entries $\tfrac12$ and off-diagonal entries $\tfrac12$. It puts equal weight on the two actions, and the off-diagonal entry records that it always blends them. Two behaviors with the same $\rho$ earn the same expected reward. In this one-step setting, expected reward is the whole objective, which is why $\rho$ is the natural summary of a policy here. From here on, when we say a policy earns a reward, we mean its expected reward $\Tr(\rho\, O)$.

The unit vector $\mathbf{x}$ has $k$ real-valued entries, one per base action, so it varies continuously over the unit sphere. One could therefore use an ordinary continuous-action RL method, for example a Gaussian policy over $\mathbf{x}$ rescaled to unit length, and never form $\rho$. Why use a density matrix? The reason is that the expected reward is exactly quadratic in $\mathbf{x}$. By Proposition~\ref{prop:secondmoment}, the expected reward then depends on the policy only through $\rho$, so $\rho$ keeps everything that matters and discards the rest. This relies on the quadratic form. A reward of higher degree in $\mathbf{x}$, such as a quartic one, would depend on higher moments of $\mathbf{x}$, and $\rho$ alone would not be enough. When the reward is quadratic, the compression pays off in several ways. A distribution over the whole unit sphere becomes one $k\times k$ matrix. The set of such matrices is convex and compact (Proposition~\ref{prop:inclusion}). Their rank trades fidelity against parameter cost (Appendix~\ref{app:rank}). They carry a canonical entropy (Appendix~\ref{sec:entropy}) and a relative entropy that gives a trust region (Theorem~\ref{thm:pinching}). Restricting them to the diagonal recovers ordinary RL exactly. So a density matrix is not the only way to represent a policy over blended actions, but it is a compact one that brings the rest of this paper's machinery with it.

We now state the basic fact that makes a non-diagonal $O$ matter.

\begin{proposition}[Coherent dominance]\label{prop:coherentdominance}
Fix $k$ actions and a probability vector $\mathbf{p}\in\Delta_k$ with $p_i>0$ for every $i$. As in Appendix~\ref{sec:theory}, write $\rho_{\mathrm{diag}}=\diag(\mathbf{p})$ for the diagonal matrix with $\mathbf{p}$ on its diagonal. Let $\mathcal D_k(\mathbf{p})$ be the set of real symmetric $\rho\in\mathcal D_k$ with $\diag(\rho)=\mathbf{p}$. By Proposition~\ref{prop:secondmoment}, these are exactly the second-moment matrices with diagonal $\mathbf{p}$. Let $O=O^\top$ have an off-diagonal entry $O_{ij}\ne0$. Then
\begin{equation}
\max_{\rho\in\mathcal D_k(\mathbf{p})} \Tr(\rho\, O) \;>\; \Tr(\rho_{\mathrm{diag}}\,O).
\end{equation}
\end{proposition}

In the blended-action environment, $p_i=\E[x_i^2]$. By Proposition~\ref{prop:secondmoment}, every $\rho\in\mathcal D_k(\mathbf{p})$ is the second-moment matrix of some policy with action probabilities $\mathbf{p}$. The classical policy with these probabilities has matrix $\rho_{\mathrm{diag}}$ and earns $\sum_ip_iO_{ii}$, the usual diagonal formula. Proposition~\ref{prop:coherentdominance} says that, if $O$ is non-diagonal, then for every action distribution with all probabilities positive, some coherent policy earns strictly more expected reward than every classical policy with the same action probabilities. A classical learner cannot close this gap. Its entire hypothesis class is $\{\rho_{\mathrm{diag}} : \mathbf{p}\in\Delta_k\}$, and it has no coherences to adjust. So the shortfall comes from the policy class, not from the optimizer, and more data or a stronger optimizer cannot remove it. Example~\ref{ex:toy} in Appendix~\ref{app:constructions} is an instance. There, the classical policy and a coherent policy both give each action probability $\tfrac12$. Yet the classical policy scores $0$, and the coherent one scores $1$.

The extra reward comes from a larger set of actions. A classical policy chooses only standard basis vectors, while a blended-action policy can choose any unit vector. So the classical policies form a strict subset of the policies available here, as the two columns of Table~\ref{tab:actionmodels} show. The matrix representation by itself adds no reward.

Proposition~\ref{prop:coherentdominance} compares policies with the same action probabilities. The next proposition compares against the best classical policy overall.

\begin{proposition}[Margin over the best classical policy]\label{prop:gap}
Let $O=O^\top\in\R^{k\times k}$. Over real symmetric $\rho\in\mathcal D_k$, the maximum of $\Tr(\rho\,O)$ is $\lambda_{\max}(O)$, the largest eigenvalue of $O$. Over classical policies, the maximum of $\sum_ip_iO_{ii}$ is $\max_iO_{ii}$. Define $\mathrm{margin}(O)=\lambda_{\max}(O)-\max_iO_{ii}$. It is never negative. It equals zero exactly when some action $i$ is uncoupled from all others, meaning $O_{ij}=0$ for every $j\ne i$, and its own reward satisfies $O_{ii}=\lambda_{\max}(O)$. In particular, $\mathrm{margin}(O)>0$ whenever every action is coupled to at least one other action.
\end{proposition}

The maximum over $\mathcal D_k$ is the low-temperature limit in Corollary~\ref{cor:annealing}. In Example~\ref{ex:toy} the margin equals $1$. The margin can also grow with $k$. For example, take $O$ with $O_{ii}=0$ and $O_{ij}=1$ for every $i\ne j$. Every entry of $O$ has magnitude at most $1$, yet $\mathrm{margin}(O)=k-1$.

The other results of Appendix~\ref{sec:entropy} also apply here, with $O$ in place of $Q_s$. The Gibbs state $\rho_\tau(O)$ maximizes $\Tr(\rho O)+\tau S(\rho)$ by Theorem~\ref{thm:gibbs}. Sharpening a policy with the softmax map never lowers its expected reward, by Theorem~\ref{thm:monotonic}.

Where could a non-diagonal $O$ come from? One construction is an evaluator that judges the chosen vector against its own $k$ reference directions, instead of against the agent's actions. For $k=2$, the evaluator might check whether $\mathbf{x}$ points along $\mathbf{b}_1=(\eb_1+\eb_2)/\sqrt2$, the two actions used together, or along $\mathbf{b}_2=(\eb_1-\eb_2)/\sqrt2$, the two actions used against each other. Its directions are the action directions rotated, hence the name.

\begin{corollary}[Rotated-basis evaluator]\label{cor:coupledmeasurement}
Let $\mathbf{b}_1,\dots,\mathbf{b}_k$ be an orthonormal basis of $\R^k$, the evaluator's reference directions. The evaluator receives the chosen unit vector $\mathbf{x}$ and assigns it to direction $n$ with probability $(\mathbf{b}_n^\top \mathbf{x})^2$. These probabilities are non-negative and sum to $\|\mathbf{x}\|^2=1$. The evaluator then pays a reward $R_n$ that depends only on $n$. Averaging over $\mathbf{x}\sim\pi$, direction $n$ is chosen with probability $\Tr(\rho\, \mathbf{b}_n\mathbf{b}_n^\top)$. Hence the expected reward is
\begin{equation}
\E[R]=\Tr(\rho\, O),\quad\text{where}\quad O=\sum_n R_n\, \mathbf{b}_n\mathbf{b}_n^\top.
\end{equation}
If the evaluator's directions are the action directions, $\mathbf{b}_n=\pm \eb_n$, then $O$ is diagonal, and this is ordinary RL. If the rewards $R_n$ are distinct, $O$ has a nonzero off-diagonal entry exactly when the evaluator's directions are not the action directions, up to reordering and sign flips. The distinctness condition matters. If some rewards repeat, a rotated set of directions can still give a diagonal $O$. For instance, if every $R_n$ equals the same value $R_0$, then $O=R_0 I$ whatever the directions are. Whenever $O$ has an off-diagonal entry, Proposition~\ref{prop:coherentdominance} applies.
\end{corollary}

For the directions $\mathbf{b}_1$ and $\mathbf{b}_2$ above, with $R_1=1$ and $R_2=-1$, the matrix $O$ has zeros on the diagonal and ones off it. This is the reward of Example~\ref{ex:toy}.

The blended-action environment is an explicit mechanism and not a claim about existing systems. Two scenarios might supply such a reward. First, an LLM pipeline could consolidate several reasoning paths into one answer, as the example below describes. Second, in robotics, sensors of different modalities, such as cameras, touch sensors, and force sensors, may each respond to a combination of actions rather than to a single action. A reward built from their squared responses, $\sum_nR_n(\mathbf{b}_n^\top \mathbf{x})^2$, again has the form $\mathbf{x}^\top O\mathbf{x}$ with $O=\sum_nR_n\mathbf{b}_n\mathbf{b}_n^\top$. It is non-diagonal whenever the sensors mix actions, unless the mixtures cancel. The sensor directions need not be orthonormal, so this example is slightly more general than Corollary~\ref{cor:coupledmeasurement}. We do not show that current systems work this way, and we treat both as possible applications.

Whatever produces $O$, whether an evaluator as in Corollary~\ref{cor:coupledmeasurement} or one of the scenarios above, the agent does not know it. As with action values in ordinary RL, it must learn $O$ from the rewards it receives. Here learning reduces to regression. By the cyclic property of the trace, the expected reward $\mathbf{x}^\top O\mathbf{x}=\Tr(\mathbf{x}\mathbf{x}^\top O)$ is linear in the entries of $O$, with the matrix $\mathbf{x}\mathbf{x}^\top$ as known features. So the agent can fit an estimate $\hat O$ by ridge regression on its past choices and rewards. It then takes a gradient step on the factor $U$, using Theorem~\ref{thm:gradient} with $\hat O$ in place of $O$. The policy being optimized chooses the eigenvector $\mathbf{v}_n$ of $\rho(U)$ with probability $\lambda_n$ (Proposition~\ref{prop:secondmoment}). To explore, the agent occasionally chooses a uniformly random unit vector instead. These choices are not part of the policy being optimized. Algorithm~\ref{alg:dmpgcoupled} in Appendix~\ref{app:algorithms} gives the steps. We do not analyze its convergence here. The Gibbs state and the softmax map mentioned earlier could replace the uniform exploration step. We do not analyze that use either.

We now return to the first scenario above. Suppose an LLM pipeline consolidates $k$ reasoning paths into one answer, and a judge scores that answer. Pipelines of this kind are already in use. Self-consistency takes a majority vote over sampled paths \cite{wang2023selfconsistency}. Mixture-of-Agents has an aggregator write one answer from several responses \cite{wang2024mixture}. Parallel-thinking systems generate several paths and combine them \cite{google2025deepthink,wen2025parathinker,zheng2025parallel}. Verifiers and LLM judges can score the results \cite{cobbe2021training,lightman2024lets,zheng2023judging}. The pipeline can learn from the judge's scores how much weight each path should receive. We consider learning during training, across many prompts. With the prompt as the state, this is a one-step problem of the kind in Definition~\ref{def:structuredaction}. Learning weights for a single prompt at test time is a small bandit problem whose answer is one weight vector. It needs no policy, so we do not pursue it. Write the weights as $x_i^2$ for a unit vector $\mathbf{x}$. If the judge rewards complementary paths and penalizes conflicting ones, these pairwise effects form a non-diagonal $O$. Consolidating a complementary pair of paths into one answer then earns strictly more than using the two paths separately with the same frequencies, in line with Proposition~\ref{prop:coherentdominance}. With non-negative weights $w_i$, we have $x_i=\sqrt{w_i}$, and the assumed score is $\sum_iw_iO_{ii}+\sum_{i\ne j}\sqrt{w_iw_j}\,O_{ij}$. So each pairwise term grows with the geometric mean of the two weights, and only complementary pairs, with $O_{ij}>0$, can raise the score. With non-negative weights, the best achievable score is the maximum of $\mathbf{x}^\top O\mathbf{x}$ over non-negative unit vectors. It equals $\lambda_{\max}(O)$ when every off-diagonal entry of $O$ is non-negative, but it can be smaller otherwise, and it is then harder to compute. So the margin of Proposition~\ref{prop:gap} applies to this application only in the non-negative case. Whether any judge scores consolidated answers in this way is an open question. A direct test compares the judge's score for a consolidated answer with the weighted average of its scores for the individual paths, using the same weights. A consistent excess for a pair of paths would indicate $O_{ij}>0$. Example~\ref{ex:llmpaths} in Appendix~\ref{app:constructions} works through this case with numbers.

The results so far rely on Definition~\ref{def:structuredaction} being one-step. If the chosen $\mathbf{x}$ also influenced the next state, $\rho$ would in general no longer determine the expected return. Requiring the transition to depend on $\mathbf{x}$ only through $\mathbf{x}\mathbf{x}^\top$ is not enough, because a nonlinear function of $\mathbf{x}\mathbf{x}^\top$ can separate two behaviors with the same second moment. Take $k=2$. One policy $\pi$ chooses $\mathbf{x}=\eb_1$ or $\mathbf{x}=\eb_2$, each with probability $\tfrac12$, so its second-moment matrix is $\rho_\pi=\tfrac12\eb_1\eb_1^\top+\tfrac12\eb_2\eb_2^\top=I/2$. Another policy $\pi'$ chooses $\mathbf{x}=(\eb_1+\eb_2)/\sqrt2$ or $\mathbf{x}=(\eb_1-\eb_2)/\sqrt2$, each with probability $\tfrac12$, giving the same second-moment matrix $\rho_{\pi'}=I/2$. Now let the next state be $s_1$ if $\mathbf{x}\mathbf{x}^\top$ is diagonal and $s_2$ otherwise. Then $\pi$ always moves to $s_1$ and $\pi'$ always moves to $s_2$, although $\rho_\pi=\rho_{\pi'}=I/2$. This rule depends on $\mathbf{x}$ only through $\mathbf{x}\mathbf{x}^\top$, but it is not linear in $\mathbf{x}\mathbf{x}^\top$. As a result, $\rho$ alone does not determine the probability that a policy moves to $s_1$. What suffices is a transition that is linear in $\mathbf{x}\mathbf{x}^\top$.

\begin{proposition}[Multi-step extension]\label{prop:multistep}
Consider a finite-horizon episodic environment, with the time step included in the state. At each non-terminal state $s$, the agent chooses a unit vector $\mathbf{x}$ and receives a random reward $R$ with $\E[R\mid s,\mathbf{x}]=\mathbf{x}^\top O_s\mathbf{x}$, where $O_s$ is real symmetric. The environment then moves to state $s'$ with probability
\begin{equation}
\mathcal P(s'\mid s,\mathbf{x})=\mathbf{x}^\top M_{s'}(s)\,\mathbf{x},
\end{equation}
where $M_{s'}(s)\succeq0$ and $\sum_{s'}M_{s'}(s)=I$. Then two policies with the same second-moment matrix $\rho_s=\E[\mathbf{x}\mathbf{x}^\top]$ at every state have the same expected return from every state.
\end{proposition}

Proposition~\ref{prop:multistep} extends Proposition~\ref{prop:secondmoment} from one step to many. Fix a state $s$. The policy chooses $\mathbf{x}$ at random, so the reward and the next state are both averaged over $\mathbf{x}$. Since $\mathbf{x}^\top O_s\mathbf{x}=\Tr(\mathbf{x}\mathbf{x}^\top O_s)$ and expectation is linear, the expected reward is $\E[\mathbf{x}^\top O_s\mathbf{x}]=\Tr(\E[\mathbf{x}\mathbf{x}^\top]\,O_s)=\Tr(\rho_sO_s)$. The same steps show that the probability of moving to $s'$ is $\E[\mathbf{x}^\top M_{s'}(s)\,\mathbf{x}]=\Tr(\rho_sM_{s'}(s))$. Both depend on the policy at $s$ only through $\rho_s$. Every transition moves to a later time step, so values can be computed backward from the final step. By induction, each value depends on the policy only through the matrices $\rho_s$.

The closed-form results of this subsection do not extend. To see the obstacle, let $V^\rho$ be the value function of the policy with second-moment matrices $\{\rho_s\}$, which is well defined by Proposition~\ref{prop:multistep}. The Bellman equation and the two expectations above give $V^\rho(s)=\Tr(\rho_sO_s)+\sum_{s'}\Tr(\rho_sM_{s'}(s))\,V^\rho(s')=\Tr(\rho_sO^\rho_s)$, where $O^\rho_s=O_s+\sum_{s'}V^\rho(s')M_{s'}(s)$. This matrix is the non-diagonal analogue of the action-value operator $Q_s$ of Section~\ref{sec:background}. The state-$s$ problem therefore resembles the one-step problem with $O_s$ replaced by $O^\rho_s$. But $O^\rho_s$ depends on $V^\rho$, and $V^\rho$ depends on the policy at later states. So Propositions~\ref{prop:coherentdominance} and~\ref{prop:gap} do not carry over directly. Extending them remains an open problem.

\section{Related work}\label{sec:related}

In DMRL, the density matrix is a classical data structure. Optimization uses classical resources. The claimed advantages are representational. No computational speedup is implied.

Quantum reinforcement learning \cite{jerbi2021parametrized,biamonte2017quantum} is different. There, the density matrix describes the physical state of a quantum device, and the hoped-for advantage is computational. Quantum-inspired RL \cite{dong2008quantum} keeps an amplitude vector over actions and updates it with Grover-style iterations, as an exploration heuristic. Quantum cognition \cite{busemeyer2012quantum} uses quantum probability descriptively, to fit human decision data. Neither of these two lines uses a density matrix as an RL policy, gives its off-diagonal entries an operational role, or proves guarantees like those developed here.

PSD matrices have also been used to model non-negative functions and probability densities \cite{marteauferey2020nonneg,rudi2021psd,marteauferey2022sampling}. There they serve density modeling and sampling. Here the PSD matrix is a policy. We use unit-trace matrices because their diagonal is a probability vector, and because their entropy and relative entropy reduce to the classical ones on diagonal matrices.

\section{Limitations}\label{sec:limitations}

Four caveats apply to our results. First, the paper is entirely theoretical. The numerical examples in Appendix~\ref{app:constructions} are illustrative calculations, not empirical evaluations. We do not measure whether the results reduce training samples on a real task. The coupled-reward results hold only in the one-step blended-action environment of Definition~\ref{def:structuredaction}, with a quadratic reward and real symmetric $\rho$ and $O$. The second-moment representation extends to multi-step environments whose transitions are linear in $\mathbf{x}\mathbf{x}^\top$ (Proposition~\ref{prop:multistep}), but the closed-form results do not. Whether real reward models or sensors provide such an interface is an open empirical question. We also do not compare against a generic PSD or Gram-matrix parameterization. So we cannot say which benefits need the trace-one normalization and matrix entropy.

Second, our guarantees are proved state by state, at a single snapshot of the policy (Remark~\ref{rem:acrossstates}). With shared parameters, monotonic improvement can fail across states, as in classical deep RL. The gain factors $g_{\mathrm{EA}}$ and $g_{\mathrm{CV}}$ (Appendix~\ref{sec:theory}) are also evaluated at the current policy. Training is nonstationary, so the theory does not show that these gains compound into fewer environment interactions.

Third, computation has a real cost. A dense implementation uses $\mathcal O(k^2)$ memory and $\mathcal O(k^3)$ time per eigendecomposition. A rank-$r$ factorization (Definition~\ref{def:factorparam}) reduces these to $\mathcal O(kr)$ parameters and $\mathcal O(kr^2)$ time. A gradient step costs $\mathcal O(kr)$ for a diagonal $Q_s$ and $\mathcal O(k^2r)$ for a dense one.

Fourth, learning coherences requires a signal. Diagonal rewards supply none (Lemma~\ref{lem:phaseirrelevance}). The two settings of Section~\ref{sec:beyond} do supply one. Outside them, supplying a signal is an open problem. Off-diagonal structure is also harder to inspect than probabilities (Remark~\ref{rem:diagnostics}).

\section{Conclusion}\label{sec:conclusion}

We asked what information an RL policy omits. A standard policy cannot record how actions relate to each other. Storing the policy as a density matrix fixes this. Its off-diagonal entries, called coherences, record pairwise structure.

Coherences change outcomes in two settings. In the first, the reference policy has coherences and the penalty is matrix relative entropy. The trained policy is then non-diagonal. Its action probabilities depend on coherences that a diagonal-only method never sees (Corollary~\ref{cor:coherentprior}, Example~\ref{ex:coherentprior}). In the second, a one-step agent chooses a blended action and the reward is a non-diagonal quadratic. Some coherent policy then earns strictly more than every classical policy with the same action probabilities. The margin over the best classical policy is $\lambda_{\max}(O)-\max_iO_{ii}$ (Propositions~\ref{prop:coherentdominance} and~\ref{prop:gap}).

Everywhere else, DMRL costs extra compute and nothing more. Its value therefore rests on the two settings above. The coherent reference policy needs data that score each sample on several actions at once. The blended-action environment needs a reward that scores blended actions. Both settings can be tested directly, starting with a synthetic environment with a known $O$. The answer will tell us when a policy should record how its actions relate, and when one number per action is enough.

% neurips_*.sty sets \flushbottom; the last body page (Limitations + Conclusion) cannot fill
% the page because References starts on the next one, so flushbottom stretches the glue
% between paragraphs and headings. Let that page end naturally.
\raggedbottom

\appendix
% neurips_*.sty sets \flushbottom, which stretches the glue around unbreakable blocks
% (algorithms, long displays) into large blank gaps. Relax it in the appendix only.
\raggedbottom
% With \raggedbottom in force, the tight 4pt \intextsep set in the preamble is no longer needed.
% Restore a more comfortable gap around the [H] algorithm floats, so the sentences just above
% each algorithm are not crowded.
\setlength{\intextsep}{8pt plus 2pt minus 2pt}

\section*{Overview of the Appendix}

The appendix has seven lettered sections, A through G. Appendix~\ref{app:rank} shows how the rank of $\rho$ trades fidelity against parameter cost. Appendices~\ref{sec:entropy} and~\ref{sec:theory} show that DMRL matches or transfers classical results in a standard setting. Appendix~\ref{sec:entropy} covers closed-form annealing, monotonic improvement of the matrix softmax map, and transfer of trust-region guarantees. Appendix~\ref{sec:theory} covers the variance reductions from eigenmode averaging and optimal coherence. It also gives the resulting sample-efficiency gain factors. These are consistency results. They are not needed for the two settings of Section~\ref{sec:beyond} in which coherences change outcomes. The one exception is the Gibbs variational principle (Theorem~\ref{thm:gibbs}), which stays in the main text. Appendix~\ref{app:proofs} gives the proofs of all results. They are grouped by the section or appendix that states each result, in the order of the paper. Appendix~\ref{app:algorithms} gives pseudocode for DMPG, density-matrix policy iteration (DMPI), adaptive-coherence policy evaluation (ACPE), and DMPG in the blended-action environment. Appendix~\ref{app:constructions} presents the worked examples. These include the $2\times2$ case, the estimator construction for standard MDPs, the coherent reference policy with numbers, and the consolidation of LLM reasoning paths. Appendix~\ref{app:remarks} contains deferred remarks. These cover the gradient convention, the support restriction of the trust region, the kernel of $\Gamma$, the feasibility scaling, the trade-off between return and variance, the adaptive estimator, and diagnostics for coherences.

\section{Rank as a capacity--complexity knob}\label{app:rank}

Definition~\ref{def:factorparam} makes rank an easy knob to set. Any nonzero $U\in\C^{k\times r}$ gives a valid $\rho(U)$ of rank at most $r$. This section asks what turning that knob costs. The proofs are in Appendix~\ref{app:proofs-rank}.

Suppose we truncate $\rho$ to its top $r$ eigenmodes. This introduces an error in any linear evaluation $\Tr(\rho O)$. The error depends only on how much spectral mass the discarded modes carry. Nothing here is specific to RL. The same bound applies to any linear evaluation of a density matrix.

\begin{definition}[Rank-$r$ capture]\label{def:capture}
Let $\rho\in\mathcal D_k$ have eigenvalues $\lambda_1\ge\lambda_2\ge\dots\ge\lambda_k\ge0$ in decreasing order, with eigenvectors $\{\mathbf{v}_n\}$ as in Section~\ref{sec:background}. For $1\le r\le k$, the \emph{rank-$r$ capture} is
\begin{equation}
\mathrm{cap}(r)=\sum_{n=1}^r\lambda_n\ \in[1/k,\,1].
\end{equation}
The \emph{rank-$r$ truncation} $\rho_r$ of $\rho$ is the renormalized top-$r$ projection
\begin{equation}
\rho_r=\frac{1}{\mathrm{cap}(r)}\sum_{n=1}^r\lambda_n\mathbf{v}_n\mathbf{v}_n^\dagger\ \in\mathcal D_k .
\end{equation}
\end{definition}

The eigenvalues sum to $1$, so $\mathrm{cap}(r)$ is the fraction of the total spectral mass held by the top $r$ modes. Also, $\lambda_1\ge1/k$. Hence $\mathrm{cap}(r)\ge1/k>0$ for every $r$, and $\rho_r$ is always well defined. The capture $\mathrm{cap}(r)$ is nondecreasing in $r$, and $\mathrm{cap}(k)=1$. The matrix $\rho_r$ is Hermitian, PSD, of unit trace, and of rank at most $r$. So $\rho_r\in\mathcal D_k$ by the same argument as Definition~\ref{def:dm}. The quantity $1-\mathrm{cap}(r)$ is the spectral mass discarded by keeping only the top $r$ modes. This discarded mass, not $r$ itself, governs the truncation error.

\begin{theorem}[Spectral truncation error]\label{thm:spectraltruncation}
Let $\rho\in\mathcal D_k$ and $1\le r\le k$. For any Hermitian observable $O$ with operator norm $\|O\|_{\mathrm{op}}$,
\begin{equation}
\bigl|\Tr(\rho O)-\Tr(\rho_r O)\bigr|\;\le\;2\,\|O\|_{\mathrm{op}}\,\bigl(1-\mathrm{cap}(r)\bigr).
\end{equation}
\end{theorem}

The left side is the error in $\Tr(\rho O)$ from using a rank-$r$ policy in place of the full one. The bound controls it through the leftover mass $1-\mathrm{cap}(r)$ alone. It assumes nothing about $O$ beyond its operator norm. It assumes nothing about which modes are discarded. It has no separate dependence on $k$ beyond what is already inside $\mathrm{cap}(r)$.

\begin{corollary}[Effective rank at a target fidelity]\label{cor:effectiverank}
For $\varepsilon\in(0,1]$, define the \emph{effective rank}
\begin{equation}
r_{\mathrm{eff}}(\varepsilon)=\min\{r\in\{1,\dots,k\}:\mathrm{cap}(r)\ge1-\varepsilon\}.
\end{equation}
Then for every observable $O$ with $\|O\|_{\mathrm{op}}\le1$, the rank-$r_{\mathrm{eff}}(\varepsilon)$ truncation evaluates within $2\varepsilon$ of $\rho$ itself.
\end{corollary}

The effective rank $r_{\mathrm{eff}}(\varepsilon)$ is the smallest rank needed so that the truncation error stays below $2\varepsilon$. It depends only on the spectrum of $\rho$. A single eigendecomposition computes it, with no search over candidate ranks.

Purity is another quantity that depends only on the spectrum. Like entropy, it measures how spread out the eigenvalues are, but it is simpler to compute.

\begin{definition}[Purity]\label{def:purity}
For $\rho\in\mathcal D_k$, the purity is $\Tr(\rho^2)=\sum_n\lambda_n^2\in[1/k,\,1]$.
\end{definition}

Purity equals $1$ exactly when $\rho$ has rank one. It equals $1/k$ exactly when all eigenvalues are equal, that is, when $\rho=I/k$. Its inverse $1/\Tr(\rho^2)$ is the participation ratio, which roughly counts how many eigenmodes carry weight. Purity can also be computed directly from the entries of $\rho$, since $\Tr(\rho^2)=\sum_i\rho_{ii}^2+\sum_{i\neq j}|\rho_{ij}|^2$. The first sum uses only the action probabilities. It is large when the policy favors a few actions and small when the policy is spread out. The second sum is the squared size of the coherences. If $\rho$ is diagonal, the second sum vanishes and purity is just $\sum_i\rho_{ii}^2$. The next result uses purity to bound the rank a policy needs.

\begin{proposition}[Purity lower-bounds effective rank]\label{prop:purityrank}
For $\rho\in\mathcal D_k$ and any $1\le r\le k$,
\begin{equation}
\mathrm{cap}(r)\le\sqrt{r\,\Tr(\rho^2)}.
\end{equation}
Consequently $r_{\mathrm{eff}}(\varepsilon)\ge(1-\varepsilon)^2/\Tr(\rho^2)$ for every $\varepsilon\in(0,1]$. So purity alone certifies that no smaller rank can reach the target capture level $1-\varepsilon$.
\end{proposition}

The proposition says that $r_{\mathrm{eff}}(\varepsilon)$ is at least $(1-\varepsilon)^2$ times the participation ratio $1/\Tr(\rho^2)$. Purity is easy to compute, so it gives a quick check on how compressible a policy can be. It also serves as a diagnostic (Remark~\ref{rem:diagnostics}).

For a spread-out spectrum with purity close to $1/k$, the bound gives $r_{\mathrm{eff}}(\varepsilon)\ge(1-\varepsilon)^2k$. So $r_{\mathrm{eff}}$ is close to $k$ for small $\varepsilon$, and no small rank will do. For a near-pure $\rho$ with purity close to $1$, the proposition gives almost no constraint, and a short separate argument is needed. Since $\Tr(\rho^2)=\sum_n\lambda_n^2\le\lambda_1$, we have $\mathrm{cap}(1)\ge\Tr(\rho^2)$. Hence $r_{\mathrm{eff}}(\varepsilon)=1$ whenever $\Tr(\rho^2)\ge1-\varepsilon$. A single eigenmode then captures almost all of $\rho$.

Together, Theorem~\ref{thm:spectraltruncation} and Proposition~\ref{prop:purityrank} make rank an explicit capacity--complexity knob, not just an implementation detail. The capture $\mathrm{cap}(r)$ gives the fidelity of a given rank in closed form. Purity gives an easily computed upper bound on it. Both hold before any coherences are estimated or trained. Both are independent of the parameter-estimation cost of reaching that rank.

\section{Rank-aware exploration: closed-form Gibbs optima and a principled annealing law}\label{sec:entropy}

Section~\ref{sec:policies} showed that coherences do not affect decisions in a standard MDP. This section shows that they still matter for exploration. The entropy regularizer depends on the whole matrix, not just its diagonal. All proofs for this section are in Appendix~\ref{app:proofs-entropy}.

The action-value operator at state $s$ is the diagonal matrix $Q_s=\sum_{i=1}^kQ(s,a_i)\eb_i\eb_i^\dagger$, an observable in the sense of Definition~\ref{def:observable}. By Lemma~\ref{lem:phaseirrelevance}, its expectation $V^\rho(s):=\Tr(\rho_sQ_s)$ equals the classical value $\sum_i\pi(a_i\mid s)Q(s,a_i)$ for every $\rho_s\in\mathcal D_k$. This holds regardless of $\rho_s$'s off-diagonal entries. The regularizer introduced next is different, as it will see those entries directly. 

\subsection{Spectral entropy and the Gibbs optimum}

The matrix free-energy objective $F_\tau(\rho)=\Tr(\rho Q_s)+\tau S(\rho)$ (Definition~\ref{def:freeenergy}) and its Gibbs optimum (Theorem~\ref{thm:gibbs}) are stated in Section~\ref{subsec:coherentprior}, where they are used.

We recall three standard facts about the matrix relative entropy of Definition~\ref{def:qre}. The proofs in Appendix~\ref{app:proofs} use them.

\begin{lemma}[Klein's inequality; see \cite{nielsen2010quantum,carlen2010trace,petz2008quantum}]\label{lem:klein}
$D(\rho\,\Vert\,\rho')\ge 0$ for all $\rho,\rho'\in\mathcal D_k$, with equality if and only if $\rho=\rho'$.
\end{lemma}

\begin{lemma}[Joint convexity; Lieb \cite{lieb1973convex}, Lindblad \cite{lindblad1974expectations}]\label{lem:jointconvex}
$(\rho,\rho')\mapsto D(\rho\,\Vert\,\rho')$ is jointly convex. In particular, for fixed $\rho'$, the sublevel set $\{\rho\in\mathcal D_k : D(\rho\,\Vert\,\rho')\le\delta\}$ is convex.
\end{lemma}

\begin{lemma}[Data-processing inequality; Lindblad \cite{lindblad1975completely}]\label{lem:dpi}
For any completely positive trace-preserving (CPTP) map $\Phi$, $D\bigl(\Phi(\rho)\,\Vert\,\Phi(\rho')\bigr)\le D(\rho\,\Vert\,\rho')$.
\end{lemma}

The comparison of the two entropies below uses the Schur--Horn relation, which Section~\ref{sec:background} stated without proof.

\begin{lemma}[Schur--Horn majorization and Schur-concavity of entropy; Schur \cite{schur1923}, Horn \cite{horn1954}, see also \cite{marshall2011inequalities}]\label{lem:schurhorn}
For $\rho\in\mathcal D_k$, the diagonal entries of $\rho$ are a doubly stochastic mixture of its eigenvalues, $\rho_{ii}=\sum_n\lambda_n|v_{n,i}|^2$. The matrix $(|v_{n,i}|^2)_{n,i}$ is doubly stochastic by unitarity, since each row and each column sums to $1$. Shannon entropy is Schur-concave, so $S(\rho)\le S(\diag(\rho))$, with equality if and only if $\rho$ is diagonal.
\end{lemma}

The lemma says that $\diag(\rho)$ is majorized by the spectrum $\boldsymbol{\lambda}$. Intuitively, averaging the eigenvalues against a doubly stochastic matrix can only spread the distribution out. A function is Schur-concave if it never decreases under such spreading, and Shannon entropy has this property. So the step from $\boldsymbol{\lambda}$ to $\diag(\rho)$ can only raise entropy.

What von Neumann entropy $S(\rho)$ adds over its classical counterpart is not the principle but the quantity being regulated. Classical maximum-entropy RL is equally principled about how spread out the action probabilities are. But that is all Shannon entropy can see. $\rho$ carries two distributions: its diagonal (the action probabilities) and its spectrum $\{\lambda_n\}$ (Section~\ref{sec:background}). $S(\rho)$ is the entropy of the spectrum, not of the diagonal. The two coincide only when $\rho$'s eigenbasis coincides with the action basis, that is, when $\rho$ is diagonal. By Lemma~\ref{lem:schurhorn}, $S(\rho)\le S(\diag(\rho))$, with equality exactly when $\rho$ is diagonal. Consequently, at a fixed diagonal, turning on any nonzero coherence strictly lowers $S$. Example~\ref{ex:toy} makes this concrete. At $\phi=\uppi/2$, $\rho$ is diagonal and $S(\rho)=\log2$ matches the diagonal's entropy exactly. At $\phi=0$, the diagonal is unchanged at $(\tfrac12,\tfrac12)$. But $\rho$ becomes rank one with eigenvalues $(1,0)$, so $S(\rho)=0$. The action probabilities look maximally uncertain while $\rho$ has collapsed onto a single mode.

Regularizing with $S(\rho)$ therefore controls a quantity classical entropy cannot express: how many modes the policy keeps active. This is precisely the exploration control this section is about. Separately, the same entropy family generates the matrix relative entropy behind the trust regions of Theorem~\ref{thm:pinching}. So the geometry used there is not a separate import.

\begin{corollary}[Annealing limits]\label{cor:annealing}
Let $\rho_\tau(Q_s)$ be the Gibbs state of Theorem~\ref{thm:gibbs}. Let $\Pi_{\max}$ be the orthogonal projector onto the top eigenspace of $Q_s$ (the top eigenvalue of $Q_s$ need not be simple). (1)~As $\tau\to\infty$, $\rho_\tau(Q_s)\to I/k$. This is maximal exploration. (2)~As $\tau\to0^+$, $\rho_\tau(Q_s)\to\Pi_{\max}/\mathrm{rank}\,\Pi_{\max}$. This is exploitation. (3)~$S(\rho_\tau(Q_s))$ is nondecreasing in $\tau$. Write $\beta=1/\tau$. Since $\beta$ decreases as $\tau$ increases, this is equivalent to $dS/d\beta=-\beta\,\Var_{\rho_\beta}(Q_s)\le0$. Here $\rho_\beta=\rho_\tau(Q_s)$ is the same Gibbs state, written in terms of $\beta$. For any $\rho\in\mathcal D_k$, $\Var_\rho(Q_s)=\Tr(\rho Q_s^2)-\bigl(\Tr(\rho Q_s)\bigr)^2$ is the variance of $Q_s$ under $\rho$.
\end{corollary}

In a standard MDP, $Q_s$ is diagonal, so the Gibbs optimum $\rho_\tau(Q_s)$ is diagonal too. Theorem~\ref{thm:gibbs} and Corollary~\ref{cor:annealing} then reduce exactly to classical maximum-entropy RL \cite{haarnoja2018soft,ziebart2008maximum}, variance identity included. That is not where the benefit lies. The same closed-form optimum and the same exact cooling schedule survive unchanged once $Q_s$ is not diagonal, the case taken up in Section~\ref{sec:beyond}. An ordinary Q-function cannot express that case, since it assigns one number per action label and has no cross-terms between them. The cooling speed is still set by the variance of $Q_s$, not by a manually tuned schedule. So the benefit is not a better annealing rule for the standard setting. It is the same trusted rule, ready to extend to a non-diagonal $Q_s$ with no loss of exactness or tractability.

\subsection{Monotonicity of the matrix softmax map}

The Gibbs state gives the best policy at a fixed temperature. We now show that raising the inverse temperature never lowers value.

\begin{definition}[Matrix softmax map]\label{def:softmaxchannel}
Given state $s$ and inverse temperature $\beta\ge0$, define $\mathcal T_{Q_s}^\beta:\mathcal D_k\to\mathcal D_k$ by
\begin{equation}
\mathcal T_{Q_s}^\beta(\rho)=\frac{e^{\beta Q_s/2}\,\rho\,e^{\beta Q_s/2}}{\Tr\bigl(e^{\beta Q_s/2}\,\rho\,e^{\beta Q_s/2}\bigr)}.
\end{equation}
\end{definition}

\begin{lemma}[Well-definedness]\label{lem:welldef}
$\mathcal T_{Q_s}^\beta$ maps $\mathcal D_k$ into itself: it preserves Hermiticity and positive semidefiniteness. The denominator is strictly positive, and normalization restores unit trace.
\end{lemma}

The map is not linear in $\rho$, so it is not a CPTP map. It is the operator analogue of multiplicative-weights reweighting. If $\rho=\diag(\pi_1,\dots,\pi_k)$, then $\bigl(\mathcal T_{Q_s}^\beta(\rho)\bigr)_{ii}=\pi_i e^{\beta Q(s,a_i)}/\sum_j\pi_j e^{\beta Q(s,a_j)}$, recovering the classical exponentiated-gradient/softmax update.

\begin{theorem}[Monotonic improvement]\label{thm:monotonic}
Fix $\rho_0\in\mathcal D_k$ and let $\tilde\rho_\beta=\mathcal T_{Q_s}^{\beta}(\rho_0)$ for $\beta\ge0$, and $J(\beta)=\Tr(\tilde\rho_\beta Q_s)$. Then
\begin{equation}
\frac{dJ}{d\beta}=\Var_{\tilde\rho_\beta}(Q_s)=\Tr\bigl(\tilde\rho_\beta Q_s^2\bigr)-\bigl(\Tr(\tilde\rho_\beta Q_s)\bigr)^2\;\ge\;0 .
\end{equation}
Hence $J(\beta)$ is nondecreasing in $\beta$, and strictly increasing unless $\supp(\rho_0)$ is contained in a single eigenspace of $Q_s$. As $\beta\to\infty$, $\tilde\rho_\beta\to\Pi_\star\rho_0\Pi_\star/\Tr(\Pi_\star\rho_0)$ and $J(\beta)\to q_\star$. Here $q_\star$ is the largest eigenvalue of $Q_s$ whose eigenspace projector $\Pi_\star$ satisfies $\Pi_\star\rho_0\ne0$.
\end{theorem}

Each infinitesimal increase of $\beta$ raises the value at a rate equal to the current variance of $Q_s$ under the policy. Progress is therefore automatic. It stalls only once the policy's support already lies in a single eigenspace of $Q_s$, that is, once every active direction has the same value.

In a standard MDP, this guarantee is nothing more than the classical softmax anneal. Raising the inverse temperature never lowers value, and the proof is the usual variance identity. The same map, derivative formula, and guarantee still hold once $Q_s$ is not diagonal. A classical softmax update cannot pose that case, since sharpening a probability vector has no cross-terms between actions. So one proof covers both the classical and the coherent case, at no extra cost.

\subsection{Trust-region preservation of classical guarantees}

Monotonic improvement has no built-in step limit. We now bound the step size while keeping the same guarantee.

The matrix analogue of KL-constrained policy optimization \cite{kakade2002approximately,schulman2015trust,schulman2017proximal}, itself closely related to natural-gradient methods \cite{amari1998natural,kakade2001natural}, constrains the matrix relative entropy instead of the classical KL divergence:

\begin{equation}\label{eq:trustregion}
\rho_s^{(t+1)}\;=\;\arg\max_{\rho\in\mathcal D_k}\;\Tr(\rho A_s^{(t)})
\quad\text{s.t.}\quad D(\rho\,\Vert\,\rho_s^{(t)})\le\delta ,
\end{equation}
where $A_s^{(t)}$ is a Hermitian advantage operator estimated at iteration $t$. We call the constraint set $\{\rho\in\mathcal D_k:D(\rho\,\Vert\,\rho_s^{(t)})\le\delta\}$ the trust region at iteration $t$.

\begin{definition}[Face generated by $\rho_s^{(t)}$]\label{def:face}
For $\rho_s^{(t)}\in\mathcal D_k$, write $\mathcal F(\rho_s^{(t)})=\{\rho\in\mathcal D_k:\supp\rho\subseteq\supp\rho_s^{(t)}\}$ for the face of $\mathcal D_k$ generated by $\rho_s^{(t)}$. This is the set of density matrices whose support fits inside $\rho_s^{(t)}$'s support. It is the matrix analogue of restricting the probability simplex to distributions supported on $\rho_s^{(t)}$'s nonzero coordinates.
\end{definition}

\begin{proposition}[Surrogate improvement and tractability]\label{prop:surrogate}
The problem stated in~\eqref{eq:trustregion} is a convex program (maximization of a linear functional over a compact convex set). Because matrix relative entropy is infinite outside $\mathcal F(\rho_s^{(t)})$, the trust region is automatically contained in this face. For every $\delta>0$, its solution satisfies $\Tr(\rho_s^{(t+1)}A_s^{(t)})\ge\Tr(\rho_s^{(t)}A_s^{(t)})$. The strict inequality holds unless $\rho_s^{(t)}$ is already a maximizer of $\Tr(\rho A_s^{(t)})$ over $\mathcal F(\rho_s^{(t)})$.
\end{proposition}

In the one-state (bandit) case, where $J(\rho)=\Tr(\rho A_s^{(t)})$ is the expected reward, the update improves it monotonically on the face over which it is defined. This restriction is not a technicality: for rank-deficient $\rho_s^{(t)}$, it cannot be dropped, as a simple two-action counterexample shows (Remark~\ref{rem:facenecessary}, Appendix~\ref{app:remarks}).

If $\rho_s^{(t)}$ has full rank, however, then $\mathcal F(\rho_s^{(t)})=\mathcal D_k$. Strict inequality then holds whenever $\rho_s^{(t)}$ is not a global maximizer over $\mathcal D_k$. Gibbs states and softmax-map images of full-rank states are full rank (proved in Appendix~\ref{app:proofs-entropy}). So the face restriction does not bind along trajectories built from them. Remark~\ref{rem:facelowrank} in Appendix~\ref{app:remarks} gives the full argument.

Proposition~\ref{prop:surrogate} improves the surrogate $\Tr(\rho A_s^{(t)})$ at a single state. For multi-state MDPs, improving the true return needs more than that. Per-state surrogate improvement only translates into trajectory-level improvement when the policy does not shift too much, as in classical conservative policy iteration (CPI) and TRPO \cite{kakade2002approximately,schulman2015trust}. So multi-state guarantees require controlling how much the induced action distributions shift across states. The next result shows the matrix trust region enforces this classical trust region automatically.

\begin{theorem}[Operator trust region implies classical trust region]\label{thm:pinching}
Let $\pi(\cdot\mid s)$ and $\pi^{(t)}(\cdot\mid s)$ denote the diagonal action distributions induced by $\rho$ and $\rho_s^{(t)}$ at state $s$. If $D(\rho\,\Vert\,\rho_s^{(t)})\le\delta$, then
\begin{equation}
D_{\mathrm{KL}}(\pi(\cdot\mid s)\,\Vert\,\pi^{(t)}(\cdot\mid s))\le\delta
\qquad\text{and}\qquad
\bigl\|\pi-\pi^{(t)}\bigr\|_1\le\sqrt{2\delta}.
\end{equation}
\end{theorem}

The proof applies the data-processing inequality (Lemma~\ref{lem:dpi}) to the pinching map $\Phi(\rho)=\sum_i\eb_i\eb_i^\dagger\rho\,\eb_i\eb_i^\dagger$, which keeps only the diagonal of $\rho$. Pinsker's inequality then gives the $\ell_1$ bound (Appendix~\ref{app:proofs-entropy}).

Consequently, per-state application of~\eqref{eq:trustregion} induces classical policies that satisfy the distance constraints behind the improvement bounds of CPI and TRPO \cite{kakade2002approximately,schulman2015trust}. Those bounds are stated in terms of the distance between the old and new policy. The bound above is on the $\ell_1$ distance, which is twice the total variation (TV). So it gives $\mathrm{TV}\le\sqrt{\delta/2}$, and this controls that distance directly. Total variation is symmetric in the two policies. So the transfer is exact for the total-variation form of these bounds. The KL bound has the direction $D_{\mathrm{KL}}(\pi\Vert\pi^{(t)})$. TRPO as usually stated constrains the reverse direction, so the KL form does not transfer verbatim. Intuitively, coherences never appear on the diagonal, so they cannot themselves move the sampled action distribution. No matter how they shift, the classical distributions $\pi(\cdot\mid s)$ and $\pi^{(t)}(\cdot\mid s)$ can drift apart no faster than $D(\rho\,\Vert\,\rho_s^{(t)})$ allows.

The same argument applies to the penalty of Section~\ref{subsec:coherentprior}. For any reference policy, $D(\rho\Vert\rho_s^{\mathrm{ref}})$ is at least the classical KL divergence between $\diag(\rho)$ and $\diag(\rho_s^{\mathrm{ref}})$. So the penalty constrains the action probabilities, and the rest of it depends on the coherences of $\rho$.

When $\rho_s^{(t)}$ and the optimizer of~\eqref{eq:trustregion} are diagonal, the matrix relative 
entropy reduces to the classical KL divergence, and~\eqref{eq:trustregion} reduces to the 
KL-constrained update underlying TRPO and, in penalized or clipped form, proximal policy optimization (PPO) 
\cite{schulman2015trust,schulman2017proximal}. DMRL thus strictly generalizes these updates. The algorithms of Appendix~\ref{app:algorithms} do not solve~\eqref{eq:trustregion}. DMPG takes gradient steps. \textit{DMPI} (density-matrix policy iteration; Algorithm~\ref{alg:dmpi}) applies the matrix softmax map of Definition~\ref{def:softmaxchannel} with a fixed $\beta$. This subsection therefore describes what a trust-region step guarantees. It does not describe those two algorithms.

\section{Variance reduction: eigenmode averaging and optimal coherence}\label{sec:theory}

This section studies how DMRL changes estimator variance. For a diagonal reward, each mechanism below matches a classical method exactly. Each becomes specific to density matrices only when the reward is non-diagonal. One elementary fact frames everything. The average of $N$ i.i.d.\ copies of an unbiased estimator $\hat J$ of $J$ has mean-squared error (MSE) $\Var(\hat J)/N$, which falls below a target value $\mathrm{MSE}^\star$ as soon as
\begin{equation}\label{eq:mse}
N\;\ge\;\frac{\Var(\hat J)}{\mathrm{MSE}^\star},
\end{equation}
so every reduction in $\Var(\hat J)$ translates directly into fewer required samples. The remaining subsections identify such reductions. Throughout the paper, $N$ rollouts form an evaluation batch and $m$ rollouts form a pilot batch (Corollary~\ref{cor:adaptive}).  Proofs of all stated results appear in Appendix~\ref{app:proofs-variance}.

\subsection{Eigenmode averaging}

Because $\rho$ is an explicit matrix, its eigenvalues $\lambda_n$ come from an eigendecomposition. They are known exactly, not estimated from samples. So there is no need to draw a random eigenmode, which would add extra sampling noise. The mixture over eigenmodes can instead be summed out analytically, using these known $\lambda_n$ directly. This option requires $\rho$ to be an explicit object with a computable spectrum. It is not available to a mixture that exists only as sampled trajectories, since there even the mixture weights must be estimated rather than known exactly. We state the theorem in general, but we read it as a consistency result. For diagonal observables it is classical stratified sampling, as the discussion after the theorem explains.

Compare two ways to estimate $J(\rho)$ by rollout, where $r_\rho=\mathrm{rank}(\rho)$ is the number of eigenmodes. \emph{Mixture sampling} spends one rollout. It draws a random mode index $W\in\{1,\dots,r_\rho\}$ with $\Pr(W=n)=\lambda_n$, runs the rollout for that mode only, and reports its value as the estimate $\hat J_{\mathrm{mix}}=X_W$. \emph{Eigenmode averaging} spends $r_\rho$ rollouts, one per eigenmode, and outputs the eigenmode-averaging (EA) estimate $\hat J_{\mathrm{EA}}=\sum_n\lambda_n X_n$. This is a spectral form of Rao--Blackwellization \cite{casella1996rao}. Given the per-mode estimates $X_1,\dots,X_{r_\rho}$, the conditional expectation of $\hat J_{\mathrm{mix}}$ is exactly $\hat J_{\mathrm{EA}}$, so eigenmode averaging cannot increase variance.

\begin{theorem}[Eigenmode averaging]\label{thm:raoblackwell}
Let $\rho=\sum_{n=1}^{r_\rho}\lambda_n\mathbf{v}_n\mathbf{v}_n^\dagger$ with $r_\rho=\mathrm{rank}(\rho)$. Let $O$ be an observable, and write $\mu_n=\mathbf{v}_n^\dagger O\mathbf{v}_n$. Then $J(\rho)=\Tr(\rho O)=\sum_n\lambda_n\mu_n$. Suppose that for each eigenmode $n$, we have an unbiased rollout estimator $X_n$ of $\mu_n$, with $\Var(X_n)=\sigma_n^2$, and that the $\{X_n\}$ are independent. 
Then both $\hat J_{\mathrm{mix}}$ and $\hat J_{\mathrm{EA}}$ are unbiased, and
\begin{equation}
\Var\bigl(\hat J_{\mathrm{mix}}\bigr)
=\underbrace{\textstyle\sum_n\lambda_n\sigma_n^2}_{\text{within-mode}}
+\underbrace{\textstyle\sum_n\lambda_n\mu_n^2-J^2}_{\text{between-mode }=\ \Var_{\boldsymbol{\lambda}}(\boldsymbol{\mu})}
\;\;\ge\;\;
\Var\bigl(\hat J_{\mathrm{EA}}\bigr)=\sum_n\lambda_n^2\sigma_n^2 .
\end{equation}
Equality holds only in a degenerate case where the between-mode variance is zero, and $\lambda_n(1-\lambda_n)\sigma_n^2=0$ for every $n$.
\end{theorem}

Theorem~\ref{thm:raoblackwell} compares the two estimators as defined, not at equal rollout budgets. Mixture sampling spends one rollout, while eigenmode averaging spends $r_\rho$. The fair comparison averages $r_\rho$ i.i.d.\ mixture samples instead of just one. This averaged estimator has variance $\bigl(\sum_n\lambda_n\sigma_n^2+\Var_{\boldsymbol{\lambda}}(\boldsymbol{\mu})\bigr)/r_\rho$.

At equal budgets, eigenmode averaging is not automatically the better choice. For strongly skewed $\boldsymbol{\lambda}$, one eigenmode carries almost all the mass and has a large $\sigma_n^2$. In that case, the averaged mixture variance can be lower than the variance of eigenmode averaging. Eigenmode averaging wins exactly when $\Var_{\boldsymbol{\lambda}}(\boldsymbol{\mu}) > r_\rho\sum_n\lambda_n^2\sigma_n^2-\sum_n\lambda_n\sigma_n^2$. The right side is zero for uniform $\boldsymbol{\lambda}$, so eigenmode averaging is never worse there. For a skewed spectrum, the right side can be large, and the between-mode variance must exceed it. To check the uniform case, take $\lambda_n=1/r_\rho$ for every $n$. Then eigenmode averaging has variance $\sum_n\lambda_n^2\sigma_n^2=\sigma^2_{\mathrm{avg}}/r_\rho$, where $\sigma^2_{\mathrm{avg}}=\frac1{r_\rho}\sum_n\sigma_n^2$ is the average within-mode variance. The averaged mixture-sampling variance $(\sigma^2_{\mathrm{avg}}+\Var_{\boldsymbol{\lambda}}(\boldsymbol{\mu}))/r_\rho$ is at least as large.

In a standard MDP, $O$ is diagonal, and eigenmode averaging reduces to a familiar classical method. Here $\mu_n=\mathbf{v}_n^\dagger O\mathbf{v}_n=\sum_i|v_{n,i}|^2\,Q(s,a_i)$ depends only on $|v_{n,i}|^2$. So the phases of $\mathbf{v}_n$ never enter $\mu_n$, even when $\rho$ has nonzero coherences. An ``eigenmode rollout'' is then an ordinary action rollout, drawn from the reweighted distribution $\mathbf{p}^{(n)}=(|v_{n,i}|^2)_{i=1}^{k}$, with $X_n$ its return. By Lemma~\ref{lem:schurhorn}, the action distribution of $\rho$ is the mixture $\mathbf{p}=(\rho_{ii})_{i=1}^k=\sum_n\lambda_n\mathbf{p}^{(n)}$. Running one rollout from each $\mathbf{p}^{(n)}$ and weighting by the known $\lambda_n$ is classical stratified sampling. It is equivalent to Rao--Blackwellization against a chosen partition \cite{casella1996rao}. Any split of $\mathbf{p}$ into a mixture gives an unbiased estimator this way, and eigenmode averaging is one particular split. If $\rho$ is diagonal as well, each eigenmode is literally one action. Eigenmode averaging then coincides exactly with the classical expected (all-action) policy gradient \cite{sutton2000policy,ciosek2018expected}. That estimator sums over every action weighted by its known probability instead of sampling one.

So for a diagonal observable, the density matrix is a convenient, rank-controlled way to generate the split. It is not the source of the variance reduction. The reduction depends on the matrix itself only for a non-diagonal observable. Then $\mu_n$ contains cross terms $\overline{v_{n,i}}\,O_{ij}\,v_{n,j}$ with $i\ne j$. These depend on the relative phases or signs of the components of $\mathbf{v}_n$, and no mixture over actions reproduces them.

Eigenmode averaging introduces no bias and needs no extra assumption. Its only price is more rollouts, one per eigenmode instead of one in total. In the low-rank regime, $r_\rho$ is small, so this price is modest. When that condition holds, removing the between-mode noise can mean fewer rollouts are needed to reach a given estimation accuracy. Appendix~\ref{subsec:multiples} quantifies this sample-efficiency gain as $g_{\mathrm{EA}}$.

\subsection{Optimal coherence}

The previous subsection removed the between-mode variance by averaging exactly, though within-mode variance remains. This subsection introduces a second, separate source of variance reduction, built from the coherences themselves.

A control variate is a zero-mean statistic added to an estimator. Adding it cannot bias the answer. If it is correlated with the estimator's noise, an appropriately signed coefficient cancels part of that noise \cite{lavenberg1981perspective,glasserman2004monte}. For a diagonal observable, the off-diagonal statistics $\hat o_{ij}$ have mean zero, so the coherences play exactly this role, and the policy's own off-diagonal entries serve as the coefficients. For a non-diagonal observable, changing the coherences also changes the policy being evaluated. The results below then compare the variance of evaluating two policies with the same action probabilities. We fix the diagonal of $\rho$, the action probabilities. We then isolate the effect of the off-diagonal entries on estimator variance. For clarity, we restrict to real coherences and observables.

For a diagonal observable, the off-diagonal entries multiply zero-mean statistics, so the same estimator can be written without a density matrix. We therefore read the results below as a consistency theorem and a bridge between matrix structure and classical variance reduction. They are not a primary algorithmic advantage.

To begin with, fix a single state $s$ and $p\in(0,1)$. Every quantity below depends on $s$. We drop $s$ from the notation, since only one state is in view. Consider the family of policies indexed by the coherence $c$, which is the off-diagonal entry,
\begin{equation}
\rho_{c}=\begin{pmatrix}p & c\\ c & 1-p\end{pmatrix},
\qquad |c|\le\sqrt{p(1-p)}\ \ \text{(PSD condition)}.
\end{equation}
Every member of this family has the same diagonal. Only the coherence $c$ varies across the family.
Let $O$ be a real symmetric observable. Then $J(\rho_{c})=p\,O_{11}+(1-p)\,O_{22}+2c\,O_{12}$. Note that the complex case requires only replacing $c\,O_{12}$ by $\re(c\,\overline{O_{12}})$.
The entries of $O$ are estimated from shared rollouts. The estimators $\hat o_{11},\hat o_{22},\hat o_{12}$ are jointly distributed and unbiased. Form the plug-in estimator $\hat J(c)=p\,\hat o_{11}+(1-p)\,\hat o_{22}+2c\,\hat o_{12}$. 

\begin{theorem}[Control-variate effect of coherence, $k=2$]\label{thm:controlvariate}
Let $\kappa=\Cov\bigl(p\,\hat o_{11}+(1-p)\,\hat o_{22},\ \hat o_{12}\bigr)$. Let $\sigma_{12}^2=\Var(\hat o_{12})$. Then $\hat J(c)$ is unbiased for $J(\rho_{c})$, and
\begin{equation}
\Var\bigl(\hat J(c)\bigr)=\Var\bigl(\hat J(0)\bigr)+4c\,\kappa+4c^2\sigma_{12}^2 .
\end{equation}
Suppose $\kappa\neq0$. Let $c_{\max}=\min\bigl(|\kappa|/\sigma_{12}^2,\ \sqrt{p(1-p)}\bigr)$. For every $c$ with $0<|c|<c_{\max}$ and $\mathrm{sign}(c)=-\mathrm{sign}(\kappa)$, $\Var(\hat J(c))<\Var(\hat J(0))$. The coherent policy is therefore evaluated with strictly smaller variance than its diagonal counterpart with the same action probabilities. The variance-minimizing coherence is $c^\star=-\kappa/(2\sigma_{12}^2)$, clipped to the PSD interval.
\end{theorem}

The general-$k$ decomposition extends this result to $k$ actions (Appendix~\ref{app:proofs-variance}). For $k=2$, the coherence vector $\mathbf{c}$ below has a single entry, the scalar $c$ above. The variance is a quadratic in the coherences. Its coefficients are known exactly. So the minimizer has a closed form. Choosing the coherence at that minimizer removes the sign condition in Theorem~\ref{thm:controlvariate}. The reduction then holds unconditionally. This is the main variance-reduction identity.

\begin{theorem}[Optimal coherence guarantees variance reduction]\label{thm:optcoherence}

For each pair $i<j$, let $\hat o_{ij}$ denote a jointly distributed, unbiased estimator of $O_{ij}$ from the same rollouts (generalizing $\hat o_{12}$ above), together with $\hat o_{ii}$ for the diagonal entries. Write $\hat J_{\mathrm{diag}}=\sum_i p_i\hat o_{ii}$, with $p_i=\rho_{ii}$, for the diagonal-only plug-in estimator (the $k=2$ case is $\hat J(0)$ in Theorem~\ref{thm:controlvariate}), and $\hat J(\mathbf{c})=\hat J_{\mathrm{diag}}+2\sum_{i<j}c_{ij}\hat o_{ij}$ for the coherence-augmented estimator. Collect the off-diagonal coherences into the vector $\mathbf{c}=(c_{ij})_{i<j}$, ordering pairs $i<j$ consistently throughout. Similarly, let $\boldsymbol{\gamma}=(\gamma_{ij})_{i<j}$ be the vector of covariances of $\hat J_{\mathrm{diag}}$ with each off-diagonal statistic, where $\gamma_{ij}=\Cov(\hat J_{\mathrm{diag}},\hat o_{ij})$. Let $\Gamma=\Cov\bigl((\hat o_{ij})_{i<j}\bigr)\succeq0$ denote the covariance matrix of the random vector $(\hat o_{ij})_{i<j}$ (i.e. $\Gamma_{(ij),(i'j')}=\Cov(\hat o_{ij},\hat o_{i'j'})$), with Moore--Penrose pseudoinverse $\Gamma^{+}$.  Then
\begin{equation}
\Var\bigl(\hat J(\mathbf{c})\bigr)=\Var\bigl(\hat J_{\mathrm{diag}}\bigr)+4\mathbf{c}^\top\boldsymbol{\gamma}+4\mathbf{c}^\top\Gamma \mathbf{c}
=\Var\bigl(\hat J_{\mathrm{diag}}\bigr)-\boldsymbol{\gamma}^\top\Gamma^{+}\boldsymbol{\gamma}+4\bigl(\mathbf{c}-\mathbf{c}^\star\bigr)^\top\Gamma\bigl(\mathbf{c}-\mathbf{c}^\star\bigr),
\end{equation}
where $\mathbf{c}^\star=-\tfrac12\Gamma^{+}\boldsymbol{\gamma}$. The set of minimizers is $\mathbf{c}^\star+\ker\Gamma$ (the null space $\{\mathbf{z}:\Gamma \mathbf{z}=0\}$). Consequently, the unconstrained minimum variance is $\Var(\hat J_{\mathrm{diag}})-\boldsymbol{\gamma}^\top\Gamma^{+}\boldsymbol{\gamma}$. So the reduction $\boldsymbol{\gamma}^\top\Gamma^{+}\boldsymbol{\gamma}$ is a fraction of $\Var(\hat J_{\mathrm{diag}})$. Denote that fraction $\eta^2$, which lies in $[0,1]$. We set $\eta^2=0$ in the trivial case $\Var(\hat J_{\mathrm{diag}})=0$. Formally, $\eta^2$ is the squared multiple correlation between $\hat J_{\mathrm{diag}}$ and the off-diagonal statistics, also known as the coefficient of determination. With this definition,
\begin{equation}\label{eq:r2reduction}
\boldsymbol{\gamma}^\top\Gamma^{+}\boldsymbol{\gamma}\;=\;\eta^2\,\Var\bigl(\hat J_{\mathrm{diag}}\bigr).
\end{equation}
\end{theorem}

$\Gamma^{+}$ inverts $\Gamma$ along the directions in which the statistics actually vary, and it is zero along directions where they are constant. Positive definiteness of $\Gamma$ is \emph{not} assumed, because $\Gamma$ can be singular. The formula for $\mathbf{c}^\star$ still gives a minimizer, since $\boldsymbol{\gamma}\in\mathrm{range}(\Gamma)$ automatically. Remark~\ref{rem:kernel} in Appendix~\ref{app:remarks} explains the role of $\ker\Gamma$.

When $O$ is diagonal, each off-diagonal statistic $\hat o_{ij}$ has mean zero. But it still varies across rollouts, and can still correlate with $\hat J_{\mathrm{diag}}$. That correlation is exactly what classical control variates exploit. So the theorem coincides with the optimally weighted classical control variate, and $\eta^2$ can be strictly positive even though $O$ is diagonal. What is specific to the density-matrix formalism shows up only when $O$ has off-diagonal entries. In general, though, $\eta^2$ in~\eqref{eq:r2reduction} has its familiar regression meaning. It is the fraction of the diagonal estimator's variance explained by the best linear predictor from the off-diagonal statistics.

An explicit construction of jointly distributed unbiased estimators $\hat o_{ij}$ satisfying Theorem~\ref{thm:optcoherence} is given in Appendix~\ref{app:constructions}. Example~\ref{ex:constructionA} covers standard MDPs, where the observable is diagonal. Indicator-based estimators are unbiased there, but become linearly dependent once $k$ grows: $\Gamma$ is exactly singular for every $k\ge3$.

Theorem~\ref{thm:optcoherence} treats $\mathbf{c}$ as unconstrained. To state the constraint, we first fix some notation. Let $\rho_{\mathrm{diag}}=\diag(p_1,\dots,p_k)$ be the diagonal part of $\rho$, where $p_i=\rho_{ii}$. For a coherence vector $\mathbf{c}=(c_{ij})_{i<j}$, let $C$ be the real symmetric matrix with $C_{ij}=C_{ji}=c_{ij}$ and zero diagonal. Then $\rho_{\mathbf{c}}=\rho_{\mathrm{diag}}+C$ has the same action probabilities as $\rho$ and coherences $\mathbf{c}$. This matrix must be PSD, so $\mathbf{c}^\star$ may not give a valid policy. The next result bounds how much this constraint can cost.

\begin{corollary}[Feasibility of the optimal coherence]\label{cor:feasibility}
Suppose $\rho_{ii}>0$ for every $i$. Here $\rho_{\alpha \mathbf{c}^\star}$ keeps the diagonal of $\rho$ fixed and scales the coherences $\mathbf{c}^\star$ by $\alpha$. The PSD cone is convex, so the set of $\alpha\ge0$ for which $\rho_{\alpha \mathbf{c}^\star}$ is PSD is an interval $[0,\alpha_{\max}]$. Its right endpoint $\alpha_{\max}$ is the largest feasible scaling. It is positive. It is finite whenever $\mathbf{c}^\star\ne0$, as the bound after this corollary shows. It is infinite only in the trivial case $\mathbf{c}^\star=0$. If $\alpha_{\max}\ge1$, the whole segment from $\rho_{\mathrm{diag}}$ ($\alpha=0$) to $\rho_{\mathbf{c}^\star}$ ($\alpha=1$, the
unconstrained optimum of Theorem~\ref{thm:optcoherence}) is feasible. If $\alpha_{\max}<1$, only the part up to $\alpha_{\max}$ is feasible. Along the feasible part, using
$\boldsymbol{\gamma}^\top\Gamma^{+}\boldsymbol{\gamma}=\eta^2\Var(\hat J_{\mathrm{diag}})$,
\begin{equation}
\Var\bigl(\hat J(\alpha \mathbf{c}^\star)\bigr)=\Var\bigl(\hat J_{\mathrm{diag}}\bigr)\bigl[1-\alpha(2-\alpha)\eta^2\bigr].
\end{equation}
The factor $\alpha(2-\alpha)$ increases on $[0,1]$ and decreases beyond. So its maximum on
$[0,\alpha_{\max}]$ is at $\alpha=\min(\alpha_{\max},1)$. When $\alpha_{\max}\ge1$, this gives the
full reduction $\eta^2$. When $\alpha_{\max}<1$, it gives the smaller guaranteed reduction
$\alpha_{\max}(2-\alpha_{\max})\eta^2$.
\end{corollary}

Positivity of every $\rho_{ii}$ matters for a concrete reason. Take any pair $i\ne j$. The $2\times2$ principal submatrix of $\rho_{\alpha \mathbf{c}^\star}$ on $\{i,j\}$ has determinant $\rho_{ii}\rho_{jj}-\alpha^2|c_{ij}^\star|^2$. Every principal submatrix of a PSD matrix is itself PSD, so this determinant cannot be negative. Hence
\begin{equation}
\alpha_{\max}\;\le\;\frac{\sqrt{\rho_{ii}\rho_{jj}}}{|c_{ij}^\star|}\qquad\text{whenever }c_{ij}^\star\ne0.
\end{equation}
If $\rho_{ii}=0$, the bound is zero, so the pair $(i,j)$ alone forces $\alpha_{\max}=0$. More generally, $\rho_{jj}\le1$ gives $\alpha_{\max}\le\sqrt{\rho_{ii}}/|c_{ij}^\star|$. So if $c_{ij}^\star\ne0$ for some $j$, and $\mathbf{c}^\star$ is held fixed, then $\alpha_{\max}$ shrinks at least as fast as $\sqrt{\rho_{ii}}$ when $\rho_{ii}\to0$.

The choice $\alpha=\min(\alpha_{\max},1)$ is only guaranteed optimal along the line $\{\alpha \mathbf{c}^\star\}$.
The exact minimizer of $\Var(\hat J(\mathbf{c}))$ over the full spectrahedron $\{\mathbf{c}:\rho_{\mathbf{c}}\succeq0\}$ always
attains a variance at least as small. When $\alpha_{\max}\ge1$, the two already coincide, since
$\alpha=1$ is the unconstrained global optimum. When $\alpha_{\max}<1$, the full spectrahedron
minimizer can do strictly better. This minimization is a convex program, but we do not solve it in closed form (see Appendix~\ref{subsec:practice}).

\subsection{Using the optimal coherence in practice}\label{subsec:practice}

The feasibility analysis above matters only when the coherences are embedded as off-diagonal entries of $\rho$ itself. This is where $\rho$ must stay PSD. In a standard MDP, the positivity assumption on $\rho_{ii}$ routinely fails as training proceeds. Corollary~\ref{cor:feasibility} needs $\rho_{ii}>0$ for every action. Annealing toward a sharper policy, greedy exploitation, and masked actions all push some $\rho_{ii}$ toward zero as training converges. Exactly there, $\alpha_{\max}$ shrinks toward zero too, by the bound after Corollary~\ref{cor:feasibility}. The guaranteed reduction along the segment $\{\alpha \mathbf{c}^\star\}$ collapses with it.

But for diagonal $O$, embedding $\mathbf{c}^\star$ inside $\rho$ was never required in the first place. The coherences carry no decision-relevant content in this case (Lemma~\ref{lem:phaseirrelevance}). The derivative of the return with respect to $\rho$ is diagonal, so it gives them no gradient (Theorem~\ref{thm:gradient}). So instead of storing $\mathbf{c}^\star$ as an entry of $\rho$, we can keep $\rho$ diagonal and place the coefficient in a separate classical control-variate term, one that never touches $\rho$. This term carries no PSD constraint. It always attains the full reduction $\eta^2$, no matter how sharp the policy becomes.
This is the choice we adopt from here on for diagonal-observable MDPs. Coherences are not embedded in $\rho$ at all. Corollary~\ref{cor:feasibility} then does not need to be checked, because no coherences are stored in $\rho$ for the PSD constraint to restrict.

The situation is different when $O$ is non-diagonal, as in the coupled-reward case of Section~\ref{subsec:coupledreward}. There the coherences shape the objective $\Tr(\rho O)$ itself (Theorem~\ref{thm:phaserelevance}). They must be embedded in $\rho$, so the feasibility constraint applies in full and $\rho_{\mathbf{c}}$ must stay PSD. Corollary~\ref{cor:feasibility} then bounds the largest scalar $\alpha_{\max}$ for which $\alpha \mathbf{c}^\star$ stays feasible.

The scaling $\alpha_{\max}$ is easy to compute. It is the reciprocal of a single extremal eigenvalue, and it depends only on $\rho_{\mathrm{diag}}$ and $\mathbf{c}$, not on individual samples. So it is computed once each time $\hat{\mathbf{c}}$ is refit from a fresh pilot batch (Corollary~\ref{cor:adaptive}), not on every evaluation rollout. Remark~\ref{rem:alphamax} in Appendix~\ref{app:remarks} gives the formula and the cost.

The feasibility constraint is not an artifact of building $\rho$ entry by entry. It holds under any parameterization, because every $2\times2$ principal minor of $\rho$ must be nonnegative. Methods that never set a coherence explicitly avoid it by construction. Two examples are DMPG, which takes gradient steps on the factor $U$ of Definition~\ref{def:factorparam}, and DMPI, which applies the matrix softmax map of Definition~\ref{def:softmaxchannel}. Remark~\ref{rem:param} in Appendix~\ref{app:remarks} explains why.

When $O$ is non-diagonal, embedding the coherences in $\rho$ also creates a trade-off. The coherences that maximize $\Tr(\rho O)$ are generally not the ones that minimize variance, yet both live in the same entries of $\rho$. Moving $\mathbf{c}$ toward the variance-optimal $\mathbf{c}^\star$ pulls it away from the return-optimal choice, and vice versa. The return is linear in $\mathbf{c}$ and the variance is an exact quadratic in $\mathbf{c}$ (Theorem~\ref{thm:optcoherence}). So balancing the two is a convex program, and sweeping the penalty weight traces out a Pareto frontier between return and estimator variance. Remark~\ref{rem:pareto} in Appendix~\ref{app:remarks} states the program and discusses how to solve it. The same remark explains why the variance-only minimization left open above is also a convex program.

Theorem~\ref{thm:optcoherence} assumes that $\boldsymbol{\gamma}$ and $\Gamma$ are known exactly. In practice, they must be estimated. The next result shows the guarantee still holds, up to a penalty that vanishes with more data. It is stated for the case where the coherences are embedded in $\rho$, so a scaling step enforces the PSD constraint. For the separate control-variate term used with diagonal observables, no scaling is needed. Then $\hat\alpha=1$ and $\mathbf{c}_{\mathrm{feas}}=\mathbf{c}^\star$. The hypothesis $\rho_{ii}>0$ is needed only by the scaling step.

\begin{corollary}[Adaptive coherence attains the guarantee from data]\label{cor:adaptive}
Suppose $\rho_{ii}>0$ for every $i$. Suppose $\boldsymbol{\gamma}$ and $\Gamma$ are estimated by sample covariances $\hat{\boldsymbol{\gamma}},\hat\Gamma$ from an independent pilot batch of $m$ rollouts, and that every rollout statistic has a finite fourth moment. Set $\hat{\mathbf{c}}=-\tfrac12(\hat\Gamma+\xi I)^{-1}\hat{\boldsymbol{\gamma}}$, using a ridge term $\xi>0$. The ridge is needed in general because $\Gamma$ may be singular, as in Example~\ref{ex:constructionA}. Let $\tilde{\mathbf{c}}=\hat\alpha\,\hat{\mathbf{c}}$, where $\hat\alpha=\min(\alpha_{\max}(\hat{\mathbf{c}}),1)$ is the largest scaling in $[0,1]$ that keeps $\rho_{\tilde{\mathbf{c}}}$ PSD (Corollary~\ref{cor:feasibility}). Conditional on the pilot batch, $\hat J(\tilde{\mathbf{c}})$ is unbiased for $J(\rho_{\tilde{\mathbf{c}}})$, and
\begin{equation}\label{eq:condvar}
\Var\bigl(\hat J(\tilde{\mathbf{c}})\mid \tilde{\mathbf{c}}\bigr)
=\Var\bigl(\hat J_{\mathrm{diag}}\bigr)-\boldsymbol{\gamma}^\top\Gamma^{+}\boldsymbol{\gamma}
+4\bigl(\tilde{\mathbf{c}}-\mathbf{c}^\star\bigr)^\top\Gamma\bigl(\tilde{\mathbf{c}}-\mathbf{c}^\star\bigr).
\end{equation}
Let $\alpha_{\mathrm{feas}}=\min(\alpha_{\max}(\mathbf{c}^\star),1)$ and $\mathbf{c}_{\mathrm{feas}}=\alpha_{\mathrm{feas}} \mathbf{c}^\star$ be the feasible target of Corollary~\ref{cor:feasibility}. Choose the ridge $\xi=\xi_m$ so that $\xi_m\to0$ and $\xi_m\sqrt m\to\infty$. Then $\tilde{\mathbf{c}}\to \mathbf{c}_{\mathrm{feas}}$ in mean square as $m\to\infty$, and
\begin{equation}\label{eq:adaptivelimit}
\Var\bigl(\hat J(\tilde{\mathbf{c}})\mid\tilde{\mathbf{c}}\bigr)\ \longrightarrow\ \Var\bigl(\hat J_{\mathrm{diag}}\bigr)\bigl[1-\alpha_{\mathrm{feas}}(2-\alpha_{\mathrm{feas}})\eta^2\bigr]
\end{equation}
in mean. If $\alpha_{\max}(\mathbf{c}^\star)\ge1$, then $\alpha_{\mathrm{feas}}=1$ and the limit is the full reduction by $\eta^2$. For a fixed ridge $\xi>0$, the pilot estimate satisfies $\E\|\hat{\mathbf{c}}-\mathbf{c}_\xi\|^2=\mathcal O(1/m)$, where $\mathbf{c}_\xi=-\tfrac12(\Gamma+\xi I)^{-1}\boldsymbol{\gamma}$ is the ridge target. The ridge itself costs at most $\xi^2\boldsymbol{\gamma}^\top(\Gamma^{+})^3\boldsymbol{\gamma}$ in the excess term of~\eqref{eq:condvar}. Without a ridge, on a linearly independent subset of the $\hat o_{ij}$ where $\Gamma\succ0$ and with no scaling needed, the excess term, multiplied by $m$, stays bounded in probability as $m\to\infty$. We make no claim about its expectation in that case. The adaptive procedure therefore attains the guaranteed reduction up to a vanishing estimation penalty. This is the standard behavior of estimated control variates \cite{casella1996rao,lavenberg1981perspective,glasserman2004monte}.
\end{corollary}

The finite fourth moment is a mild condition. It holds whenever per-step rewards are bounded. The variance in the corollary is conditional on the pilot batch, which fixes $\tilde{\mathbf{c}}$. This is the variance seen by the $N$ evaluation rollouts, so the gain factors of Appendix~\ref{subsec:multiples} are built on it. For diagonal $O$, the total variance equals the optimum $\Var(\hat J_{\mathrm{diag}})-\boldsymbol{\gamma}^\top\Gamma^{+}\boldsymbol{\gamma}$ plus the expected estimation penalty $4\,\E[(\tilde{\mathbf{c}}-\mathbf{c}^\star)^\top\Gamma(\tilde{\mathbf{c}}-\mathbf{c}^\star)]$. For non-diagonal $O$, the total variance also includes the variability of the target $J(\rho_{\tilde{\mathbf{c}}})$ across pilot batches. Remark~\ref{rem:adaptive} in Appendix~\ref{app:remarks} derives both statements.

The $m$ pilot rollouts play a different role from the rollouts averaged in~\eqref{eq:mse}. They estimate the covariance objects $\boldsymbol{\gamma}$ and $\Gamma$, not the target $J$. Once $\hat{\mathbf{c}}$ is fixed, a separate evaluation batch of size $N$ realizes the reduced variance, which Appendix~\ref{subsec:multiples} converts into the gain factor $1/(1-\eta^2)$. That factor does not include the pilot cost. For a total budget of $m+N$ rollouts, the pilot is a one-time overhead amortized over $N$. The pilot cost scales with the dimension of $\Gamma$, not with $N$. Remark~\ref{rem:adaptive} gives that dimension.

The resulting procedure, \textit{ACPE} (adaptive-coherence policy evaluation; Algorithm~\ref{alg:acpe} of Appendix~\ref{app:algorithms}), rolls out a pilot batch, forms $\hat{\boldsymbol{\gamma}},\hat\Gamma$, sets $\hat{\mathbf{c}}$, and then evaluates on a fresh batch. The guaranteed reduction is relative to the statistics chosen. A poor choice, with $\boldsymbol{\gamma}=0$, yields zero reduction, 
but never an increase. Classical control variates have the same property: an uncorrelated covariate contributes nothing, but never hurts \cite{glasserman2004monte}. Picking good covariates is therefore a feature-design problem the matrix formulation inherits, not one it creates.

\subsection{From variance reduction to sample-efficiency gain factors}\label{subsec:multiples}

The results proved above are variance reductions. A variance ratio also says how many times fewer samples are needed for the same accuracy. So every ratio proved above is a sample-efficiency gain factor for its estimation problem. We list these factors here so that the paper's quantitative predictions are easy to state.

\textit{The eigenmode-averaging (EA) gain factor.} Theorem~\ref{thm:raoblackwell} supplies a sample-efficiency gain factor with a closed form. At matched rollout budgets, $r_\rho$ rollouts on each side, comparing the two estimators gives
\begin{equation}
g_{\mathrm{EA}} \;=\; \frac{1}{r_\rho}\cdot\frac{\sum_n\lambda_n\sigma_n^2+\Var_{\boldsymbol{\lambda}}(\boldsymbol{\mu})}{\sum_n\lambda_n^2\sigma_n^2}.
\end{equation}
As discussed after Theorem~\ref{thm:raoblackwell}, a uniform spectrum gives $g_{\mathrm{EA}}=1+\Var_{\boldsymbol{\lambda}}(\boldsymbol{\mu})/\sigma^2_{\mathrm{avg}}$, which is never below one. A skewed spectrum can push $g_{\mathrm{EA}}$ below one. A simple case has equal within-mode variances and zero between-mode variance. Then $g_{\mathrm{EA}}=1/\bigl(r_\rho\Tr(\rho^2)\bigr)$, where $\Tr(\rho^2)$ is the purity of $\rho$. This is below one for every non-uniform spectrum, and it falls further as the spectrum becomes more skewed. In practice, only $\boldsymbol{\lambda}$ is known exactly. The $\sigma_n^2$ and $\mu_n$ must be estimated from rollouts, so a check of whether $g_{\mathrm{EA}}>1$ is only as reliable as those estimates. Given them, the closed form says whether eigenmode averaging pays off, and by how much.

We state the relevant caveat here rather than let the closed form imply more than it does. For diagonal $O$, $g_{\mathrm{EA}}$ is realizable without any density matrix. It is exactly the gain from classical stratified sampling against any $r_\rho$-component mixture decomposition of the action distribution (the discussion following Theorem~\ref{thm:raoblackwell}). The density matrix supplies a differentiable, rank-controlled way to generate such a decomposition, not a mechanism unavailable to classical estimators. The gain factor is density-matrix-specific only once $O$ is non-diagonal, as in the coupled-reward case of Section~\ref{subsec:coupledreward}.

\textit{The control-variate (CV) gain factor.} Theorem~\ref{thm:optcoherence} gives a second, separate reduction identity, \eqref{eq:r2reduction}. It converts directly into a sample-efficiency gain factor for policy evaluation:
\begin{equation}
g_{\mathrm{CV}}\;=\;\frac{\Var\bigl(\hat J_{\mathrm{diag}}\bigr)}{\Var\bigl(\hat J(\mathbf{c}^\star)\bigr)}\;=\;\frac{1}{1-\eta^2},
\end{equation}
up to the vanishing estimation penalty of Corollary~\ref{cor:adaptive}. In round numbers: $\eta^2=0.5$ doubles sample efficiency, $\eta^2=0.8$ gives a fivefold gain, and $\eta^2=0.9$ gives a tenfold gain. $\eta^2=0$ gives exactly $1\times$. So the gain factor is never below one. If $\mathbf{c}^\star$ must be scaled down to keep $\rho$ PSD, Corollary~\ref{cor:feasibility} gives the exact shrunk factor instead. At $\alpha=\min(\alpha_{\max},1)$, this is $g_{\mathrm{CV}}(\alpha)=1/[1-\alpha(2-\alpha)\eta^2]$.

The choice of baseline matters, and we state it plainly rather than let a favorable baseline inflate the claim. Against a vanilla estimator with no control variates, $g_{\mathrm{CV}}$ is a real gain. Against a strong classical baseline that already uses optimally tuned control variates, the picture is different for diagonal observables: the coherence mechanism coincides exactly with the classical optimum, so the predicted gain over that baseline is $1$, meaning no improvement. We treat this as the correct and expected outcome. A formalism that claimed to beat a provably optimal classical method in the regime where the two objects coincide would be making an error somewhere. The edge over classical practice appears only when the observable is non-diagonal (Theorem~\ref{thm:phaserelevance}). Only then can two policies with the same action probabilities earn different rewards, and only coherences tell them apart.

\section{Deferred proofs}\label{app:proofs}

\subsection{Proofs for Section~\ref{sec:foundations}}\label{app:proofs-foundations}

\begin{proof}[Proof of the first part of Proposition~\ref{prop:inclusion}]
Given $\pi(\cdot\mid s)\in\Delta_k$, define $\rho_s=\sum_{i=1}^k \pi(a_i\mid s)\eb_i\eb_i^\dagger$. In matrix form, $\rho_s$ is diagonal with entries $\pi(a_1\mid s),\dots,\pi(a_k\mid s)$. It is Hermitian because its entries are real. It is positive semidefinite because a diagonal matrix is PSD exactly when its diagonal entries are nonnegative, and each $\pi(a_i\mid s)\ge0$. Its trace is $\sum_i\pi(a_i\mid s)=1$ since $\pi(\cdot\mid s)\in\Delta_k$. So $\rho_s\in\mathcal D_k$. Taking the diagonal recovers the original probabilities. Indeed, since $\{\eb_i\}$ is orthonormal, $\Tr(\rho_s\eb_i\eb_i^\dagger)=\eb_i^\dagger\rho_s\eb_i=\pi(a_i\mid s)$.

For the converse, let $\Lambda$ be any density matrix diagonal in the action basis, with diagonal entries $\Lambda_{11},\dots,\Lambda_{kk}$. Positive semidefiniteness forces $\Lambda_{ii}\ge0$ for each $i$, and unit trace forces $\sum_i\Lambda_{ii}=1$. So $(\Lambda_{11},\dots,\Lambda_{kk})\in\Delta_k$. This assignment is exactly the inverse of the map $\pi\mapsto\rho_s$ above. So the map is a bijection between $\Delta_k$ and the diagonal density matrices.

Strictness for $k\ge2$ holds because non-diagonal density matrices exist. For example, $\rho=\tfrac12(\eb_1+\eb_2)(\eb_1+\eb_2)^\dagger$ has off-diagonal entries $\rho_{12}=\rho_{21}=\tfrac12\ne0$. It is Hermitian and PSD, since it is a rank-one projector scaled to unit trace. So it lies in $\mathcal D_k$ but not among the diagonal matrices. Hence $\Delta_k$ embeds as a strict subset of $\mathcal D_k$.
\end{proof}

\begin{proof}[Proof of the second part of Proposition~\ref{prop:inclusion}]
A Hermitian $k\times k$ matrix is determined by its $k$ real diagonal entries together with its $k(k-1)/2$ independent off-diagonal entries above the diagonal. Each off-diagonal entry is a complex number, equivalently two real numbers, since the entries below the diagonal are fixed as the complex conjugates of these. This gives $k+2\cdot\frac{k(k-1)}{2}=k^2$ real degrees of freedom for a general Hermitian matrix.

The trace-one constraint $\Tr(\rho)=1$ is a single real linear equation on these $k^2$ coordinates. It removes exactly one degree of freedom, leaving an affine subspace of real dimension $k^2-1$.

Positive semidefiniteness does not remove any further dimension. The matrix $I/k$ lies in the interior of the PSD cone, since all its eigenvalues equal $1/k>0$, and small Hermitian perturbations of a positive definite matrix stay positive definite. So the trace-one hyperplane meets the PSD cone in a set with nonempty interior relative to that hyperplane, and this intersection has the same dimension, $k^2-1$, as the hyperplane itself. Hence $\dim\mathcal D_k = k^2-1$.

Convexity: if $\rho_1,\rho_2\in\mathcal D_k$ and $t\in[0,1]$, then $t\rho_1+(1-t)\rho_2$ is Hermitian, being a real combination of Hermitian matrices. It is PSD, since the PSD cone is convex: for any vector $\mathbf{v}$, $\mathbf{v}^\dagger(t\rho_1+(1-t)\rho_2)\mathbf{v}=t\,\mathbf{v}^\dagger\rho_1\mathbf{v}+(1-t)\,\mathbf{v}^\dagger\rho_2\mathbf{v}\ge0$. It has trace $t\Tr(\rho_1)+(1-t)\Tr(\rho_2)=t+(1-t)=1$. So $t\rho_1+(1-t)\rho_2\in\mathcal D_k$.

Compactness: $\mathcal D_k$ is closed, as an intersection of the closed PSD cone with the closed trace-one hyperplane. It is also bounded. For $\rho\succeq0$ with eigenvalues $\lambda_1,\dots,\lambda_k\ge0$ summing to $1$, the Frobenius norm satisfies $\|\rho\|_F^2=\sum_n\lambda_n^2\le\bigl(\sum_n\lambda_n\bigr)^2=1$, so $\|\rho\|_F\le1$ for every $\rho\in\mathcal D_k$. A closed and bounded subset of a finite-dimensional space is compact.

The simplex statement is the first part of the proposition.
\end{proof}

\begin{proof}[Proof of Lemma~\ref{lem:phaseirrelevance}]
$\Tr(\rho O)=\sum_{i,j}\rho_{ij}O_{ji}=\sum_i\rho_{ii}O_{ii}$, since $O_{ji}=0$ for $i\ne j$. The right-hand side involves only the diagonal of $\rho$, so off-diagonal entries and their phases have no effect.
\end{proof}

\begin{proof}[Proof of Theorem~\ref{thm:phaserelevance}]
Fix $i\ne j$ with $O_{ij}\ne0$ and, for $c_{ij}\in(0,1/k]$ and $\phi\in[0,2\uppi)$, define
\[
\rho(\phi)\;=\;\frac{I}{k}\;+\;c_{ij}\Bigl(e^{\mathrm{i}\phi}\eb_i\eb_j^\dagger+e^{-\mathrm{i}\phi}\eb_j\eb_i^\dagger\Bigr).
\]

\emph{Hermiticity and trace.} The added term is Hermitian, since swapping $i$ and $j$ sends $e^{\mathrm{i}\phi}\eb_i\eb_j^\dagger$ to $e^{-\mathrm{i}\phi}\eb_j\eb_i^\dagger$, the other summand, so the term equals its own conjugate transpose. Since $I/k$ is also Hermitian, $\rho(\phi)$ is Hermitian. Every diagonal entry of the added term is zero. The matrix $\eb_i\eb_j^\dagger$ has a single nonzero entry at position $(i,j)$, and $\eb_j\eb_i^\dagger$ has one at position $(j,i)$. So $\Tr\rho(\phi)=\Tr(I/k)=1$.

\emph{Spectrum.} Away from the two-dimensional subspace $\mathrm{span}\{\eb_i,\eb_j\}$, $\rho(\phi)$ acts as $I/k$, giving the eigenvalue $1/k$ with multiplicity $k-2$. On $\mathrm{span}\{\eb_i,\eb_j\}$, $\rho(\phi)$ restricts to the $2\times2$ block $\begin{pmatrix}1/k & c_{ij} e^{\mathrm{i}\phi}\\ c_{ij} e^{-\mathrm{i}\phi}&1/k\end{pmatrix}$. This is a standard computation: for a matrix with equal diagonal entries $1/k$ and off-diagonal entries of magnitude $c_{ij}$, the off-diagonal terms shift the two eigenvalues symmetrically around $1/k$ by exactly $\pm c_{ij}$, giving eigenvalues $1/k\pm c_{ij}$. So the full spectrum of $\rho(\phi)$ is $\{1/k+c_{ij},\,1/k-c_{ij},\,1/k,\dots,1/k\}$. All eigenvalues are nonnegative exactly when $c_{ij}\le1/k$, so $\rho(\phi)\succeq0$ under the stated range of $c_{ij}$. Being Hermitian, PSD, and unit trace, $\rho(\phi)\in\mathcal D_k$ for every $\phi$.

\emph{Shared diagonal.} Every diagonal entry of the added term is zero, so every $\rho(\phi)$ has the same diagonal, $(1/k,\dots,1/k)$, regardless of $\phi$.

\emph{The expectation varies with $\phi$.} Using the trace identity $\Tr(\mathbf{u}\mathbf{v}^\dagger O)=\mathbf{v}^\dagger O\mathbf{u}$, which follows from the cyclic property of the trace,
\[
\Tr\bigl(\rho(\phi)O\bigr)=\Tr\Bigl(\frac{O}{k}\Bigr)+c_{ij} e^{\mathrm{i}\phi}\eb_j^\dagger O\eb_i+c_{ij} e^{-\mathrm{i}\phi}\eb_i^\dagger O\eb_j
=\frac{\Tr O}{k}+c_{ij} e^{\mathrm{i}\phi}O_{ji}+c_{ij} e^{-\mathrm{i}\phi}O_{ij}.
\]
Since $O$ is Hermitian, $O_{ji}=\overline{O_{ij}}$. Writing $O_{ij}=|O_{ij}|e^{\mathrm{i}\arg O_{ij}}$, so that $O_{ji}=|O_{ij}|e^{-\mathrm{i}\arg O_{ij}}$, the last two terms become
\[
c_{ij}|O_{ij}|e^{\mathrm{i}(\phi-\arg O_{ij})}+c_{ij}|O_{ij}|e^{-\mathrm{i}(\phi-\arg O_{ij})}=2c_{ij}|O_{ij}|\cos(\phi-\arg O_{ij}) .
\]
So
\[
\Tr\bigl(\rho(\phi)O\bigr)=\frac{\Tr O}{k}+2c_{ij}\,\lvert O_{ij}\rvert\cos\bigl(\phi-\arg O_{ij}\bigr) ,
\]
which, as $\phi$ ranges over $[0,2\uppi)$, sweeps every value in an interval centered at $\Tr(O)/k$ of width $4c_{ij}|O_{ij}|>0$.
\end{proof}

\begin{proof}[Proof of Lemma~\ref{lem:factorwelldef}]
Let $U\in\C^{k\times r}$ be nonzero and set $\Psi=UU^\dagger$. The matrix $\Psi$ is Hermitian, since $(UU^\dagger)^\dagger=UU^\dagger$. It is positive semidefinite, since $\mathbf{v}^\dagger\Psi \mathbf{v}=\|U^\dagger \mathbf{v}\|^2\ge0$ for every $\mathbf{v}\in\C^k$. Its trace is $\Tr(UU^\dagger)=\|U\|_F^2$, which is strictly positive because $U\ne0$. Dividing by this positive scalar keeps $\Psi$ Hermitian and PSD, and makes the trace equal to $1$. So $\rho(U)=\Psi/\Tr(\Psi)\in\mathcal D_k$. Finally, $\mathrm{rank}(UU^\dagger)\le\mathrm{rank}(U)\le\min(k,r)=r$, and scaling by a positive number does not change the rank. So $\mathrm{rank}\,\rho(U)\le r$.
\end{proof}

\begin{proof}[Proof of Theorem~\ref{thm:gradient}]
Write $f=\Tr(\Psi O)$, so $h=\Tr(\Psi)$ and $J(U)=f/h$. For a Hermitian perturbation $d\Psi$ of $\Psi$, linearity of the trace gives $df=\Tr(d\Psi\,O)$ and $dh=\Tr(d\Psi)$. By the quotient rule,
\[
dJ=\frac{df}{h}-\frac{f\,dh}{h^2}=\frac{df}{h}-J(U)\frac{dh}{h}=\frac{\Tr(d\Psi\,O)-J(U)\Tr(d\Psi)}{h}=\frac{\Tr\bigl(d\Psi\,(O-J(U) I)\bigr)}{h} ,
\]
using $f/h=J(U)$ and linearity of the trace to combine the two terms into one.

Since $\Psi=UU^\dagger$, the perturbation is $d\Psi=(dU)U^\dagger+U(dU)^\dagger$. Substituting, and writing $B=(O-J(U)I)/h$, which is Hermitian since $O$ is Hermitian and $J(U)$ is real,
\[
dJ=\Tr\bigl((dU)U^\dagger B\bigr)+\Tr\bigl(U(dU)^\dagger B\bigr) .
\]
The second term is the complex conjugate of the first: by cyclicity,
\[
\Tr\bigl(U(dU)^\dagger B\bigr)=\Tr\bigl((dU)^\dagger BU\bigr) ,
\]
and taking the conjugate transpose of the first term's argument, using $B=B^\dagger$,
\[
\bigl((dU)U^\dagger B\bigr)^\dagger=B^\dagger U(dU)^\dagger=BU(dU)^\dagger ,
\]
whose trace is $\Tr(BU(dU)^\dagger)=\Tr((dU)^\dagger BU)$ by cyclicity again, matching the second term. So $\Tr(U(dU)^\dagger B)=\overline{\Tr((dU)U^\dagger B)}$, since $\overline{\Tr(Y_1)}=\Tr(Y_1^\dagger)$ for any matrix $Y_1$. Hence
\[
dJ=\Tr\bigl((dU)\,U^\dagger B\bigr)+\overline{\Tr\bigl((dU)\,U^\dagger B\bigr)}
=2\,\re\Tr\bigl((dU)\,U^\dagger B\bigr) .
\]
Under the real inner product $\langle Y_1,Y_2\rangle=\re\Tr(Y_1^\dagger Y_2)$ on $\C^{k\times r}$, the change becomes
\[
dJ=2\,\re\Tr\bigl((dU)U^\dagger B\bigr)=2\langle BU,dU\rangle .
\]
This identifies two objects. The first is the Wirtinger gradient $\nabla_{\overline{U}}J=BU=\tfrac1h(O-J(U)I)U$, which is the convention where $U$ and $\overline{U}$ are treated as independent. The second is the steepest-ascent direction in the real inner product above, which is $2BU$ and so is twice the Wirtinger gradient.

Now let $U$ be real and $O$ be real symmetric. Then $B$ is real symmetric, $dU$ is real, and the same computation gives $dJ=2\Tr\bigl((dU)^\top BU\bigr)$. The ordinary real gradient is therefore $2BU=\tfrac{2}{h}(O-J(U)I)U$, as stated. If $O$ is complex Hermitian and $U$ is real, then $\Tr(\rho(U)O)$ involves only the real part of $O$. In that case the formula holds with $O$ replaced by $\re O$.
\end{proof}

\subsection{Proofs for Section~\ref{sec:policies}}\label{app:proofs-policies}

\begin{proof}[Proof of Proposition~\ref{prop:MDPbehavior}]
Fix a state $s$, and let $a$ be drawn from $\pi(\cdot\mid s)$, where $\pi(a_i\mid s)=(\rho_s)_{ii}$. The expected reward at $s$ is
\[
\E[R(s,a)]=\sum_i(\rho_s)_{ii}\,R(s,a_i)=\sum_i(\rho_s)_{ii}\,(O_R)_{ii}=\Tr(\rho_sO_R) ,
\]
where the last step is Lemma~\ref{lem:phaseirrelevance} applied to the diagonal observable
\[
O_R=\diag(R(s,a_1),\dots,R(s,a_k)) .
\]
So the one-step expected reward depends on $\rho_s$ only through its diagonal.

This fact propagates to the whole trajectory. The trajectory distribution factorizes as
\[
P_0(s_0)\prod_t \pi(a_t\mid s_t)\,\mathcal P(s_{t+1}\mid s_t,a_t) ,
\]
where $P_0$ is the initial-state distribution and $\mathcal P$ is the environment's transition kernel, neither of which depends on the policy. The only factor that depends on the policy at all is $\pi(a_t\mid s_t)=(\rho_{s_t})_{a_ta_t}$, the diagonal entry of $\rho_{s_t}$ at the sampled action. No off-diagonal entry of $\rho_{s_t}$ enters this factor, or any other factor of the trajectory distribution, or the reward $R(s_t,a_t)$, which is a fixed function of the realized state-action pair. So the entire trajectory distribution, and every quantity computed from it (returns, value functions, visitation measures), depends on $\{\rho_s\}$ only through $\{\diag(\rho_s)\}$. The achievable behaviors therefore coincide exactly with those of the classical policy $\pi(a_i\mid s)=(\rho_s)_{ii}$.
\end{proof}

\subsection{Proofs for Section~\ref{sec:beyond}}\label{app:proofs-beyond}

\begin{proof}[Proof of Corollary~\ref{cor:coherentprior}]
Apply Theorem~\ref{thm:gibbs} with $Q_s$ replaced by $\tilde Q_s=Q_s+\tau\log\rho_s^{\mathrm{ref}}$. This is valid because $\tilde Q_s$ is Hermitian, being a sum of two Hermitian operators, and Theorem~\ref{thm:gibbs} places no other condition on its argument. This gives the stated form of $\rho_\tau(\tilde Q_s)$ directly.

For the limits, write $\rho_\tau(\tilde Q_s)=\exp(Q_s/\tau+\log\rho_s^{\mathrm{ref}})/Z$. As $\tau\to\infty$, $Q_s/\tau\to0$, so $\exp(Q_s/\tau+\log\rho_s^{\mathrm{ref}})\to\exp(\log\rho_s^{\mathrm{ref}})=\rho_s^{\mathrm{ref}}$ and $Z\to\Tr(\rho_s^{\mathrm{ref}})=1$. Hence $\rho_\tau(\tilde Q_s)\to\rho_s^{\mathrm{ref}}$. The case $\tau\to0^+$ needs a different argument, because $\tilde Q_s$ itself depends on $\tau$. We use the fact that $\rho_\tau(\tilde Q_s)$ maximizes $F_\tau^{\mathrm{ref}}$. Let $q_{\max}$ be the largest eigenvalue of $Q_s$, with eigenspace projector $\Pi_{\max}$. If $Q_s=q_{\max}I$, the top eigenspace is the whole space and there is nothing to prove. Otherwise let $q_2<q_{\max}$ be the second largest eigenvalue and set $\Delta_q=q_{\max}-q_2>0$.

Take a unit vector $\mathbf{y}$ in the top eigenspace and set $\rho_0=\mathbf{y}\mathbf{y}^\dagger$. Then $\Tr(\rho_0Q_s)=q_{\max}$ and
\[
D(\rho_0\Vert\rho_s^{\mathrm{ref}})=-\Tr(\rho_0\log\rho_s^{\mathrm{ref}})\le-\log\lambda_{\min}(\rho_s^{\mathrm{ref}}),
\]
which is finite because $\rho_s^{\mathrm{ref}}\succ0$. Write $B_{\mathrm{ref}}=-\log\lambda_{\min}(\rho_s^{\mathrm{ref}})$. Since $\rho_\tau(\tilde Q_s)$ maximizes $F_\tau^{\mathrm{ref}}$ and $D\ge0$ (Lemma~\ref{lem:klein}),
\[
\Tr(\rho_\tau(\tilde Q_s) Q_s)\ \ge\ \Tr(\rho_\tau(\tilde Q_s) Q_s)-\tau D(\rho_\tau(\tilde Q_s)\Vert\rho_s^{\mathrm{ref}})\ \ge\ F_\tau^{\mathrm{ref}}(\rho_0)\ \ge\ q_{\max}-\tau B_{\mathrm{ref}} .
\]
The eigenvalues of $Q_s$ outside the top eigenspace are at most $q_2$, so $\Tr(\rho_\tau(\tilde Q_s) Q_s)\le q_{\max}\Tr(\rho_\tau(\tilde Q_s)\Pi_{\max})+q_2\Tr\bigl(\rho_\tau(\tilde Q_s)(I-\Pi_{\max})\bigr)=q_{\max}-\Delta_q\bigl[1-\Tr(\rho_\tau(\tilde Q_s)\Pi_{\max})\bigr]$. Combining the two displays gives
\[
1-\Tr(\rho_\tau(\tilde Q_s)\Pi_{\max})\ \le\ \tau B_{\mathrm{ref}}/\Delta_q\ \to\ 0 .
\]
So $\rho_\tau(\tilde Q_s)$ puts all its weight on the top eigenspace in the limit. If that eigenvalue is simple, $\Pi_{\max}$ has rank one and $\Tr(\rho_\tau(\tilde Q_s)\Pi_{\max})\to1$ forces $\rho_\tau(\tilde Q_s)\to\Pi_{\max}$. Indeed, $\Tr(\rho_\tau(\tilde Q_s)(I-\Pi_{\max}))\to0$ and $\rho_\tau(\tilde Q_s)\succeq0$ imply that the blocks of $\rho_\tau(\tilde Q_s)$ outside $\Pi_{\max}\rho_\tau(\tilde Q_s)\Pi_{\max}=\Tr(\rho_\tau(\tilde Q_s)\Pi_{\max})\,\Pi_{\max}$ tend to zero. If the top eigenvalue is degenerate, this argument does not determine the limit inside the top eigenspace, and that limit can depend on $\rho_s^{\mathrm{ref}}$.
\end{proof}

\begin{proof}[Proof of Proposition~\ref{prop:coherencetransfer}]
Write $\tilde Q_s=Q_s+\tau\log\rho_s^{\mathrm{ref}}$ as in Corollary~\ref{cor:coherentprior}. Since $Q_s$ is diagonal in the action basis, its off-diagonal entries are all zero. So for $i\ne j$, $(\tilde Q_s)_{ij}=\tau(\log\rho_s^{\mathrm{ref}})_{ij}$. This means $\tilde Q_s$ is diagonal in the action basis if and only if $\log\rho_s^{\mathrm{ref}}$ is.

Next, $\log\rho_s^{\mathrm{ref}}$ and $\rho_s^{\mathrm{ref}}$ share the same eigenvectors, since if $\rho_s^{\mathrm{ref}}=\sum_n\lambda_n'\mathbf{v}_n'\mathbf{v}_n'^\dagger$ then $\log\rho_s^{\mathrm{ref}}=\sum_n(\log\lambda_n')\mathbf{v}_n'\mathbf{v}_n'^\dagger$ by the same spectral decomposition. A Hermitian operator is diagonal in the action basis exactly when its eigenvectors can be taken to be the standard basis vectors $\eb_i$. So $\log\rho_s^{\mathrm{ref}}$ is diagonal in the action basis if and only if $\rho_s^{\mathrm{ref}}$ is.

Finally, $\rho_\tau(\tilde Q_s)=f(\tilde Q_s)$ where $f(y)=e^{y/\tau}/Z$ is a strictly increasing function of a real eigenvalue $y$, applied to $\tilde Q_s$ through its spectral decomposition. A strictly increasing scalar function preserves eigenvectors and never merges distinct eigenvalues into one, so $\rho_\tau(\tilde Q_s)$ shares its eigenspaces exactly with $\tilde Q_s$. In particular, $\rho_\tau(\tilde Q_s)$ is diagonal in the action basis if and only if $\tilde Q_s$ is.

Chaining the three equivalences gives the claim. $\rho_\tau(\tilde Q_s)$ is diagonal in the action basis if and only if $\tilde Q_s$ is, if and only if $\log\rho_s^{\mathrm{ref}}$ is, if and only if $\rho_s^{\mathrm{ref}}$ is.
\end{proof}

\begin{proof}[Proof of Proposition~\ref{prop:secondmoment}]
Let $\rho=\E_\pi[\mathbf{x}\mathbf{x}^\top]$. For a unit vector $\mathbf{x}$, the matrix $\mathbf{x}\mathbf{x}^\top$ is symmetric and PSD, and its trace is $\|\mathbf{x}\|^2=1$. Taking the expectation over $\pi$ preserves all three properties, so $\rho\in\mathcal D_k$. By linearity of the trace, $\E[R]=\E[\mathbf{x}^\top O\mathbf{x}]=\E[\Tr(O\,\mathbf{x}\mathbf{x}^\top)]=\Tr(O\rho)$. For the converse, take any real symmetric $\rho\in\mathcal D_k$ and write its spectral decomposition $\rho=\sum_n\lambda_n\mathbf{v}_n\mathbf{v}_n^\top$ with real orthonormal eigenvectors $\mathbf{v}_n$. Let $\pi$ choose $\mathbf{v}_n$ with probability $\lambda_n$. Then $\E_\pi[\mathbf{x}\mathbf{x}^\top]=\sum_n\lambda_n\mathbf{v}_n\mathbf{v}_n^\top=\rho$. The entry formulas follow from $(\mathbf{x}\mathbf{x}^\top)_{ij}=x_ix_j$.
\end{proof}

\begin{proof}[Proof of Proposition~\ref{prop:coherentdominance}]
Write $\rho=\rho_{\mathrm{diag}}+C$, where $C$ is hollow, meaning zero on the diagonal, and collects the off-diagonal entries. Then $\Tr(\rho\,O)=\Tr(\rho_{\mathrm{diag}}O)+\Tr(C O)$. The second term is linear in $C$, and it is not constant, since $O_{ij}\ne0$. The feasible set of $C$ is convex, being the intersection of the PSD cone with the affine slice $\diag(\rho)=\mathbf{p}$. Every $p_i$ is strictly positive, so $\rho_{\mathrm{diag}}\succ0$, and the feasible set therefore contains a neighborhood of $C=0$ inside the space of hollow symmetric matrices. A non-constant linear function never attains its maximum at an interior point of a convex set. So the maximizing $C$ is nonzero, and the maximum strictly exceeds the value at $C=0$, which is $\Tr(\rho_{\mathrm{diag}}O)$.
\end{proof}

\begin{proof}[Proof of Proposition~\ref{prop:gap}]
Write $\rho=\sum_n\lambda_n\mathbf{v}_n\mathbf{v}_n^\top$ with $\lambda_n\ge0$, $\sum_n\lambda_n=1$, and orthonormal $\mathbf{v}_n$. Then $\Tr(\rho O)=\sum_n\lambda_n\,\mathbf{v}_n^\top O\mathbf{v}_n\le\lambda_{\max}(O)$ by the Rayleigh bound. Equality holds when $\rho$ is supported on the top eigenspace. So the maximum over $\mathcal D_k$ is $\lambda_{\max}(O)$. For a classical policy, $\Tr(\diag(\mathbf{p})\,O)=\sum_ip_iO_{ii}\le\max_iO_{ii}$, with equality at a point mass. Taking $\rho=\eb_i\eb_i^\dagger$ shows $\lambda_{\max}(O)\ge O_{ii}$ for every $i$. So $\mathrm{margin}(O)\ge0$.

Suppose $\mathrm{margin}(O)=0$. Pick $i$ with $O_{ii}=\max_jO_{jj}$. Then $\eb_i^\top O\eb_i=\lambda_{\max}(O)$. The Rayleigh quotient reaches its maximum only on the top eigenspace, so $\eb_i$ is an eigenvector of $O$ with eigenvalue $\lambda_{\max}(O)$. That means $O_{ji}=0$ for every $j\ne i$, and $O_{ii}=\lambda_{\max}(O)$. Conversely, suppose some $i$ has $O_{ij}=0$ for all $j\ne i$ and $O_{ii}=\lambda_{\max}(O)$. Then $\max_jO_{jj}\ge O_{ii}=\lambda_{\max}(O)\ge\max_jO_{jj}$, so $\mathrm{margin}(O)=0$. The last claim follows because, if every action is coupled to at least one other action, no $i$ satisfies this condition.
\end{proof}

\begin{proof}[Proof of Corollary~\ref{cor:coupledmeasurement}]
Since $\{\mathbf{b}_n\}$ is an orthonormal basis and $\|\mathbf{x}\|=1$, we have $\sum_n(\mathbf{b}_n^\top \mathbf{x})^2=\|\mathbf{x}\|^2=1$. So the evaluator's probabilities are valid. Averaging over the chosen vector gives $\Pr(n)=\E[(\mathbf{b}_n^\top \mathbf{x})^2]=\E[\mathbf{b}_n^\top \mathbf{x}\mathbf{x}^\top \mathbf{b}_n]=\Tr(\rho\, \mathbf{b}_n\mathbf{b}_n^\top)$. Hence $\E[R]=\sum_nR_n\Tr(\rho\, \mathbf{b}_n\mathbf{b}_n^\top)=\Tr(\rho\, O)$.

For the second claim, suppose the $R_n$ are distinct. Then the eigenvalues of $O$ are distinct, and each eigenspace is spanned by one $\mathbf{b}_n$. The matrix $O$ is diagonal in the standard basis exactly when every $\mathbf{b}_n$ is a standard basis vector up to sign. That is, the $\mathbf{b}_n$ are the action directions, possibly reordered or with flipped signs. If every $R_n$ equals $R_0$, then $O=R_0\sum_n\mathbf{b}_n\mathbf{b}_n^\top=R_0 I$, which is diagonal for every basis.
\end{proof}

\begin{proof}[Proof of Proposition~\ref{prop:multistep}]
The transition probabilities are valid, since $\mathbf{x}^\top M_{s'}(s)\mathbf{x}\ge0$ and
\[
\sum_{s'}\mathbf{x}^\top M_{s'}(s)\,\mathbf{x}=\mathbf{x}^\top\Bigl(\sum_{s'}M_{s'}(s)\Bigr)\mathbf{x}=\mathbf{x}^\top \mathbf{x}=1.
\]
Fix a policy $\pi$, and let $\rho_s$ be its second-moment matrix at state $s$. Let $V^\pi$ be its value function, so $V^\pi(s)=0$ at terminal states. At any other state, averaging over the chosen $\mathbf{x}$ gives
\[
V^\pi(s)=\E\Bigl[\mathbf{x}^\top O_s\mathbf{x}+\sum_{s'}\mathbf{x}^\top M_{s'}(s)\,\mathbf{x}\,V^\pi(s')\Bigr]=\Tr\Bigl(\rho_s\Bigl(O_s+\sum_{s'}V^\pi(s')M_{s'}(s)\Bigr)\Bigr).
\]
The right side depends on the policy only through $\rho_s$ and the values of the successor states. Because the time step is part of the state, every successor $s'$ lies at a later step. Backward induction from the final step therefore determines $V^\pi$ at every state from $\{\rho_s\}$ alone. Two policies with the same $\rho_s$ at every state thus have the same value at every state, which is the expected return from that state. We can therefore write $V^\rho$ for this common value function.
\end{proof}

\subsection{Proofs for Appendix~\ref{app:rank}}\label{app:proofs-rank}

\begin{proof}[Proof of Theorem~\ref{thm:spectraltruncation}]
Write $\mu_n=\mathbf{v}_n^\dagger O\mathbf{v}_n$ for $n=1,\dots,k$. Then $\Tr(\rho O)=\sum_{n=1}^k\lambda_n\mu_n$. The same identity applied to $\rho_r$ gives $\Tr(\rho_rO)=\tfrac{1}{\mathrm{cap}(r)}\sum_{n=1}^r\lambda_n\mu_n$. Each $\mathbf{v}_n$ is a unit vector, so $|\mu_n|=|\mathbf{v}_n^\dagger O\mathbf{v}_n|\le\|O\|_{\mathrm{op}}$ for every $n$, by definition of the operator norm.

Let $T_{\le r}=\sum_{n=1}^r\lambda_n\mu_n$ and $T_{>r}=\sum_{n=r+1}^k\lambda_n\mu_n$. Then $\Tr(\rho O)=T_{\le r}+T_{>r}$ and $\Tr(\rho_rO)=T_{\le r}/\mathrm{cap}(r)$, so
\[
\Tr(\rho O)-\Tr(\rho_rO)=\Bigl(1-\frac{1}{\mathrm{cap}(r)}\Bigr)T_{\le r}+T_{>r} .
\]
Since $|\mu_n|\le\|O\|_{\mathrm{op}}$ and every $\lambda_n\ge0$, we get $|T_{\le r}|\le\|O\|_{\mathrm{op}}\sum_{n=1}^r\lambda_n=\|O\|_{\mathrm{op}}\mathrm{cap}(r)$. The same reasoning, using $\sum_{n=1}^k\lambda_n=1$, gives $|T_{>r}|\le\|O\|_{\mathrm{op}}\sum_{n=r+1}^k\lambda_n=\|O\|_{\mathrm{op}}(1-\mathrm{cap}(r))$. Since $\mathrm{cap}(r)\in(0,1]$, we have $1-\tfrac{1}{\mathrm{cap}(r)}\le0$, so $\bigl|1-\tfrac1{\mathrm{cap}(r)}\bigr|=\tfrac{1-\mathrm{cap}(r)}{\mathrm{cap}(r)}$. Combining these bounds,
\begin{align*}
\bigl|\Tr(\rho O)-\Tr(\rho_rO)\bigr|
&\le\frac{1-\mathrm{cap}(r)}{\mathrm{cap}(r)}\,|T_{\le r}|+|T_{>r}|\\
&\le\frac{1-\mathrm{cap}(r)}{\mathrm{cap}(r)}\cdot\|O\|_{\mathrm{op}}\mathrm{cap}(r)+\|O\|_{\mathrm{op}}(1-\mathrm{cap}(r))\\
&=2\,\|O\|_{\mathrm{op}}(1-\mathrm{cap}(r)) .\qedhere
\end{align*}
\end{proof}

\begin{proof}[Proof of Corollary~\ref{cor:effectiverank}]
By definition of $r_{\mathrm{eff}}(\varepsilon)$, $\mathrm{cap}(r_{\mathrm{eff}}(\varepsilon))\ge1-\varepsilon$, so $1-\mathrm{cap}(r_{\mathrm{eff}}(\varepsilon))\le\varepsilon$. Applying Theorem~\ref{thm:spectraltruncation} at $r=r_{\mathrm{eff}}(\varepsilon)$ with $\|O\|_{\mathrm{op}}\le1$ gives $\bigl|\Tr(\rho O)-\Tr(\rho_rO)\bigr|\le2\varepsilon$.
\end{proof}

\begin{proof}[Proof of Proposition~\ref{prop:purityrank}]
By the Cauchy--Schwarz inequality applied to the vectors $(\lambda_1,\dots,\lambda_r)$ and $(1,\dots,1)\in\R^r$,
\[
\mathrm{cap}(r)^2=\Bigl(\sum_{n=1}^r\lambda_n\Bigr)^2\le r\sum_{n=1}^r\lambda_n^2\le r\sum_{n=1}^k\lambda_n^2=r\,\Tr(\rho^2) .
\]
The second inequality drops the nonnegative terms $\lambda_{r+1}^2,\dots,\lambda_k^2$. The final equality uses $\Tr(\rho^2)=\sum_n\lambda_n^2$, which follows from the spectral decomposition $\rho^2=\sum_n\lambda_n^2\mathbf{v}_n\mathbf{v}_n^\dagger$. We always have $\Tr\rho=1$, so $\rho\ne0$ and thus $\Tr(\rho^2)>0$. Rearranging gives $r\ge \mathrm{cap}(r)^2/\Tr(\rho^2)$. Taking $r=r_{\mathrm{eff}}(\varepsilon)$, for which $\mathrm{cap}(r)\ge1-\varepsilon$, gives $r_{\mathrm{eff}}(\varepsilon)\ge(1-\varepsilon)^2/\Tr(\rho^2)$.
\end{proof}

\subsection{Proofs for Appendix~\ref{sec:entropy}}\label{app:proofs-entropy}

\begin{proof}[Proof of Theorem~\ref{thm:gibbs}]
Let $Z=\Tr e^{Q_s/\tau}$, so that $\rho_\tau(Q_s)=e^{Q_s/\tau}/Z$. Since $\rho_\tau(Q_s)$ is a positive scalar multiple of $e^{Q_s/\tau}$, taking logarithms gives $\log\rho_\tau(Q_s)=Q_s/\tau-(\log Z)\,I$. The operator $Q_s/\tau$ commutes with itself, so no ordering issues arise, and dividing by the scalar $Z$ simply subtracts $\log Z$ from the exponent. For any $\rho\in\mathcal D_k$,
\[
\Tr(\rho\log\rho_\tau(Q_s))=\Tr\Bigl(\rho\Bigl(\frac{Q_s}{\tau}-(\log Z)I\Bigr)\Bigr)=\frac{1}{\tau}\Tr(\rho Q_s)-(\log Z)\Tr(\rho)=\frac{1}{\tau}\Tr(\rho Q_s)-\log Z ,
\]
using linearity of the trace and $\Tr(\rho)=1$. So, using $S(\rho)=-\Tr(\rho\log\rho)$ (Definition~\ref{def:vne}),
\[
D(\rho\,\Vert\,\rho_\tau(Q_s))
=\Tr(\rho\log\rho)-\Tr(\rho\log\rho_\tau(Q_s))
=-S(\rho)-\frac{1}{\tau}\Tr(\rho Q_s)+\log Z .
\]
Rearranging this identity to isolate $F_\tau(\rho)=\Tr(\rho Q_s)+\tau S(\rho)$ gives
\[
F_\tau(\rho)=\tau\log Z-\tau\,D(\rho\,\Vert\,\rho_\tau(Q_s)).
\]
By Klein's inequality (Lemma~\ref{lem:klein}), $D(\rho\,\Vert\,\rho_\tau(Q_s))\ge0$ for every $\rho\in\mathcal D_k$, with equality exactly at $\rho=\rho_\tau(Q_s)$. Since $\tau>0$, this gives $F_\tau(\rho)\le\tau\log Z$ for every $\rho$, with equality exactly when $\rho=\rho_\tau(Q_s)$. So $\rho_\tau(Q_s)$ is the unique maximizer of $F_\tau$ over $\mathcal D_k$, and the maximal value is $\tau\log Z=\tau\log\Tr e^{Q_s/\tau}$.
\end{proof}

\begin{proof}[Proof of Corollary~\ref{cor:annealing}]
Write $\beta=1/\tau$, so that $\rho_\beta=\rho_\tau(Q_s)=e^{\beta Q_s}/Z(\beta)$ with $Z(\beta)=\Tr e^{\beta Q_s}$.

(1) As $\beta\to0$, every eigenvalue of $e^{\beta Q_s}$ tends to $e^0=1$, so $e^{\beta Q_s}\to I$ and $Z(\beta)\to k$. Hence $\rho_\beta\to I/k$.

(2) Diagonalize $Q_s=\sum_i q_i\mathbf{y}_i\mathbf{y}_i^\dagger$ with maximum eigenvalue $q_{\max}$, attained on an eigenspace with projector $\Pi_{\max}$. Dividing numerator and denominator of $\rho_\beta$ by $e^{\beta q_{\max}}$ gives
\[
\rho_\beta=\frac{\sum_ie^{\beta(q_i-q_{\max})}\mathbf{y}_i\mathbf{y}_i^\dagger}{\sum_je^{\beta(q_j-q_{\max})}} .
\]
For $q_i<q_{\max}$, $e^{\beta(q_i-q_{\max})}\to0$ as $\beta\to\infty$, while for $q_i=q_{\max}$ this factor stays at $1$ for every $\beta$. So the numerator tends to $\Pi_{\max}$, the sum over the top eigenvectors, and the denominator tends to $\mathrm{rank}\,\Pi_{\max}$, their count. Hence $\rho_\beta\to\Pi_{\max}/\mathrm{rank}\,\Pi_{\max}$.

(3) Since $\rho_\beta$ is a function of $Q_s$, it shares the eigenbasis $\{\mathbf{y}_i\}$ of $Q_s$, with eigenvalues $\lambda_i=e^{\beta q_i}/Z(\beta)$. This makes the entropy calculation purely classical, treating $\{\lambda_i\}$ as a probability vector. Write $\langle Q_s\rangle_\beta=\sum_i\lambda_iq_i=\Tr(\rho_\beta Q_s)$ for the mean of $Q_s$ under $\rho_\beta$. Then
\begin{align*}
S(\rho_\beta)&=-\sum_i\lambda_i\log \lambda_i=-\sum_i\lambda_i\bigl(\beta q_i-\log Z(\beta)\bigr)\\
&=-\beta\sum_i\lambda_iq_i+\log Z(\beta)\sum_i\lambda_i=\log Z(\beta)-\beta\langle Q_s\rangle_\beta ,
\end{align*}
using $\sum_i\lambda_i=1$. Since $Z(\beta)=\sum_ie^{\beta q_i}$, differentiating under the sum gives $Z'(\beta)=\sum_iq_ie^{\beta q_i}$, so
\[
\frac{Z'(\beta)}{Z(\beta)}=\sum_iq_i\,\frac{e^{\beta q_i}}{Z(\beta)}=\sum_iq_i\lambda_i=\langle Q_s\rangle_\beta .
\]
Differentiating $S(\rho_\beta)=\log Z(\beta)-\beta\langle Q_s\rangle_\beta$ in $\beta$, and using $\tfrac{d}{d\beta}\log Z(\beta)=Z'(\beta)/Z(\beta)=\langle Q_s\rangle_\beta$,
\[
\frac{dS}{d\beta}=\langle Q_s\rangle_\beta-\langle Q_s\rangle_\beta-\beta\frac{d\langle Q_s\rangle_\beta}{d\beta}=-\beta\frac{d\langle Q_s\rangle_\beta}{d\beta} .
\]
It remains to identify $d\langle Q_s\rangle_\beta/d\beta$ with $\Var_{\rho_\beta}(Q_s)$. Differentiating $\langle Q_s\rangle_\beta=Z'(\beta)/Z(\beta)$ by the quotient rule,
\[
\frac{d\langle Q_s\rangle_\beta}{d\beta}=\frac{Z''(\beta)Z(\beta)-Z'(\beta)^2}{Z(\beta)^2}=\frac{Z''(\beta)}{Z(\beta)}-\langle Q_s\rangle_\beta^2 .
\]
Since $Z''(\beta)=\sum_iq_i^2e^{\beta q_i}$, we get $Z''(\beta)/Z(\beta)=\sum_iq_i^2\lambda_i=\Tr(\rho_\beta Q_s^2)$, so
\[
\frac{d\langle Q_s\rangle_\beta}{d\beta}=\Tr(\rho_\beta Q_s^2)-\bigl(\Tr(\rho_\beta Q_s)\bigr)^2=\Var_{\rho_\beta}(Q_s)\ge0 .
\]
Substituting back, $dS/d\beta=-\beta\,\Var_{\rho_\beta}(Q_s)\le0$ for $\beta\ge0$. Entropy is therefore nonincreasing in $\beta$, that is, nondecreasing in $\tau=1/\beta$.
\end{proof}

\begin{proof}[Proof of Lemma~\ref{lem:welldef}]
Write $Y=e^{\beta Q_s/2}$. Since $Q_s$ is Hermitian, it has a spectral decomposition $Q_s=\sum_iq_i\mathbf{y}_i\mathbf{y}_i^\dagger$ with real eigenvalues $q_i$, so $Y=\sum_ie^{\beta q_i/2}\mathbf{y}_i\mathbf{y}_i^\dagger$ is Hermitian with strictly positive eigenvalues $e^{\beta q_i/2}>0$. So $Y$ is Hermitian and positive definite, and in particular invertible.

\emph{Hermiticity of the numerator $Y\rho Y$.} $(Y\rho Y)^\dagger=Y^\dagger\rho^\dagger Y^\dagger=Y\rho Y$, using $Y^\dagger=Y$ and $\rho^\dagger=\rho$.

\emph{Positive semidefiniteness of the numerator.} For any $\mathbf{v}\in\C^k$, $\mathbf{v}^\dagger(Y\rho Y)\mathbf{v}=(Y\mathbf{v})^\dagger\rho(Y\mathbf{v})\ge0$, since $\rho\succeq0$ and $Y\mathbf{v}$ is just another vector in $\C^k$.

\emph{Strict positivity of the denominator.} By cyclicity of the trace, $\Tr(Y\rho Y)=\Tr(\rho Y^2)$. Since $Y^2\succ0$, with eigenvalues $e^{\beta q_i}>0$, we have $Y^2-\lambda_{\min}(Y^2)I\succeq0$, so $\Tr(\rho Y^2)\ge\lambda_{\min}(Y^2)\Tr(\rho)=\lambda_{\min}(Y^2)>0$. So the denominator $\Tr(Y\rho Y)$ is a strictly positive real number.

Dividing the Hermitian, PSD numerator by this positive scalar preserves both Hermiticity and positive semidefiniteness, and rescales the trace to exactly $1$. So $\mathcal T_{Q_s}^\beta(\rho)=Y\rho Y/\Tr(Y\rho Y)\in\mathcal D_k$.
\end{proof}

\begin{proof}[Proof of Theorem~\ref{thm:monotonic}]
Since $Q_s$ is a fixed Hermitian matrix, $e^{\beta Q_s/2}$ commutes with $Q_s$ for every $\beta$, both being functions of the same operator. Cyclicity of the trace then gives
\[
\Tr\bigl(\rho_0\,e^{\beta Q_s/2}Q_se^{\beta Q_s/2}\bigr)=\Tr\bigl(\rho_0\,Q_s\,e^{\beta Q_s}\bigr) ,
\]
so
\[
J(\beta)=\frac{\tilde N(\beta)}{\tilde Z(\beta)} ,
\]
where $\tilde N(\beta)=\Tr\bigl(\rho_0\,Q_s\,e^{\beta Q_s}\bigr)$ and $\tilde Z(\beta)=\Tr\bigl(\rho_0\,e^{\beta Q_s}\bigr)$.

\emph{Derivatives.} Since $Q_s$ does not depend on $\beta$, the ordinary rule $\tfrac{d}{d\beta}e^{\beta Q_s}=Q_se^{\beta Q_s}$ holds. Differentiating under the trace,
\[
\tilde N'(\beta)=\Tr\bigl(\rho_0Q_s^2e^{\beta Q_s}\bigr),\qquad
\tilde Z'(\beta)=\Tr\bigl(\rho_0Q_se^{\beta Q_s}\bigr)=\tilde N(\beta) .
\]
By the quotient rule,
\begin{align*}
J'(\beta)&=\frac{\tilde N'(\beta)\tilde Z(\beta)-\tilde N(\beta)\tilde Z'(\beta)}{\tilde Z(\beta)^2}
=\frac{\tilde N'(\beta)}{\tilde Z(\beta)}-\frac{\tilde N(\beta)^2}{\tilde Z(\beta)^2}\\
&=\Tr(\tilde\rho_\beta Q_s^2)-\bigl(\Tr(\tilde\rho_\beta Q_s)\bigr)^2=\Var_{\tilde\rho_\beta}(Q_s)\ge0 ,
\end{align*}
where the second-to-last equality uses $\tilde\rho_\beta=e^{\beta Q_s/2}\rho_0e^{\beta Q_s/2}/\tilde Z(\beta)$: by cyclicity, $\Tr(\tilde\rho_\beta Q_s^2)=\tilde N'(\beta)/\tilde Z(\beta)$ and $\Tr(\tilde\rho_\beta Q_s)=\tilde N(\beta)/\tilde Z(\beta)=J(\beta)$. Nonnegativity holds because a variance is always nonnegative.

\emph{Equality case.} Recall $J(\beta)=\Tr(\tilde\rho_\beta Q_s)$. Expanding,
\[
\Var_{\tilde\rho_\beta}(Q_s)=\Tr\bigl((Q_s-J(\beta)I)\tilde\rho_\beta(Q_s-J(\beta)I)\bigr)=\bigl\|(Q_s-J(\beta)I)\tilde\rho_\beta^{1/2}\bigr\|_F^2 ,
\]
where the first equality follows from $\Tr(\tilde\rho_\beta Q_s^2)-2J(\beta)\Tr(\tilde\rho_\beta Q_s)+J(\beta)^2\Tr(\tilde\rho_\beta)=\Tr(\tilde\rho_\beta Q_s^2)-J(\beta)^2$, matching the variance, and the second uses $\Tr(B\tilde\rho_\beta B)=\|B\tilde\rho_\beta^{1/2}\|_F^2$ for Hermitian $B$, by cyclicity. A Frobenius norm vanishes exactly when its argument is zero, so the variance is zero iff $(Q_s-J(\beta)I)\tilde\rho_\beta^{1/2}=0$. Equivalently, every vector in $\supp(\tilde\rho_\beta)$ (the range of $\tilde\rho_\beta^{1/2}$) must lie in the eigenspace $\ker(Q_s-J(\beta)I)$ of $Q_s$. Since $\supp(\tilde\rho_\beta)=e^{\beta Q_s/2}\supp(\rho_0)$, the invertible operator $e^{\beta Q_s/2}$ maps the support of $\rho_0$ onto the support of $\tilde\rho_\beta$. Every eigenspace of $Q_s$ is invariant under $e^{\beta Q_s/2}$, since $e^{\beta Q_s/2}$ is itself a function of $Q_s$. So this containment holds iff $\supp(\rho_0)\subseteq\ker(Q_s-J(\beta)I)$, that is, iff $\supp(\rho_0)$ already lies in a single eigenspace of $Q_s$. \emph{The limit $\beta\to\infty$.} Write $Q_s=\sum_qq\,\Pi_q$, where $\Pi_q$ is the projector onto the eigenspace with eigenvalue $q$. Then $e^{\beta Q_s/2}=\sum_qe^{\beta q/2}\Pi_q$. Let $q_\star$ and $\Pi_\star$ be as in the theorem. For $q>q_\star$ we have $\Pi_q\rho_0=0$. This is because $\Tr(\Pi_q\rho_0\Pi_q)=\Tr(\Pi_q\rho_0)=0$ and $\rho_0\succeq0$ force $\rho_0^{1/2}\Pi_q=0$. Multiply the numerator and the denominator of $\tilde\rho_\beta$ by $e^{-\beta q_\star}$. The numerator becomes
\[
\sum_{q,q'\le q_\star}e^{\beta(q+q'-2q_\star)/2}\,\Pi_q\rho_0\Pi_{q'} .
\]
The term $q=q'=q_\star$ equals $\Pi_\star\rho_0\Pi_\star$. Every other term has a negative exponent and vanishes as $\beta\to\infty$. The denominator is the trace of the numerator, so it tends to $\Tr(\Pi_\star\rho_0\Pi_\star)=\Tr(\Pi_\star\rho_0)$. This is positive, since $\Pi_\star\rho_0\neq0$. Hence $\tilde\rho_\beta\to\Pi_\star\rho_0\Pi_\star/\Tr(\Pi_\star\rho_0)$. This limit lies in the eigenspace of $Q_s$ with eigenvalue $q_\star$, so $J(\beta)\to q_\star$.
\end{proof}

\begin{proof}[Proof of Proposition~\ref{prop:surrogate}]
\emph{Convexity and existence.} The objective $\rho\mapsto\Tr(\rho A_s^{(t)})$ is linear in $\rho$, hence continuous. The feasible set in~\eqref{eq:trustregion} is the intersection of $\mathcal D_k$, convex and compact by Proposition~\ref{prop:inclusion}, with the sublevel set $\{\rho:D(\rho\,\Vert\,\rho_s^{(t)})\le\delta\}$. This sublevel set is convex by Lemma~\ref{lem:jointconvex}, since joint convexity of $D$ implies convexity of its sublevel sets in the first argument. It is also closed, because $D(\cdot\,\Vert\,\rho_s^{(t)})$ is lower semicontinuous. An intersection of a compact set with a closed set is compact, so the feasible set is compact. A continuous function on a nonempty compact set attains a maximum, by the extreme value theorem, so a maximizer $\rho_s^{(t+1)}$ exists as long as the feasible set is nonempty. It is nonempty because Klein's inequality (Lemma~\ref{lem:klein}) gives $D(\rho_s^{(t)}\,\Vert\,\rho_s^{(t)})=0\le\delta$, so $\rho_s^{(t)}$ itself is feasible. Since $\rho_s^{(t+1)}$ maximizes over a set that contains $\rho_s^{(t)}$, $\Tr(\rho_s^{(t+1)}A_s^{(t)})\ge\Tr(\rho_s^{(t)}A_s^{(t)})$.

\emph{Every feasible point lies in the face.} By Definition~\ref{def:qre}, $D(\rho\,\Vert\,\rho_s^{(t)})=+\infty$ whenever $\supp\rho\not\subseteq\supp\rho_s^{(t)}$. So any $\rho$ with $D(\rho\,\Vert\,\rho_s^{(t)})\le\delta<\infty$ must satisfy $\supp\rho\subseteq\supp\rho_s^{(t)}$, meaning the feasible set is contained in the face $\mathcal F(\rho_s^{(t)})$. This is exactly why strictness must be, and is, stated relative to $\mathcal F(\rho_s^{(t)})$ (Remark~\ref{rem:facenecessary}).

\emph{Strictness on the face.} Suppose there exists $\rho''\in\mathcal F(\rho_s^{(t)})$ with $\Tr(\rho''A_s^{(t)})>\Tr(\rho_s^{(t)}A_s^{(t)})$, and consider the segment $\rho_\chi=(1-\chi)\rho_s^{(t)}+\chi\rho''$, $\chi\in[0,1]$. This segment lies in $\mathcal F(\rho_s^{(t)})$ for every $\chi$, since a face of a convex set is itself convex and both endpoints lie in it.

Restrict attention to the subspace $\supp\rho_s^{(t)}$. On this subspace $\rho_s^{(t)}$ is positive definite by definition of the support, so $\log\rho_s^{(t)}$ is finite there. The map $\rho\mapsto\Tr(\rho\log\rho)$ is continuous on density matrices supported on this subspace, using the convention $0\log0=0$. The map $\rho\mapsto\Tr(\rho\log\rho_s^{(t)})$ is linear, hence continuous, on the same subspace. So $\chi\mapsto D(\rho_\chi\,\Vert\,\rho_s^{(t)})$ is continuous on $[0,1]$, with value $0$ at $\chi=0$ by Klein's inequality applied to $\rho_s^{(t)}$ against itself.

By this continuity, $D(\rho_\chi\,\Vert\,\rho_s^{(t)})\le\delta$ for all sufficiently small $\chi>0$, so $\rho_\chi$ is feasible for such $\chi$. Meanwhile, linearity of the objective gives
\[
\Tr(\rho_\chi A_s^{(t)})=\Tr(\rho_s^{(t)}A_s^{(t)})+\chi\bigl(\Tr(\rho''A_s^{(t)})-\Tr(\rho_s^{(t)}A_s^{(t)})\bigr)>\Tr(\rho_s^{(t)}A_s^{(t)})
\]
for every $\chi\in(0,1]$, since the bracketed term is strictly positive. So for small enough $\chi>0$, $\rho_\chi$ is a feasible point whose objective value strictly exceeds $\Tr(\rho_s^{(t)}A_s^{(t)})$. So the maximum over $\mathcal F(\rho_s^{(t)})$ strictly exceeds $\Tr(\rho_s^{(t)}A_s^{(t)})$ whenever such a $\rho''$ exists.

\emph{Full rank is preserved along algorithmic trajectories.} Gibbs states $e^{Q_s/\tau}/\Tr e^{Q_s/\tau}$ have every eigenvalue strictly positive, as in the proof of Lemma~\ref{lem:welldef}, so they are positive definite, hence full rank. For the matrix softmax map of Definition~\ref{def:softmaxchannel}, $\mathcal T_{Q_s}^{\beta}(\rho)$ is proportional to $Y\rho Y$ with $Y=e^{\beta Q_s/2}$ invertible, since all its eigenvalues $e^{\beta q_i/2}$ are strictly positive. Congruence by an invertible matrix, the map $\rho\mapsto Y\rho Y$ with $Y=Y^\dagger$, preserves the inertia of a Hermitian matrix, meaning the number of positive, negative, and zero eigenvalues, by Sylvester's law of inertia. In particular it preserves the rank, so $\mathrm{rank}\,\mathcal T_{Q_s}^{\beta}(\rho)=\mathrm{rank}\,\rho$: a full-rank input yields a full-rank output. So once an iterate is full rank, every later application of the matrix softmax map keeps it full rank. When $\rho_s^{(t)}$ is full rank, its support is the whole space, so $\mathcal F(\rho_s^{(t)})=\mathcal D_k$. The strictness guarantee above, which a priori only holds relative to the face $\mathcal F(\rho_s^{(t)})$, therefore coincides with strictness relative to all of $\mathcal D_k$.
\end{proof}

\begin{proof}[Proof of Theorem~\ref{thm:pinching}]
Let $\Phi(\rho)=\sum_i \eb_i\eb_i^\dagger\rho \eb_i\eb_i^\dagger$ be the pinching (completely dephasing) channel in the action basis. Each term $\rho\mapsto \eb_i\eb_i^\dagger\rho \eb_i\eb_i^\dagger$ is completely positive, being conjugation by the single Kraus operator $\eb_i\eb_i^\dagger$, so the sum $\Phi$ is completely positive. It is trace preserving because $\sum_i(\eb_i\eb_i^\dagger)^\dagger(\eb_i\eb_i^\dagger)=\sum_i\eb_i\eb_i^\dagger=I$. This is the completeness relation for the orthonormal action basis. So $\Phi$ is CPTP.

Applying $\Phi$ to any $\rho$ keeps only the diagonal entries in the action basis. Indeed, $\eb_i^\dagger\rho \eb_i=\rho_{ii}$, so each summand equals $\rho_{ii}\,\eb_i\eb_i^\dagger$. Hence $\Phi(\rho)=\diag(\rho_{11},\dots,\rho_{kk})=\diag(\pi(\cdot\mid s))$, where $\pi(\cdot\mid s)$ is the distribution given by the diagonal of $\rho$, and similarly $\Phi(\rho_s^{(t)})=\diag(\pi^{(t)}(\cdot\mid s))$.

By the data-processing inequality (Lemma~\ref{lem:dpi}) applied to $\Phi$,
\[
D\bigl(\Phi(\rho)\,\Vert\,\Phi(\rho_s^{(t)})\bigr)\le D(\rho\,\Vert\,\rho_s^{(t)})\le\delta .
\]
Since $\Phi(\rho)$ and $\Phi(\rho_s^{(t)})$ are both diagonal, the matrix relative entropy between them reduces exactly to the classical KL divergence between their diagonals. This is because Definition~\ref{def:qre} specializes to the classical formula when both arguments are simultaneously diagonal, so $D(\Phi(\rho)\,\Vert\,\Phi(\rho_s^{(t)}))=D_{\mathrm{KL}}(\pi(\cdot\mid s)\,\Vert\,\pi^{(t)}(\cdot\mid s))$. Combining,
\[
D_{\mathrm{KL}}(\pi(\cdot\mid s)\,\Vert\,\pi^{(t)}(\cdot\mid s))\le\delta .
\]
The classical Pinsker inequality then gives $\tfrac12\|\pi-\pi^{(t)}\|_1^2\le D_{\mathrm{KL}}(\pi(\cdot\mid s)\,\Vert\,\pi^{(t)}(\cdot\mid s))\le\delta$, so $\|\pi(\cdot\mid s)-\pi^{(t)}(\cdot\mid s)\|_1\le\sqrt{2\delta}$.

The transfer of the CPI and TRPO bounds is then immediate in total variation. Those guarantees depend on the new policy only through the per-state distance between the old and new action distributions. Total variation is half the $\ell_1$ distance, so $\mathrm{TV}(\pi,\pi^{(t)})\le\sqrt{\delta/2}$. Total variation is symmetric in the two policies. So the total-variation form of the guarantees carries over, with this bound as the per-state distance. The KL bound has the direction $D_{\mathrm{KL}}(\pi\Vert\pi^{(t)})$, which is the reverse of the usual TRPO constraint.
\end{proof}

\subsection{Proofs for Appendix~\ref{sec:theory}}\label{app:proofs-variance}

\begin{proof}[Proof of Theorem~\ref{thm:raoblackwell}]
\emph{Unbiasedness.} For $\hat J_{\mathrm{mix}}=X_W$: conditional on $W=n$, $X_W=X_n$, so $\E[X_W\mid W=n]=\E[X_n]=\mu_n$. Averaging over $W$, $\E[X_W]=\sum_n\Pr(W=n)\,\E[X_W\mid W=n]=\sum_n\lambda_n\mu_n=J$, using $\Pr(W=n)=\lambda_n$. For $\hat J_{\mathrm{EA}}=\sum_n\lambda_nX_n$: by linearity, $\E[\hat J_{\mathrm{EA}}]=\sum_n\lambda_n\E[X_n]=\sum_n\lambda_n\mu_n=J$.

\emph{Variance of $\hat J_{\mathrm{mix}}$.} By the law of total variance, conditioning on the sampled mode $W$,
\[
\Var(X_W)=\underbrace{\E\bigl[\Var(X_W\mid W)\bigr]}_{\text{within-mode}}+\underbrace{\Var\bigl(\E[X_W\mid W]\bigr)}_{\text{between-mode}} .
\]
For the first term, $\Var(X_W\mid W=n)=\Var(X_n)=\sigma_n^2$, so $\E[\Var(X_W\mid W)]=\sum_n\lambda_n\sigma_n^2$. For the second term, $\E[X_W\mid W=n]=\mu_n$, so $\E[X_W\mid W]$ takes value $\mu_n$ with probability $\lambda_n$, giving $\Var\bigl(\E[X_W\mid W]\bigr)=\sum_n\lambda_n\mu_n^2-\bigl(\sum_n\lambda_n\mu_n\bigr)^2=\sum_n\lambda_n\mu_n^2-J^2$. Adding the two terms gives the stated decomposition.

\emph{Variance of $\hat J_{\mathrm{EA}}$.} $\hat J_{\mathrm{EA}}=\sum_n\lambda_nX_n$ is a fixed linear combination of the independent random variables $\{X_n\}$ with deterministic coefficients $\lambda_n$, so $\Var(\hat J_{\mathrm{EA}})=\sum_n\lambda_n^2\Var(X_n)=\sum_n\lambda_n^2\sigma_n^2$.

\emph{Comparing the two.} Since $0\le\lambda_n\le1$, squaring gives $\lambda_n^2\le\lambda_n$, so $\lambda_n^2\sigma_n^2\le\lambda_n\sigma_n^2$ term by term, and summing, $\sum_n\lambda_n^2\sigma_n^2\le\sum_n\lambda_n\sigma_n^2$. The between-mode term $\sum_n\lambda_n\mu_n^2-J^2$ is itself a variance, of the discrete random variable $\mu_W$, hence nonnegative by Jensen's inequality applied to $y\mapsto y^2$. Combining,
\[
\Var(\hat J_{\mathrm{EA}})=\sum_n\lambda_n^2\sigma_n^2\ \le\ \sum_n\lambda_n\sigma_n^2\ \le\ \sum_n\lambda_n\sigma_n^2+\Bigl(\sum_n\lambda_n\mu_n^2-J^2\Bigr)=\Var(\hat J_{\mathrm{mix}}) ,
\]
which is the claimed inequality. Equality throughout requires the between-mode term to vanish, $\sum_n\lambda_n\mu_n^2=J^2$, and every gap $\lambda_n\sigma_n^2-\lambda_n^2\sigma_n^2=\lambda_n(1-\lambda_n)\sigma_n^2$ to vanish.
\end{proof}

\begin{proof}[Proof of Theorem~\ref{thm:controlvariate}]
\emph{Unbiasedness.} $\E[\hat J(c)]=p\,\E[\hat o_{11}]+(1-p)\,\E[\hat o_{22}]+2c\,\E[\hat o_{12}]=p\,O_{11}+(1-p)\,O_{22}+2c\,O_{12}=J(\rho_{c})$, using unbiasedness of each estimator and linearity of expectation. This holds for every $c$, so in particular $\hat J(0)$ is unbiased for $J(\rho_0)=p\,O_{11}+(1-p)\,O_{22}$.

\emph{Variance identity.} Comparing definitions, $\hat J(c)=\hat J(0)+2c\,\hat o_{12}$. Expanding the variance of this sum,
\[
\Var\bigl(\hat J(c)\bigr)=\Var\bigl(\hat J(0)\bigr)+2\cdot2c\,\Cov\bigl(\hat J(0),\hat o_{12}\bigr)+(2c)^2\Var(\hat o_{12})
=\Var\bigl(\hat J(0)\bigr)+4c\,\kappa+4c^2\sigma_{12}^2 ,
\]
using $\kappa=\Cov(\hat J(0),\hat o_{12})$ and $\sigma_{12}^2=\Var(\hat o_{12})$.

\emph{Where the quadratic dips below its value at $c=0$.} Let $f(c)=4c\kappa+4c^2\sigma_{12}^2=\Var(\hat J(c))-\Var(\hat J(0))$. Factoring, $f(c)=4c(\kappa+c\sigma_{12}^2)$, which has roots at $c=0$ and $c=-\kappa/\sigma_{12}^2$. Since $\kappa\ne0$, the Cauchy--Schwarz inequality gives $\sigma_{12}^2>0$. So the parabola $f$ opens upward, so $f(c)<0$ exactly on the open interval strictly between its two roots, that is, strictly between $0$ and $-\kappa/\sigma_{12}^2$. Because $\kappa\ne0$, this interval is nonempty, and any $c$ in it must have $\mathrm{sign}(c)=-\mathrm{sign}(\kappa)$. For every such $c$ with $0<|c|<|\kappa|/\sigma_{12}^2$, $f(c)<0$, that is, $\Var(\hat J(c))<\Var(\hat J(0))$.

\emph{Unconstrained minimizer.} Differentiating, $f'(c)=4\kappa+8c\sigma_{12}^2$, zero at $c=-\kappa/(2\sigma_{12}^2)$. Since $f''(c)=8\sigma_{12}^2>0$, this critical point is the global minimizer of $f$, hence of $\Var(\hat J(c))=\Var(\hat J(0))+f(c)$. So the unconstrained minimizer is $c^\star=-\kappa/(2\sigma_{12}^2)$, which lies inside the interval above, being exactly halfway between the roots $0$ and $-\kappa/\sigma_{12}^2$.

\emph{Feasibility.} The family $\rho_{c}$ is a valid density matrix only when $|c|\le\sqrt{p(1-p)}$, the PSD condition stated above. Intersect the sign-determined interval $0<|c|<|\kappa|/\sigma_{12}^2$ with the feasibility bound $|c|\le\sqrt{p(1-p)}$. The result contains the range $0<|c|<c_{\max}$, where $c_{\max}=\min(|\kappa|/\sigma_{12}^2,\sqrt{p(1-p)})$. If $\sqrt{p(1-p)}$ is the smaller bound, its endpoint is also feasible. We state the strict range for simplicity. Every $c$ in this range is both strictly variance-reducing and PSD-feasible. Now consider minimizing $\Var(\hat J(c))$ over the feasible interval $|c|\le\sqrt{p(1-p)}$ rather than over all of $\R$. If $c^\star$ already satisfies $|c^\star|\le\sqrt{p(1-p)}$, it remains the constrained minimizer, since the unconstrained minimum of the convex function $f$ then already lies inside the feasible interval. Otherwise, since $f$ is a convex parabola, its minimum over the feasible interval is attained at the nearer boundary point, $\pm\sqrt{p(1-p)}$ with sign matching $c^\star$. This is exactly what is meant by clipping $c^\star$ to the PSD interval.
\end{proof}

Theorem~\ref{thm:optcoherence}'s proof needs the general-$k$ case of this decomposition. We state and prove it first. Work with real symmetric $\rho$ and $O$ for notational simplicity (the Hermitian case replaces each pair $(i,j)$, $i<j$, by its real and imaginary parts, doubling the coordinates).

\begin{proposition}[Variance decomposition, general $k$]\label{prop:vardecomp}
Let $\rho\in\mathcal D_k$ be real symmetric with diagonal entries $p_i$ and off-diagonal entries $c_{ij}$. Let the entries of a real symmetric observable $O$ be estimated by jointly distributed unbiased estimators $\{\hat o_{ii}\}_i\cup\{\hat o_{ij}\}_{i<j}$ from shared rollouts. Define the plug-in estimator
\[
\hat J(\rho)\;=\;\sum_i p_i\,\hat o_{ii}\;+\;2\sum_{i<j}c_{ij}\,\hat o_{ij}.
\]
Then $\hat J(\rho)$ is unbiased for $J(\rho)=\Tr(\rho O)$. Writing $\hat J_{\mathrm{diag}}=\sum_ip_i\hat o_{ii}$, $\gamma_{ij}=\Cov(\hat J_{\mathrm{diag}},\hat o_{ij})$, $\sigma_{ij}^2=\Var(\hat o_{ij})$, and $\Gamma_{(ij),(i'j')}=\Cov(\hat o_{ij},\hat o_{i'j'})$ for distinct pairs (the off-diagonal entries of the matrix $\Gamma$ of Theorem~\ref{thm:optcoherence}), its variance decomposes as
\[
\Var\bigl(\hat J(\rho)\bigr)
=\Var\bigl(\hat J_{\mathrm{diag}}\bigr)
+4\sum_{i<j}c_{ij}\,\gamma_{ij}
+4\sum_{i<j}c_{ij}^2\,\sigma_{ij}^2
+8\!\!\sum_{\substack{(i<j)\,\prec\,(i'<j')}}\!\!c_{ij}c_{i'j'}\,\Gamma_{(ij),(i'j')},
\]
where the last sum runs over unordered pairs of distinct off-diagonal index pairs and $\prec$ denotes any fixed total order on the pairs. In particular, $\nabla_{c_{ij}}\Var\bigl(\hat J(\rho)\bigr)\big|_{\mathbf{c}_{\mathrm{off}}=0}=4\gamma_{ij}$. Suppose in addition that $p_i>0$ for every $i$, and that $\gamma_{ij}\ne0$ for some pair. Then there are coherences $c_{ij}$ of arbitrarily small magnitude, with $\mathrm{sign}(c_{ij})=-\mathrm{sign}(\gamma_{ij})$ and all other coherences zero, such that $\rho$ is PSD and the variance is strictly below that of the diagonal ensemble with the same action probabilities.
\end{proposition}

The positivity of the $p_i$ is needed only for the last claim. If $p_i=0$, then every PSD $\rho$ with diagonal $\mathbf{p}$ has $c_{ij}=0$ for all $j$, because the $2\times2$ minor on $\{i,j\}$ equals $-c_{ij}^2$. This is the same obstruction as in Corollary~\ref{cor:feasibility}.
\begin{proof}
\emph{Unbiasedness.} By linearity of expectation, $\E[\hat J(\rho)]=\sum_ip_i\E[\hat o_{ii}]+2\sum_{i<j}c_{ij}\E[\hat o_{ij}]=\sum_ip_iO_{ii}+2\sum_{i<j}c_{ij}O_{ij}$, using unbiasedness of each estimator. This equals $J(\rho)=\Tr(\rho O)$, since for a real symmetric matrix, $\Tr(\rho O)=\sum_i\rho_{ii}O_{ii}+2\sum_{i<j}\rho_{ij}O_{ij}$: the pair $(i,j)$ and $(j,i)$ contribute the same product $\rho_{ij}O_{ij}$ twice, by symmetry of both matrices.

\emph{Setting up the quadratic form.} Collect the estimators into a random vector $\hat{\mathbf o}=(\hat o_{ii})_i\cup(\hat o_{ij})_{i<j}$, and the coefficients into $\boldsymbol{\psi}=(p_i)_i\cup(2c_{ij})_{i<j}$, so $\hat J(\rho)=\boldsymbol{\psi}^\top\hat{\mathbf o}$ and $\Var(\hat J(\rho))=\boldsymbol{\psi}^\top\Sigma \boldsymbol{\psi}$, where $\Sigma=\Cov(\hat{\mathbf o})$.

\emph{Expanding the quadratic form.} Write $\boldsymbol{\psi}^\top\Sigma \boldsymbol{\psi}=\sum_{\ell,\ell'}\psi_\ell \psi_{\ell'}\Sigma_{\ell\ell'}$ and split the index set into the diagonal block $\{ii\}$ and the off-diagonal block $\{ij:i<j\}$. The sum breaks into four groups, according to whether each of the two indices $\ell,\ell'$ falls in the diagonal or the off-diagonal block.

Group 1 (diagonal-diagonal): both indices range over diagonal entries, with coefficients $p_i$ (not doubled). This is exactly $\Var(\hat J_{\mathrm{diag}})$, since $\hat J_{\mathrm{diag}}=\sum_ip_i\hat o_{ii}$ has this quadratic form as its variance.

Group 2 (diagonal-off-diagonal cross terms): a pair with one diagonal index and one off-diagonal index $ij$ appears twice in the double sum, by symmetry of $\Sigma$. Each appearance carries coefficient $p_i\cdot2c_{ij}$, summed over the diagonal indices making up $\hat J_{\mathrm{diag}}$. Using $\Cov(\hat J_{\mathrm{diag}},\hat o_{ij})=\gamma_{ij}$, this group totals $2\cdot2\sum_{i<j}c_{ij}\gamma_{ij}=4\sum_{i<j}c_{ij}\gamma_{ij}$.

Group 3 (off-diagonal variances): each off-diagonal index $ij$ paired with itself contributes
\[
(2c_{ij})^2\Var(\hat o_{ij})=4c_{ij}^2\sigma_{ij}^2 ,
\]
totaling $4\sum_{i<j}c_{ij}^2\sigma_{ij}^2$.

Group 4 (distinct off-diagonal pairs): an unordered pair of distinct off-diagonal indices $\{ij,i'j'\}$ appears twice in the double sum, each time with coefficient $2c_{ij}\cdot2c_{i'j'}$, contributing $2\cdot4c_{ij}c_{i'j'}\Gamma_{(ij),(i'j')}=8c_{ij}c_{i'j'}\Gamma_{(ij),(i'j')}$ per unordered pair, totaling $8\sum_{(ij)\prec(i'j')}c_{ij}c_{i'j'}\Gamma_{(ij),(i'j')}$.

Adding the four groups gives exactly the displayed identity.

\emph{The gradient claim.} Differentiating this quadratic in $c_{ij}$ and evaluating at $\mathbf{c}_{\mathrm{off}}=0$, every term with a factor of $c_{i'j'}$ (Groups 3 and 4) vanishes, since those terms are at least quadratic in the off-diagonal coefficients. Only the linear Group 2 term survives, giving $\nabla_{c_{ij}}\Var(\hat J(\rho))\big|_{\mathbf{c}_{\mathrm{off}}=0}=4\gamma_{ij}$.

\emph{Feasibility of small perturbations.} If $\gamma_{ij}\ne0$ for some pair, choosing $c_{ij}$ with $\mathrm{sign}(c_{ij})=-\mathrm{sign}(\gamma_{ij})$ and small enough magnitude makes the linear term $4c_{ij}\gamma_{ij}<0$. This linear term dominates the quadratic terms of Groups 3 and 4 for sufficiently small $\|\mathbf{c}_{\mathrm{off}}\|$, since those shrink faster as the coefficients shrink. It remains to check that such a small perturbation keeps $\rho$ PSD. Here we use the extra assumption $p_i>0$ for every $i$. Then $\rho_{\mathrm{diag}}=\diag(\mathbf{p})$ is positive definite, so its smallest eigenvalue is strictly positive. The eigenvalues of a Hermitian matrix depend continuously on its entries. So $\diag(\mathbf{p})+C$ stays positive definite for every hollow symmetric $C$ of small enough norm. In particular, the choice above, with a single small nonzero coherence $c_{ij}$, gives a valid density matrix with diagonal $\mathbf{p}$.
\end{proof}

\begin{proof}[Proof of Theorem~\ref{thm:optcoherence}]
\emph{The first display.} This is Proposition~\ref{prop:vardecomp} written in vector notation. The linear term $4\sum_{i<j}c_{ij}\gamma_{ij}$ becomes $4\mathbf{c}^\top\boldsymbol{\gamma}$. The sum $4\sum_{i<j}c_{ij}^2\sigma_{ij}^2$ becomes the diagonal part of $4\mathbf{c}^\top\Gamma \mathbf{c}$, using $\Gamma_{(ij),(ij)}=\sigma_{ij}^2$. The cross-pair sum $8\sum_{(ij)\prec(i'j')}c_{ij}c_{i'j'}\Gamma_{(ij),(i'j')}$ becomes the remaining part of $4\mathbf{c}^\top\Gamma \mathbf{c}$, whose coefficients $\Gamma_{(ij),(i'j')}$ are the off-diagonal entries of $\Gamma$, and the fact that each unordered pair contributes twice to the quadratic form $\mathbf{c}^\top\Gamma \mathbf{c}$.

\emph{Step 1, $\boldsymbol{\gamma}$ lies in the range of $\Gamma$.} Let $\mathbf{z}\in\ker\Gamma$, so $\Gamma \mathbf{z}=0$. Writing $\hat{\mathbf{o}}=(\hat o_{ij})_{i<j}$, $\Var(\mathbf{z}^\top\hat{\mathbf{o}})=\mathbf{z}^\top\Cov(\hat{\mathbf{o}})\mathbf{z}=\mathbf{z}^\top\Gamma \mathbf{z}=0$. A random variable with zero variance is almost surely equal to its own mean, so $\mathbf{z}^\top\hat{\mathbf{o}}$ is almost surely constant. A constant has zero covariance with any other random variable, so $\mathbf{z}^\top\boldsymbol{\gamma}=\Cov(\hat J_{\mathrm{diag}},\mathbf{z}^\top\hat{\mathbf{o}})=0$. This holds for every $\mathbf{z}\in\ker\Gamma$, so $\boldsymbol{\gamma}$ is orthogonal to $\ker\Gamma$. Since $\ker\Gamma$ and $\mathrm{range}(\Gamma)$ are orthogonal complements for a symmetric matrix, $\boldsymbol{\gamma}\in\mathrm{range}(\Gamma)$. Because $\Gamma\Gamma^{+}$ is the orthogonal projector onto $\mathrm{range}(\Gamma)$, a standard property of the Moore--Penrose pseudoinverse, this gives $\Gamma\Gamma^{+}\boldsymbol{\gamma}=\boldsymbol{\gamma}$.

\emph{Step 2, completing the square.} Set $\mathbf{c}^\star=-\tfrac12\Gamma^{+}\boldsymbol{\gamma}$ and expand
\[
4(\mathbf{c}-\mathbf{c}^\star)^\top\Gamma(\mathbf{c}-\mathbf{c}^\star)=4\mathbf{c}^\top\Gamma \mathbf{c}-8\mathbf{c}^\top\Gamma \mathbf{c}^\star+4\mathbf{c}^{\star\top}\Gamma \mathbf{c}^\star .
\]
Substituting $\mathbf{c}^\star=-\tfrac12\Gamma^{+}\boldsymbol{\gamma}$ into the middle term, $-8\mathbf{c}^\top\Gamma \mathbf{c}^\star=4\mathbf{c}^\top\Gamma\Gamma^{+}\boldsymbol{\gamma}=4\mathbf{c}^\top\boldsymbol{\gamma}$, using Step 1. Substituting into the last term, $4\mathbf{c}^{\star\top}\Gamma \mathbf{c}^\star=\boldsymbol{\gamma}^\top\Gamma^{+}\Gamma\Gamma^{+}\boldsymbol{\gamma}=\boldsymbol{\gamma}^\top\Gamma^{+}\boldsymbol{\gamma}$, using the pseudoinverse identity $\Gamma^{+}\Gamma\Gamma^{+}=\Gamma^{+}$. So
\[
4(\mathbf{c}-\mathbf{c}^\star)^\top\Gamma(\mathbf{c}-\mathbf{c}^\star)=4\mathbf{c}^\top\Gamma \mathbf{c}+4\mathbf{c}^\top\boldsymbol{\gamma}+\boldsymbol{\gamma}^\top\Gamma^{+}\boldsymbol{\gamma} .
\]
Rearranging and adding $\Var(\hat J_{\mathrm{diag}})$ to both sides recovers the second display,
\[
\Var\bigl(\hat J(\mathbf{c})\bigr)=\Var\bigl(\hat J_{\mathrm{diag}}\bigr)-\boldsymbol{\gamma}^\top\Gamma^{+}\boldsymbol{\gamma}+4\bigl(\mathbf{c}-\mathbf{c}^\star\bigr)^\top\Gamma\bigl(\mathbf{c}-\mathbf{c}^\star\bigr) .
\]

\emph{Step 3, minimizers.} Since $\Gamma\succeq0$, the last term is nonnegative for every $\mathbf{c}$, with equality exactly when $\mathbf{c}-\mathbf{c}^\star\in\ker\Gamma$, since a PSD quadratic form vanishes exactly on the kernel of the matrix. So $\Var(\hat J(\mathbf{c}))$ is minimized exactly on the affine set $\mathbf{c}^\star+\ker\Gamma$, at the value $\Var(\hat J_{\mathrm{diag}})-\boldsymbol{\gamma}^\top\Gamma^{+}\boldsymbol{\gamma}$.

\emph{Step 4, sign and size of the reduction, and the $\eta^2$ identity.} Since $\boldsymbol{\gamma}\in\mathrm{range}(\Gamma)$ (Step 1) and $\Gamma^{+}$ acts as the inverse of $\Gamma$ on $\mathrm{range}(\Gamma)$, the quantity $\boldsymbol{\gamma}^\top\Gamma^{+}\boldsymbol{\gamma}$ is nonnegative, because $\Gamma^{+}$ is itself PSD. It is strictly positive whenever $\boldsymbol{\gamma}\ne0$, since $\boldsymbol{\gamma}^\top\Gamma^{+}\boldsymbol{\gamma}=0$ together with $\boldsymbol{\gamma}\in\mathrm{range}(\Gamma)$ forces $\boldsymbol{\gamma}=0$. The minimum variance in Step 3 cannot be negative. So the reduction $\boldsymbol{\gamma}^\top\Gamma^{+}\boldsymbol{\gamma}$ is at most $\Var(\hat J_{\mathrm{diag}})$. Hence the ratio
\[
\eta^2=\frac{\boldsymbol{\gamma}^\top\Gamma^{+}\boldsymbol{\gamma}}{\Var(\hat J_{\mathrm{diag}})}
\]
lies in $[0,1]$ whenever $\Var(\hat J_{\mathrm{diag}})>0$. If $\Var(\hat J_{\mathrm{diag}})=0$, then $\boldsymbol{\gamma}=0$, and we set $\eta^2=0$. In both cases \eqref{eq:r2reduction} holds. This ratio is exactly the coefficient of determination from linear regression. Here $\boldsymbol{\gamma}^\top\Gamma^{+}\boldsymbol{\gamma}$ is the variance of the best linear predictor of $\hat J_{\mathrm{diag}}$ from the covariates $(\hat o_{ij})_{i<j}$, that is, the projection of $\hat J_{\mathrm{diag}}$ onto their span in the covariance inner product. Dividing by $\Var(\hat J_{\mathrm{diag}})$ then gives the fraction of variance in $\hat J_{\mathrm{diag}}$ explained by the off-diagonal statistics.
\end{proof}

\begin{proof}[Proof of Corollary~\ref{cor:feasibility}]
\emph{Existence of $\alpha_{\max}$.} At $\alpha=0$, $\rho_{\alpha \mathbf{c}^\star}=\rho_{\mathrm{diag}}$, whose eigenvalues equal its diagonal entries, all strictly positive by assumption. The eigenvalues of a Hermitian matrix vary continuously with its entries, and the entries of $\rho_{\alpha \mathbf{c}^\star}$ vary linearly in $\alpha$, since only the off-diagonal entries change. So the smallest eigenvalue of $\rho_{\alpha \mathbf{c}^\star}$ stays strictly positive, hence $\rho_{\alpha \mathbf{c}^\star}\succeq0$, for all $\alpha$ in some interval $[0,\alpha_0]$ with $\alpha_0>0$. The PSD cone is convex, so the set of feasible $\alpha\ge0$ is an interval $[0,\alpha_{\max}]$ with $\alpha_{\max}\ge\alpha_0$. The closed form below computes $\alpha_{\max}$.

\emph{The variance along this segment.} Substitute $\mathbf{c}=\alpha \mathbf{c}^\star$ into $\Var(\hat J(\mathbf{c}))=\Var(\hat J_{\mathrm{diag}})+4\mathbf{c}^\top\boldsymbol{\gamma}+4\mathbf{c}^\top\Gamma \mathbf{c}$ from Theorem~\ref{thm:optcoherence}. The linear term becomes
\[
4(\alpha \mathbf{c}^\star)^\top\boldsymbol{\gamma}=4\alpha\Bigl(-\tfrac12\Gamma^{+}\boldsymbol{\gamma}\Bigr)^\top\boldsymbol{\gamma}=-2\alpha\,\boldsymbol{\gamma}^\top\Gamma^{+}\boldsymbol{\gamma} ,
\]
using $\mathbf{c}^\star=-\tfrac12\Gamma^{+}\boldsymbol{\gamma}$ and symmetry of $\Gamma^{+}$. The quadratic term becomes
\[
4(\alpha \mathbf{c}^\star)^\top\Gamma(\alpha \mathbf{c}^\star)=\alpha^2\boldsymbol{\gamma}^\top\Gamma^{+}\Gamma\Gamma^{+}\boldsymbol{\gamma}=\alpha^2\boldsymbol{\gamma}^\top\Gamma^{+}\boldsymbol{\gamma} ,
\]
using $\Gamma^{+}\Gamma\Gamma^{+}=\Gamma^{+}$. Adding these,
\[
4(\alpha \mathbf{c}^\star)^\top\boldsymbol{\gamma}+4(\alpha \mathbf{c}^\star)^\top\Gamma(\alpha \mathbf{c}^\star)=-\alpha(2-\alpha)\,\boldsymbol{\gamma}^\top\Gamma^{+}\boldsymbol{\gamma} .
\]
For $\alpha\in(0,2)$, $\alpha(2-\alpha)>0$, so this is strictly negative whenever $\boldsymbol{\gamma}\ne0$. Adding $\Var(\hat J_{\mathrm{diag}})$ to both sides and using $\boldsymbol{\gamma}^\top\Gamma^{+}\boldsymbol{\gamma}=\eta^2\Var(\hat J_{\mathrm{diag}})$ from \eqref{eq:r2reduction} gives the stated formula, $\Var(\hat J(\alpha \mathbf{c}^\star))=\Var(\hat J_{\mathrm{diag}})[1-\alpha(2-\alpha)\eta^2]$.

\emph{Monotonicity and the choice of $\alpha$.} The factor $\alpha(2-\alpha)=1-(1-\alpha)^2$ is increasing on $[0,1]$ and decreasing beyond, so its maximum on $[0,\alpha_{\max}]$ occurs at $\alpha=\min(\alpha_{\max},1)$. This choice maximizes the guaranteed reduction achievable along the segment $\{\alpha \mathbf{c}^\star\}$: the full reduction $\eta^2$ when $\alpha_{\max}\ge1$, and the smaller amount $\alpha_{\max}(2-\alpha_{\max})\eta^2$ when $\alpha_{\max}<1$.

\emph{Closed form for $\alpha_{\max}$.} Write $\rho_{\alpha \mathbf{c}^\star}=\rho_{\mathrm{diag}}+\alpha C$. Here $C$ is the hollow Hermitian matrix built from $\mathbf{c}^\star$'s off-diagonal entries. Since $\rho_{\mathrm{diag}}\succ0$, factor it out on both sides:
\[
\rho_{\mathrm{diag}}+\alpha C=\rho_{\mathrm{diag}}^{1/2}\bigl(I+\alpha K\bigr)\rho_{\mathrm{diag}}^{1/2},\qquad K:=\rho_{\mathrm{diag}}^{-1/2}C\rho_{\mathrm{diag}}^{-1/2}.
\]
$K$ is Hermitian and $\rho_{\mathrm{diag}}^{1/2}$ is invertible, so this is a congruence transformation. Sylvester's law of inertia says congruence preserves the sign pattern of eigenvalues. So $\rho_{\mathrm{diag}}+\alpha C\succeq0$ exactly when $I+\alpha K\succeq0$. Since $K$ is Hermitian, it has real eigenvalues $\lambda_1(K)\le\dots\le\lambda_k(K)$. In the same eigenbasis, $I+\alpha K$ has eigenvalues $1+\alpha\lambda_i(K)$. This matrix is PSD exactly when $1+\alpha\lambda_i(K)\ge0$ for every $i$. For $\alpha\ge0$, the tightest constraint comes from the smallest eigenvalue $\lambda_1(K)=\lambda_{\min}(K)$. If $\lambda_{\min}(K)\ge0$, every constraint holds for all $\alpha\ge0$. There is then no upper bound. If $\lambda_{\min}(K)<0$, the constraint requires $\alpha\le-1/\lambda_{\min}(K)$.
Since $-\lambda_{\min}(K)=\lambda_{\max}(-K)$, both cases combine into
\[
\alpha_{\max}=1/\max\bigl[0,\lambda_{\max}(-\rho_{\mathrm{diag}}^{-1/2}C\rho_{\mathrm{diag}}^{-1/2})\bigr].
\]
The $\max$ with $0$ covers the unconstrained case. Dividing by $0$ there signals no upper bound. This case needs $\lambda_{\min}(K)\ge0$. But $K$ has zero diagonal, so $\Tr K=0$. A nonzero $K$ with zero trace has a negative eigenvalue. Hence $\alpha_{\max}$ is finite whenever $\mathbf{c}^\star\ne0$, and unbounded only when $\mathbf{c}^\star=0$.
\end{proof}

\begin{proof}[Proof of Corollary~\ref{cor:adaptive}]
Throughout, write $\|\mathbf{z}\|_\Gamma^2=\mathbf{z}^\top\Gamma \mathbf{z}$. Define the scaling map $\varphi(\mathbf{c})=\alpha(\mathbf{c})\,\mathbf{c}$ with $\alpha(\mathbf{c})=\min\bigl(1,\,1/\max[0,\lambda_{\max}(-K(\mathbf{c}))]\bigr)$, where $K(\mathbf{c})=\rho_{\mathrm{diag}}^{-1/2}C\rho_{\mathrm{diag}}^{-1/2}$ and $C$ is the hollow matrix built from $\mathbf{c}$. By the proof of Corollary~\ref{cor:feasibility}, $\alpha(\mathbf{c})=\min(\alpha_{\max}(\mathbf{c}),1)$. So $\tilde{\mathbf{c}}=\varphi(\hat{\mathbf{c}})$ and $\mathbf{c}_{\mathrm{feas}}=\varphi(\mathbf{c}^\star)$.

\emph{Step 1, conditional unbiasedness and variance.} Condition on the pilot batch, so $\tilde{\mathbf{c}}$ is fixed. Because the pilot batch and the evaluation batch are independent, the evaluation-batch statistics used to form $\hat J(\tilde{\mathbf{c}})$ keep their original unconditional means and covariances $\boldsymbol{\gamma},\Gamma$ even after conditioning. So Theorem~\ref{thm:optcoherence} applies directly with the fixed coefficient $\mathbf{c}=\tilde{\mathbf{c}}$. This gives conditional unbiasedness, $\E[\hat J(\tilde{\mathbf{c}})\mid\tilde{\mathbf{c}}]=J(\rho_{\tilde{\mathbf{c}}})$, and the conditional variance identity~\eqref{eq:condvar}, which is the second display of Theorem~\ref{thm:optcoherence} evaluated at $\mathbf{c}=\tilde{\mathbf{c}}$.

\emph{Step 2, the ridge bias.} Let $\Gamma=\sum_{\lambda>0}\lambda\,\mathbf{u}\mathbf{u}^\top$ be the spectral decomposition on its range. We know $\boldsymbol{\gamma}\in\mathrm{range}(\Gamma)$ from the proof of Theorem~\ref{thm:optcoherence}. So $\mathbf{c}^\star=-\tfrac12\sum_\lambda\lambda^{-1}(\mathbf{u}^\top\boldsymbol{\gamma})\,\mathbf{u}$ and $\mathbf{c}_\xi=-\tfrac12\sum_\lambda(\lambda+\xi)^{-1}(\mathbf{u}^\top\boldsymbol{\gamma})\,\mathbf{u}$, with no component in $\ker\Gamma$. Subtracting, and using $\lambda^{-1}-(\lambda+\xi)^{-1}=\xi/(\lambda(\lambda+\xi))$,
\[
4\|\mathbf{c}_\xi-\mathbf{c}^\star\|_\Gamma^2=\sum_\lambda\frac{\xi^2}{\lambda(\lambda+\xi)^2}(\mathbf{u}^\top\boldsymbol{\gamma})^2\ \le\ \xi^2\boldsymbol{\gamma}^\top(\Gamma^{+})^3\boldsymbol{\gamma} .
\]
In particular $\mathbf{c}_\xi\to \mathbf{c}^\star$ as $\xi\to0$.

\emph{Step 3, sampling error at a fixed ridge.} A sample covariance is PSD, so $\hat\Gamma+\xi I\succeq\xi I$ and $\|(\hat\Gamma+\xi I)^{-1}\|\le1/\xi$. The same bound holds for $\Gamma+\xi I$. A direct check shows
\[
\hat{\mathbf{c}}-\mathbf{c}_\xi=-\tfrac12(\hat\Gamma+\xi I)^{-1}\bigl[(\hat{\boldsymbol{\gamma}}-\boldsymbol{\gamma})-(\hat\Gamma-\Gamma)(\Gamma+\xi I)^{-1}\boldsymbol{\gamma}\bigr] .
\]
Since $\boldsymbol{\gamma}\in\mathrm{range}(\Gamma)$, we have $\|(\Gamma+\xi I)^{-1}\boldsymbol{\gamma}\|\le\|\Gamma^{+}\boldsymbol{\gamma}\|=2\|\mathbf{c}^\star\|$. Hence
\[
\|\hat{\mathbf{c}}-\mathbf{c}_\xi\|\ \le\ \frac1{2\xi}\Bigl(\|\hat{\boldsymbol{\gamma}}-\boldsymbol{\gamma}\|+2\|\mathbf{c}^\star\|\,\|\hat\Gamma-\Gamma\|\Bigr) .
\]
Every entry of $\hat{\boldsymbol{\gamma}}-\boldsymbol{\gamma}$ and $\hat\Gamma-\Gamma$ is an average of $m$ independent centered terms, up to an $\mathcal O(1/m)$ bias from centering at the sample mean. Each term has finite variance because the statistics have finite fourth moments. So $\E\|\hat{\boldsymbol{\gamma}}-\boldsymbol{\gamma}\|^2$ and $\E\|\hat\Gamma-\Gamma\|^2$ are both $\mathcal O(1/m)$. Therefore $\E\|\hat{\mathbf{c}}-\mathbf{c}_\xi\|^2\le C_0/(\xi^2m)$ for a constant $C_0$ that does not depend on $m$ or $\xi$. For fixed $\xi$ this is $\mathcal O(1/m)$.

\emph{Step 4, the scaling.} The map $t\mapsto\min(1,1/\max[0,t])$ is $1$-Lipschitz, and $\lambda_{\max}(-K(\mathbf{c}))$ is Lipschitz in $\mathbf{c}$, because $K$ is linear in $\mathbf{c}$. So $\alpha$ is Lipschitz with some constant $L_\alpha$, and $\alpha\in[0,1]$. Writing $\varphi(\hat{\mathbf{c}})-\varphi(\mathbf{c}_\xi)=\alpha(\hat{\mathbf{c}})(\hat{\mathbf{c}}-\mathbf{c}_\xi)+(\alpha(\hat{\mathbf{c}})-\alpha(\mathbf{c}_\xi))\,\mathbf{c}_\xi$ gives
\[
\|\tilde{\mathbf{c}}-\varphi(\mathbf{c}_\xi)\|\ \le\ \bigl(1+L_\alpha\|\mathbf{c}_\xi\|\bigr)\,\|\hat{\mathbf{c}}-\mathbf{c}_\xi\| .
\]
Also, $\varphi$ is continuous, so $\varphi(\mathbf{c}_\xi)\to\varphi(\mathbf{c}^\star)=\mathbf{c}_{\mathrm{feas}}$ as $\xi\to0$.

\emph{Step 5, the limit.} Take $\xi=\xi_m$ with $\xi_m\to0$ and $\xi_m\sqrt m\to\infty$. Then $\|\mathbf{c}_{\xi_m}\|$ stays bounded, and Steps 3 and 4 give $\E\|\tilde{\mathbf{c}}-\varphi(\mathbf{c}_{\xi_m})\|^2\le(1+L_\alpha\|\mathbf{c}_{\xi_m}\|)^2C_0/(\xi_m^2m)\to0$. Together with $\varphi(\mathbf{c}_{\xi_m})\to \mathbf{c}_{\mathrm{feas}}$, this shows $\tilde{\mathbf{c}}\to \mathbf{c}_{\mathrm{feas}}$ in mean square. The right side of~\eqref{eq:condvar} is a continuous function of $\tilde{\mathbf{c}}$. It is also bounded, because every feasible $\tilde{\mathbf{c}}$ satisfies $|\tilde c_{ij}|\le\sqrt{\rho_{ii}\rho_{jj}}$. So its expectation converges to its value at $\mathbf{c}_{\mathrm{feas}}=\alpha_{\mathrm{feas}} \mathbf{c}^\star$. At that point, $4\|\mathbf{c}_{\mathrm{feas}}-\mathbf{c}^\star\|_\Gamma^2=(1-\alpha_{\mathrm{feas}})^2\,4\|\mathbf{c}^\star\|_\Gamma^2=(1-\alpha_{\mathrm{feas}})^2\boldsymbol{\gamma}^\top\Gamma^{+}\boldsymbol{\gamma}$, and
\[
\Var(\hat J_{\mathrm{diag}})-\boldsymbol{\gamma}^\top\Gamma^{+}\boldsymbol{\gamma}+(1-\alpha_{\mathrm{feas}})^2\boldsymbol{\gamma}^\top\Gamma^{+}\boldsymbol{\gamma}=\Var(\hat J_{\mathrm{diag}})\bigl[1-\alpha_{\mathrm{feas}}(2-\alpha_{\mathrm{feas}})\eta^2\bigr] ,
\]
which is~\eqref{eq:adaptivelimit}. When $\alpha_{\max}(\mathbf{c}^\star)\ge1$, we have $\alpha_{\mathrm{feas}}=1$ and the limit is $\Var(\hat J_{\mathrm{diag}})(1-\eta^2)$. If no scaling is applied, $\varphi$ is the identity and $\mathbf{c}_{\mathrm{feas}}=\mathbf{c}^\star$. The same argument then holds. The excess term is at most $4\|\Gamma\|\,\|\tilde{\mathbf{c}}-\mathbf{c}^\star\|^2$, so its expectation also tends to zero.

\emph{Step 6, the unregularized rate.} On a linearly independent subset of the covariates, $\Gamma\succ0$ and $\mathbf{c}^\star=-\tfrac12\Gamma^{-1}\boldsymbol{\gamma}$. By the central limit theorem, $\sqrt m(\hat{\boldsymbol{\gamma}}-\boldsymbol{\gamma},\hat\Gamma-\Gamma)$ converges in distribution to a Gaussian, again using finite fourth moments. The map $(\boldsymbol{\gamma},\Gamma)\mapsto-\tfrac12\Gamma^{-1}\boldsymbol{\gamma}$ is smooth near $(\boldsymbol{\gamma},\Gamma)$. So the delta method shows that $\sqrt m(\hat{\mathbf{c}}-\mathbf{c}^\star)$ converges in distribution, and hence stays bounded in probability. If $\hat{\mathbf{c}}$ is already feasible, $\tilde{\mathbf{c}}=\hat{\mathbf{c}}$. The excess term in~\eqref{eq:condvar} is then $4(\hat{\mathbf{c}}-\mathbf{c}^\star)^\top\Gamma(\hat{\mathbf{c}}-\mathbf{c}^\star)$. Since $\Gamma$ is fixed, $m$ times this term also stays bounded in probability. This is a statement in probability. We do not bound the expectation, because $\hat\Gamma^{-1}$ may have heavy moments.

\emph{Step 7, the diagonal-observable case.} When $O$ is diagonal, $J(\rho_{\mathbf{c}})=\sum_ip_iO_{ii}$ depends only on the fixed diagonal, not on $\mathbf{c}$ (Lemma~\ref{lem:phaseirrelevance}). So $\E[\hat J(\tilde{\mathbf{c}})\mid\tilde{\mathbf{c}}]=J(\rho_{\tilde{\mathbf{c}}})=J$ is the same constant regardless of the realized $\tilde{\mathbf{c}}$. By the law of total variance,
\[
\Var\bigl(\hat J(\tilde{\mathbf{c}})\bigr)=\E\bigl[\Var(\hat J(\tilde{\mathbf{c}})\mid\tilde{\mathbf{c}})\bigr]+\Var\bigl(\E[\hat J(\tilde{\mathbf{c}})\mid\tilde{\mathbf{c}}]\bigr) .
\]
Since $\E[\hat J(\tilde{\mathbf{c}})\mid\tilde{\mathbf{c}}]=J$ is constant, the second term vanishes. Taking the expectation of~\eqref{eq:condvar} over the pilot batch, and noting that $\Var(\hat J_{\mathrm{diag}})-\boldsymbol{\gamma}^\top\Gamma^{+}\boldsymbol{\gamma}$ does not depend on the pilot batch, gives
\[
\Var\bigl(\hat J(\tilde{\mathbf{c}})\bigr)=\Var\bigl(\hat J_{\mathrm{diag}}\bigr)-\boldsymbol{\gamma}^\top\Gamma^{+}\boldsymbol{\gamma}+4\,\E\bigl[(\tilde{\mathbf{c}}-\mathbf{c}^\star)^\top\Gamma(\tilde{\mathbf{c}}-\mathbf{c}^\star)\bigr] .
\]
This is~\eqref{eq:condvar} with the random excess term replaced by its expectation. By Step 5, the adaptive procedure attains the guaranteed reduction of Corollary~\ref{cor:feasibility} up to a penalty that vanishes as the pilot batch grows.
\end{proof}

\section{Algorithms}\label{app:algorithms}

This appendix gives pseudocode for the four algorithms used in the paper. Each comes with the short prose needed to read it alongside the main text.

\emph{DMPG} (density-matrix policy gradient) ascends the objective $J(\theta)=\E_{s}\bigl[\Tr(\rho_s(\theta)Q_s)\bigr]$. Here $\rho_s=\rho(U_\theta(s))$, where $U_\theta(s)$ is a parametric factor such as the output of a neural network. The state $s$ follows the state-visitation distribution under $\pi$. As in the policy gradient theorem \cite{sutton2000policy}, each update treats $Q_s$ and the visitation distribution as fixed. With on-policy sampling and exact values, this gives the true gradient in expectation.

Let $Q_s$ be the diagonal matrix of value or advantage estimates $\hat Q(s,a_i)$ for all actions. Setting $O=Q_s$ in Theorem~\ref{thm:gradient} gives the gradient at each state. As in that theorem, $h=\Tr(UU^\dagger)$ is the normalizer of $\rho(U)$. For real $U$, the gradient is
\[
\nabla_UJ_s=\tfrac{2}{h}(Q_s-J_sI)U,\qquad J_s=\sum_ip_i\hat Q(s,a_i).
\]
Here $p_i=(UU^\top)_{ii}/h$ is the action probability. Since $Q_s$ is diagonal, $J_s$ depends on $U$ only through $\mathbf{p}$. So DMPG is an expected-action policy gradient under this row-norm parameterization, with $Q_s$ held fixed. For $r=1$ it is the square-root parameterization $p_i=u_i^2/\|\mathbf{u}\|^2$.

The chain rule carries this gradient to the parameters $\theta$. In the complex case, $\partial J_s/\partial\theta_\ell=2\,\re\Tr\bigl((\nabla_{\overline{U}}J_s)^\dagger\,\partial U_\theta(s)/\partial\theta_\ell\bigr)$, and backpropagation through $U_\theta$ computes it. For real $U$, this is the ordinary chain rule applied to the real gradient above. Section~\ref{sec:limitations} gives the per-state costs. For a diagonal $Q_s$, the cost is a factor of $r$ above the $\mathcal O(k)$ of a softmax over logits. So for a diagonal reward, DMPG has no computational advantage over an ordinary softmax policy. Its value is that the same gradient formula applies unchanged when $Q_s$ is non-diagonal.

The entropy regularizer $\tau S(\rho_s)$ is optional. Theorem~\ref{thm:gradient} does not cover it, because $S$ is not linear in $\rho$. For full-rank $\rho_s$, its gradient is
\[
\nabla_{\overline{U}}S(\rho_s)=-\tfrac1h\bigl(\log\rho_s+S(\rho_s)I\bigr)U.
\]
This is the formula of Theorem~\ref{thm:gradient} with $O$ replaced by $-\log\rho_s$, evaluated at the current $\rho_s$. If $\rho_s$ is rank-deficient, the logarithm is taken on the support of $\rho_s$, which contains the columns of $U$.

\begin{algorithm}[H]
\caption{Density-Matrix Policy Gradient (DMPG)}\label{alg:dmpg}
\begin{algorithmic}[1]
\STATE Initialize a parametric factor map $U_\theta(s)\in\C^{k\times r}$. Choose a step size $\alpha_{\mathrm{step}}>0$.
\FOR{iteration $t=1,\dots,T$}
  \STATE Collect trajectories by sampling actions from the diagonal of $\rho_s=U_\theta(s)U_\theta(s)^\dagger/\Tr\bigl(U_\theta(s)U_\theta(s)^\dagger\bigr)$.
  \STATE Estimate advantages $\hat A(s,a)$ (critic or Monte Carlo) and form $Q_s$ (or the advantage operator $A_s^{(t)}$).
  \STATE Compute $\nabla_\theta J(\theta)$ via Theorem~\ref{thm:gradient} and backpropagation through $U_\theta(s)$. Optionally add the gradient of the entropy regularizer $\tau\,S(\rho_s)$ of Appendix~\ref{sec:entropy}, given above.
  \STATE Update $\theta\leftarrow\theta+\alpha_{\mathrm{step}}\,\nabla_\theta J(\theta)$.
\ENDFOR
\end{algorithmic}
\end{algorithm}

\emph{DMPI} (density-matrix policy iteration) alternates evaluation with the improvement map of Definition~\ref{def:softmaxchannel}. Its per-state improvement guarantee (Theorem~\ref{thm:monotonic}) is exact when $Q^{(t)}$ is computed exactly. With simulated estimates of $Q^{(t)}$, it holds only approximately.

\begin{algorithm}[H]
\caption{Density-Matrix Policy Iteration (DMPI)}\label{alg:dmpi}
\begin{algorithmic}[1]
\STATE Initialize $\rho_s^{(0)}$ (e.g., a strictly positive diagonal or $I/k$).
\FOR{$t=0,1,\dots$ until convergence}
  \STATE \emph{Evaluation:} estimate $Q^{(t)}(s,a)$ (dynamic programming or simulation) and set $V^{(t)}(s)=\Tr\bigl(\rho_s^{(t)}Q_s^{(t)}\bigr)$.
  \STATE \emph{Improvement:} set $\rho_s^{(t+1)}=\mathcal T_{Q_s^{(t)}}^{\beta}\bigl(\rho_s^{(t)}\bigr)$ (Definition~\ref{def:softmaxchannel}). For exact $Q_s^{(t)}$, Theorem~\ref{thm:monotonic} guarantees $\Tr\bigl(\rho_s^{(t+1)}Q_s^{(t)}\bigr)\ge\Tr\bigl(\rho_s^{(t)}Q_s^{(t)}\bigr)$ at every state. Since $Q_s^{(t)}$ is diagonal, behavior depends only on the diagonal (Section~\ref{sec:policies}). The classical policy-improvement theorem then applies and gives $V^{(t+1)}\ge V^{(t)}$.
\ENDFOR
\end{algorithmic}
\end{algorithm}

With a diagonal $Q_s$, both DMPG and DMPI leave a zero-probability action at zero, so every $\rho_{ii}$ should start positive. Remark~\ref{rem:facelowrank} (Appendix~\ref{app:remarks}) explains why. It also relates this limit to the face restriction of Appendix~\ref{sec:entropy}.

The third algorithm, ACPE, does not train a policy. It reduces the variance of on-policy evaluation for a fixed policy, as described in Appendix~\ref{sec:theory}. It is an evaluation method only, and we do not give a gradient version. The pseudocode handles the two cases of that section with a single \textbf{if} branch.

\begin{algorithm}[H]
\caption{Adaptive-Coherence Policy Evaluation (ACPE)}\label{alg:acpe}
\begin{algorithmic}[1]
\STATE \textbf{Input:} state $s$, current diagonal $\diag(\rho_s)$, pilot batch size $m$, evaluation batch size $N$, ridge schedule $\xi=\xi_m$ with $\xi_m\to0$ and $\xi_m\sqrt m\to\infty$.
\STATE Roll out $m$ pilot episodes. Form unbiased entry estimators $\{\hat o_{ii}\},\{\hat o_{ij}\}$ and compute the sample covariances $\hat{\boldsymbol{\gamma}},\hat\Gamma$. For diagonal $O$, Example~\ref{ex:constructionA} (Appendix~\ref{app:constructions}) gives these estimators. Drop any term with $\rho_{ii}=0$ from $\hat J_{\mathrm{diag}}$, since its weight is zero.
\STATE Set $\hat{\mathbf{c}}\leftarrow-\tfrac12(\hat\Gamma+\xi I)^{-1}\hat{\boldsymbol{\gamma}}$.
\IF{$O$ is non-diagonal}
  \STATE Scale $\hat{\mathbf{c}}\leftarrow\alpha\,\hat{\mathbf{c}}$ with the largest $\alpha\in[0,1]$ that keeps $\rho_{\alpha\hat{\mathbf{c}}}\succeq0$ (Corollary~\ref{cor:feasibility}). Such $\alpha>0$ exists whenever $\rho_{ii}>0$ for every index $i$ in a nonzero entry of $\hat{\mathbf{c}}$. Otherwise $\alpha=0$ and $\hat{\mathbf{c}}=0$, which recovers the diagonal estimator. The policy being evaluated is then the one with coherences $\hat{\mathbf{c}}$.
\ELSE
  \STATE Keep $\hat{\mathbf{c}}$ unscaled, as a control-variate term outside $\rho_s$ (Appendix~\ref{sec:theory}). No PSD constraint applies.
\ENDIF
\STATE Roll out $N$ evaluation episodes. In each episode, compute $\hat J(\hat{\mathbf{c}})=\sum_i(\rho_s)_{ii}\,\hat o_{ii}+2\sum_{i<j}\hat c_{ij}\hat o_{ij}$. Return the average over the $N$ episodes.
\STATE \textbf{Guarantee.} Given the pilot batch, the variance of one evaluation episode approaches the best value that the feasible coherences allow as $m\to\infty$ (Corollary~\ref{cor:adaptive}). For a fixed ridge, the corollary also bounds the extra variance that the ridge adds. The average over $N$ episodes has $1/N$ times the per-episode variance.
\end{algorithmic}
\end{algorithm}

The fourth algorithm is the procedure of Section~\ref{subsec:coupledreward}. It applies DMPG in the blended-action environment of Definition~\ref{def:structuredaction}, where the reward operator $O$ is unknown and non-diagonal. It shows that the environment is trainable end to end. It is not tuned for efficiency.

\begin{algorithm}[H]
\caption{DMPG in a blended-action environment (bandit case)}\label{alg:dmpgcoupled}
\begin{algorithmic}[1]
\STATE Initialize a real factor $U\in\R^{k\times r}$. Choose a ridge weight $\xi>0$ and an exploration probability $\zeta\in(0,1)$. Choose a step size $\alpha_{\mathrm{step}}>0$.
\FOR{iteration $t=1,\dots,T$}
  \STATE Set $h=\Tr(UU^\top)$ and $\rho=UU^\top/h$. Compute its eigenpairs $(\lambda_n,\mathbf{v}_n)$ from a thin SVD of $U$.
  \STATE With probability $1-\zeta$, draw $n$ with probability $\lambda_n$ and choose $\mathbf{x}^{(t)}=\mathbf{v}_n$. Otherwise choose a uniformly random unit vector.
  \STATE Observe the reward $R^{(t)}$, with $\E[R^{(t)}\mid \mathbf{x}^{(t)}]=\mathbf{x}^{(t)\top} O\mathbf{x}^{(t)}$.
  \STATE Refit $\hat O$ by ridge regression of $R^{(1)},\dots,R^{(t)}$ on the features $\mathbf{x}^{(t')}\mathbf{x}^{(t')\top}$, $t'\le t$, with penalty $\xi\|\hat O\|_F^2$.
  \STATE Set $J\leftarrow\Tr(\rho\hat O)$ and update $U\leftarrow U+\alpha_{\mathrm{step}}\cdot\tfrac{2}{h}(\hat O-JI)U$.
\ENDFOR
\end{algorithmic}
\end{algorithm}

\section{Explicit examples}\label{app:constructions}
This appendix has four items. The first is a worked example illustrating phase relevance (Section~\ref{subsec:phase}). The second is the construction of jointly distributed unbiased estimators $\hat o_{ij}$ used to instantiate Theorem~\ref{thm:optcoherence} in a standard MDP (Appendix~\ref{sec:theory}). The third is a numeric example showing a coherent reference policy in action (Section~\ref{subsec:coherentprior}). The fourth is an application to consolidating LLM reasoning paths, which illustrates the coupled reward (Section~\ref{subsec:coupledreward}).

\begin{example}[A worked $2\times 2$ case]\label{ex:toy}
Let $k=2$, $O=\begin{pmatrix}0&1\\1&0\end{pmatrix}$, and $\rho=\tfrac12\begin{pmatrix}1&\cos\phi\\ \cos\phi&1\end{pmatrix}$. Then $\Tr(\rho O)=\cos\phi$. Without coherence ($\phi=\frac{\uppi}{2}$), the uniform diagonal policy yields expected reward $0$. But maximal constructive coherence ($\phi=0$) yields $+1$ and maximal destructive coherence ($\phi=\uppi$) yields $-1$. Every classical policy earns $0$ here, because the diagonal of $O$ is zero. In particular, no classical policy moves the expectation away from $0$ while keeping both actions equally likely.

In the blended-action environment of Definition~\ref{def:structuredaction}, this case is realized by an agent that chooses $(1,1)/\sqrt2$ with probability $(1+\cos\phi)/2$ and $(1,-1)/\sqrt2$ otherwise. Each chosen vector puts weight $1/2$ on each action. The two vectors have mean rewards $+1$ and $-1$, so $\E[R]=\cos\phi$.
\end{example}

\begin{example}[Construction of estimators for standard MDPs]\label{ex:constructionA}
In a standard MDP, the observable is diagonal, so $O_{ij}=0$. Any rollout statistic with exactly zero mean qualifies as $\hat o_{ij}$. From each episode with realized action $a\sim\diag(\rho_s)$ and return $G$, take
\[
\hat o_{ii}=\mathbf 1\{a=a_i\}\,G/\rho_{ii},\qquad
\hat o_{ij}=\tfrac12\bigl(\mathbf 1\{a=a_i\}-\rho_{ii}\bigr)-\tfrac12\bigl(\mathbf 1\{a=a_j\}-\rho_{jj}\bigr),
\]
where $\mathbf 1\{\cdot\}$ denotes the indicator function, which equals $1$ if its argument is true and $0$ otherwise. Here $\hat o_{ii}$ needs $\rho_{ii}>0$. An action with $\rho_{ii}=0$ never occurs, and its term in $\hat J_{\mathrm{diag}}$ has weight zero, so we drop it. Both $\hat o_{ii}$ and $\hat o_{ij}$ are unbiased. Given $a=a_i$, the return $G$ has mean $Q(s,a_i)$, so $\hat o_{ii}$ has mean $Q(s,a_i)$. The statistic $\hat o_{ij}$ has mean exactly $O_{ij}=0$.

This construction also makes the $\hat o_{ij}$ redundant once $k$ grows. The reason is that $\hat o_{ij}=\tfrac12(\Omega_i-\Omega_j)$, where $\Omega_i=\mathbf 1\{a=a_i\}-\rho_{ii}$. The $k$ values $\Omega_1,\dots,\Omega_k$ always sum to zero. So they carry only $k-1$ free dimensions of information. The $k(k-1)/2$ pairwise differences built from them span at most $k-1$ dimensions too, no matter how many pairs there are. In fact the span has exactly $k-1$ dimensions when every $\rho_{ii}>0$. For $k=3$, $\hat o_{12}+\hat o_{23}-\hat o_{13}$ vanishes identically. Hence $\Gamma$ is singular for every $k\ge3$, since $k(k-1)/2>k-1$ there.
This is why Theorem~\ref{thm:optcoherence} is stated with the pseudoinverse $\Gamma^{+}$ rather than an inverse. In practice, one keeps only a maximal linearly independent subset of the $\hat o_{ij}$, so $\Gamma$ becomes invertible on the retained set. Alternatively, when $\Gamma$ and $\boldsymbol{\gamma}$ must be estimated from data rather than known exactly, one estimates $\mathbf{c}^\star$ with the ridge-regularized formula of Corollary~\ref{cor:adaptive}. This formula stays well defined even when $\Gamma$ is singular. It requires no subset selection at all.
\end{example}

\begin{example}[A coherent prior moves the trained policy]\label{ex:coherentprior}
Take $k=2$ actions with $Q_s=\diag(1,-0.3)$ and $\tau=0.7$. Let the reference policy have equal diagonal entries $0.5$ and a real coherence $c$, so
\[
\rho_s^{\mathrm{ref}}=\begin{pmatrix}0.5 & c\\ c& 0.5\end{pmatrix}, \qquad 0\le c<0.5\quad\text{(PSD requires }|c|\le0.5\text{)}.
\]
A classical method only ever sees $\diag(\rho_s^{\mathrm{ref}})=(0.5,0.5)$. Its regularized optimum is the standard prior-weighted softmax, $p_i\propto(\rho_s^{\mathrm{ref}})_{ii}\,e^{Q(s,a_i)/\tau}$, which gives $\mathbf{p}=(0.865,0.135)$ for every value of $c$. The coherence $c$ is simply invisible to it.

The DMRL optimum $\rho_\tau(\tilde Q_s)$ from Corollary~\ref{cor:coherentprior} depends on $c$ directly, through $\log\rho_s^{\mathrm{ref}}$. Computing it from the exact formula gives $\diag(\rho_\tau(\tilde Q_s))=(0.850,0.150)$ at $c=0.2$ and $(0.751,0.249)$ at $c=0.45$. The coherence of $\rho_\tau(\tilde Q_s)$ grows with $c$ as well, with $|(\rho_\tau(\tilde Q_s))_{12}|=0.160$ and $0.398$ at these two values.

At $c=0$, both methods agree exactly, matching Proposition~\ref{prop:coherencetransfer}. As $c$ grows, $\rho_\tau(\tilde Q_s)$ picks up a coherence of its own, and its action distribution moves steadily away from the classical answer. The classical method cannot follow, since it never had access to $c$ in the first place.

The size of this shift is worth noting. Proposition~\ref{prop:coherencetransfer} only says that $\rho_\tau(\tilde Q_s)$ is non-diagonal. It says nothing about how far the sampled action distribution actually moves. Here it moves by about eleven percentage points at $c=0.45$, from $0.865$ to $0.751$ on action~1. The shift is in the diagonal, so it changes which action gets played and how often. It is not confined to entries that sampling never uses.
\end{example}

\begin{example}[Consolidating LLM reasoning paths]\label{ex:llmpaths}
For a given prompt, an LLM produces $k$ reasoning paths. For a math problem, path~1 might be an algebraic derivation, path~2 a geometric argument, and path~3 a numerical check. An aggregator writes one final answer that gives path $i$ a weight $w_i\ge0$, with $\sum_iw_i=1$. A judge, such as a verifier model, then scores the answer.

We write $w_i=x_i^2$ for a unit vector $\mathbf{x}$ with non-negative entries, and we assume the judge's expected score is $\mathbf{x}^\top O\mathbf{x}$. The diagonal entry $O_{ii}$ is the score of path $i$ used on its own. The off-diagonal entry $O_{ij}$ measures how paths $i$ and $j$ interact when they are consolidated. It is positive when the two paths complement each other, for example when a numerical check makes an algebraic derivation more convincing. It is negative when they conflict. With the prompt as the state, this is the blended-action environment of Definition~\ref{def:structuredaction}.

A policy is trained once and then used on every prompt. For each prompt $s$, the policy is a distribution over unit vectors $\mathbf{x}$ with non-negative entries. By Proposition~\ref{prop:secondmoment}, its second-moment matrix $\rho_s$ is all that matters for the expected score. One training step runs as follows.
\begin{enumerate}
\item Take a training prompt $s$ and generate its $k$ paths.
\item Sample a unit vector $\mathbf{x}$ from the policy, which gives the weights $x_i^2$.
\item Aggregate the paths with these weights, and let the judge return a reward $R$.
\item Use the pair $(\mathbf{x},R)$ to update the policy parameters, for example with a gradient step based on Theorem~\ref{thm:gradient}.
\end{enumerate}
After training, the policy outputs weights for a new prompt directly, with no judge calls. What it learns are pairwise patterns that hold across prompts, such as a numerical check strengthening an algebraic derivation. These patterns are the coherences of $\rho_s$. Such patterns can carry over from one prompt to another only if the paths have consistent roles. For example, path~1 is always the algebraic derivation and path~3 is always the numerical check. Otherwise the pair $(1,3)$ means something different on every prompt, and there is no stable structure to learn.

A classical policy learns to pick one path per prompt, with some probabilities. This is a learned form of best-of-$k$ selection. The DMRL policy can also consolidate several paths into one answer.

Take two paths, an algebraic derivation A and a numerical check C, with
\[
O=\begin{pmatrix}0.6 & 0.3\\ 0.3 & 0.5\end{pmatrix}.
\]
Path A alone scores $0.6$, and path C alone scores $0.5$. The positive off-diagonal entry says that the two paths complement each other. Compare three strategies.
\begin{itemize}
\item \textit{Best single path.} Always using A scores $0.6$. No policy that uses one path per answer can do better.
\item \textit{Separate use.} Using A in $58.2\%$ of answers and C in the rest, one path per answer, scores $0.582\cdot0.6+0.418\cdot0.5\approx0.558$.
\item \textit{Consolidation.} Using weight $0.582$ on A and $0.418$ on C within every answer scores $\lambda_{\max}(O)\approx0.854$. The extra $2\cdot0.3\cdot\sqrt{0.582\cdot0.418}\approx0.296$ is the bonus from the positive interaction between the two paths.
\end{itemize}
The consolidation and the separate strategy use the two paths with the same frequencies, yet the consolidation scores more, as Proposition~\ref{prop:coherentdominance} predicts. The consolidation also beats the best single path, as Proposition~\ref{prop:gap} predicts, since $\lambda_{\max}(O)\approx0.854$ exceeds $\max_iO_{ii}=0.6$. The weights come from the top eigenvector of $O$, which is approximately $(0.763,0.646)$ and squares to $(0.582,0.418)$. This eigenvector has non-negative entries because the off-diagonal entry of $O$ is non-negative. So the weights are feasible, and the consolidation reaches $\lambda_{\max}(O)$.

The sign restriction affects which of the general results carry over. Because $w_i=x_i^2$, the aggregator cannot use the signs of $x_i$, so the policy may choose only non-negative vectors. Every coherence $\rho_{ij}=\E[x_ix_j]$ is then non-negative. A negative off-diagonal entry of $O$ cannot be exploited. So Proposition~\ref{prop:coherentdominance} holds here when some $O_{ij}>0$. Adding a small positive coherence between such a pair raises the score, which gives the strict gain. Proposition~\ref{prop:gap} holds as stated when every off-diagonal entry of $O$ is non-negative. Then the Perron--Frobenius theorem gives a non-negative top eigenvector, so $\lambda_{\max}(O)$ is still reached. Both conditions hold for the matrix above. The training procedure must also keep the vectors non-negative. A gradient step on $U$ can produce eigenvectors with mixed signs, and the aggregator cannot use those as weights.
\end{example}

\section{Deferred remarks}\label{app:remarks}
This appendix contains remarks deferred from the main text to keep it focused on the core results. The first concerns the gradient convention of Section~\ref{sec:foundations}. The next two concern the trust-region face restriction of Appendix~\ref{sec:entropy}. The next five concern the optimal coherence of Appendix~\ref{sec:theory}. The last two concern the combination of guarantees across states and the diagnostics mentioned in Section~\ref{sec:limitations}.

\begin{remark}[The Wirtinger gradient]\label{rem:wirtinger}
Theorem~\ref{thm:gradient} states the gradient with respect to $\overline{U}$. This is the Wirtinger convention. It treats $U$ and $\overline{U}$ as independent variables, so $J$ can be differentiated with respect to $\overline{U}$ while $U$ is held fixed. Write $U=X+\mathrm{i}Y$ with $X$ and $Y$ real. Then $\nabla_{\overline{U}}J=\tfrac12(\nabla_XJ+\mathrm{i}\nabla_YJ)$.

This gives $\nabla_XJ+\mathrm{i}\nabla_YJ=2\nabla_{\overline{U}}J$. Since $J$ is real valued, this is the direction of steepest ascent of $J$ under the real inner product $\re\Tr(V^\dagger W)$. So a gradient step on a complex $U$ moves along $2\nabla_{\overline{U}}J=\tfrac{2}{h}(O-J(U)I)U$, where $h=\Tr(UU^\dagger)$ as in the theorem. This has the same form as the real gradient in the theorem.

Readers who prefer to avoid complex calculus can restrict $U$ to be real and $O$ to be real symmetric, and use the real gradient directly. If $O$ is complex Hermitian and $U$ is real, then $\Tr(\rho(U)O)$ involves only the real part of $O$, and the same formula holds with $\re O$ in place of $O$.
\end{remark}

\begin{remark}[The support restriction is necessary]\label{rem:facenecessary}
For rank-deficient $\rho_s^{(t)}$, the restriction to $\mathcal F(\rho_s^{(t)})$ cannot be dropped. Take $k=2$, $\rho_s^{(t)}=\eb_1\eb_1^\dagger$ (a policy always picking action~1), and $A_s^{(t)}=\diag(0,1)$ (rewarding action~2 only). Note that $\rho_s^{(t)}$ is diagonal in this example, so this is an ordinary classical policy. The global maximum of $\Tr(\rho A_s^{(t)})$ over $\mathcal D_2$ is $1>0=\Tr(\rho_s^{(t)}A_s^{(t)})$, attained at $\eb_2\eb_2^\dagger$. The policy always picking action~2 is clearly reachable in principle, and strictly better. Yet $D(\rho\,\Vert\,\rho_s^{(t)})<\infty$ forces $\supp\rho\subseteq\supp\rho_s^{(t)}=\mathrm{span}\{\eb_1\}$. The reason is that $\rho_s^{(t)}$ assigns action~2 exactly zero probability. Any move toward giving it positive probability costs infinite relative entropy, no matter how good action~2 is. Since $\rho$ is PSD with $\Tr\rho=1$, the only matrix supported on this one-dimensional space is the projector onto it. So $\rho=\eb_1\eb_1^\dagger=\rho_s^{(t)}$. The face $\mathcal F(\rho_s^{(t)})$ is then the single point $\rho_s^{(t)}$. For every $\delta>0$, the trust region literally admits no feasible point other than $\rho_s^{(t)}$, and no improvement is possible. In general, the strict inequality can only be claimed relative to $\rho_s^{(t)}$'s support face, as Proposition~\ref{prop:surrogate} states.
\end{remark}

\begin{remark}[Trust-region face restriction under low rank]\label{rem:facelowrank}
Proposition~\ref{prop:surrogate} showed that the face restriction $\mathcal F(\rho_s^{(t)})$ never limits the optimum when $\rho_s^{(t)}$ is full rank. That covers Gibbs states and iterates of the matrix softmax map started from a full-rank state. As defined in Algorithm~\ref{alg:dmpi}, DMPI operates directly on $\rho_s$. It is initialized at a full-rank state, such as a strictly positive diagonal or $I/k$. Since the matrix softmax map preserves full rank, every DMPI iterate stays full rank, and Proposition~\ref{prop:surrogate}'s guarantee applies at its full strength throughout. Under the factor parameterization of Definition~\ref{def:factorparam}, $\rho_s^{(t)}=\rho(U)$ has rank at most $r$. Whenever $r<k$, $\rho_s^{(t)}$ is not full rank. Then $\mathcal F(\rho_s^{(t)})$ is a proper sub-face of $\mathcal D_k$. Any update constrained to solve~\eqref{eq:trustregion} on $\rho_s^{(t)}$ would only be guaranteed to improve relative to that face, not against maximizers of $\Tr(\rho A_s^{(t)})$ outside $\rho_s^{(t)}$'s support. DMPG avoids this face restriction, as it does not solve~\eqref{eq:trustregion} at all. It takes unconstrained gradient steps on $U$, using the exact gradient of Theorem~\ref{thm:gradient}, which holds at any rank. The range of $U$ (the subspace its columns span) determines which directions $\rho_s^{(t)}$ currently has weight on. Nothing in the gradient update fixes this subspace in place. Successive steps can shift it toward different directions in $\C^k$. So while each $\rho_s^{(t)}$ always has rank at most $r$, the particular $r$-dimensional subspace it occupies can change over the course of training. The caveat above is therefore not a concern for DMPI when it is initialized at full rank, as suggested. It applies to any algorithm that solves a trust-region problem on a low-rank iterate.

DMPI and DMPG do not solve a trust-region problem, but they share a related limit for a diagonal $Q_s$. Before normalization, the improvement map multiplies each $\rho_{ii}$ by $e^{\beta q_i}$. Consider DMPG with a tabular $U$ that is updated directly. A step without the entropy term then rescales row $i$ of $U$ by a scalar. So an action with zero probability stays at zero in both algorithms. In DMPG, the column space of $U$ can still rotate, but it stays inside the coordinate subspace of the actions with positive probability. With a network $U_\theta(s)$, a step in $\theta$ need not keep a zero row at zero exactly. The statement above is therefore for the tabular case. Starting with every $\rho_{ii}>0$ avoids this limit.

This is not a new pathology introduced by the density-matrix framework. The counterexample of Remark~\ref{rem:facenecessary} is a diagonal, deterministic policy, so it is an ordinary classical policy. The same restriction holds for any classical trust region of the form $D_{\mathrm{KL}}(\pi(\cdot\mid s)\Vert\pi^{(t)}(\cdot\mid s))\le\delta$. This KL is infinite when $\pi(\cdot\mid s)$ puts mass on an action with $\pi^{(t)}(a_i\mid s)=0$. TRPO as usually stated constrains the other direction, $D_{\mathrm{KL}}(\pi^{(t)}(\cdot\mid s)\Vert\pi(\cdot\mid s))$. Even so, importance-ratio methods such as PPO and sampled TRPO surrogates share a blind spot, because an action with $\pi^{(t)}(a_i\mid s)=0$ is never sampled. A rank-deficient $\rho_s^{(t)}$ plays the same role as a zero-probability action in the classical case. It marks a direction the trust region can never reach.
\end{remark}

\begin{remark}[The kernel of $\Gamma$]\label{rem:kernel}
Let $\hat{\mathbf{o}}=(\hat o_{ij})_{i<j}$ be the vector of off-diagonal statistics. Let $\mathbf{o}_{\mathrm{off}}=(O_{ij})_{i<j}$ be the vector of off-diagonal entries of $O$, ordered like $\mathbf{c}$. Take any $\mathbf{z}\in\ker\Gamma$. Then $\Var(\mathbf{z}^\top\hat{\mathbf{o}})=\mathbf{z}^\top\Gamma\mathbf{z}=0$, so $\mathbf{z}^\top\hat{\mathbf{o}}$ is almost surely constant. Its value is its mean $\mathbf{z}^\top\mathbf{o}_{\mathrm{off}}$.

This has two consequences. First, moving $\mathbf{c}$ by $\mathbf{z}$ shifts $\hat J(\mathbf{c})$ by the fixed constant $2\mathbf{z}^\top\mathbf{o}_{\mathrm{off}}$. A constant shift does not change the variance, so $\Var(\hat J(\mathbf{c}))$ is unchanged. Second, a constant statistic is uncorrelated with $\hat J_{\mathrm{diag}}$, so $\mathbf{z}^\top\boldsymbol{\gamma}=0$. Hence $\boldsymbol{\gamma}$ is orthogonal to $\ker\Gamma$ and lies in $\mathrm{range}(\Gamma)$. So $\Gamma\Gamma^{+}\boldsymbol{\gamma}=\boldsymbol{\gamma}$, which means that $\mathbf{c}^\star=-\tfrac12\Gamma^{+}\boldsymbol{\gamma}$ satisfies the first-order condition $\Gamma\mathbf{c}=-\tfrac12\boldsymbol{\gamma}$. This is why $\mathbf{c}^\star$ is a minimizer, and why the minimizers form the set $\mathbf{c}^\star+\ker\Gamma$.

When $O$ is diagonal, $\mathbf{o}_{\mathrm{off}}=0$, so the shift is zero. When $O$ is non-diagonal, the return $J(\rho_{\mathbf{c}})$ changes by the same amount $2\mathbf{z}^\top\mathbf{o}_{\mathrm{off}}$ while the variance does not. Remark~\ref{rem:pareto} uses this fact.
\end{remark}

\begin{remark}[Computing the feasibility scaling]\label{rem:alphamax}
The formula for $\alpha_{\max}$ is derived in the proof of Corollary~\ref{cor:feasibility} in Appendix~\ref{app:proofs-variance}. The derivation applies to any coherence vector $\mathbf{c}$. Let $C$ be the hollow symmetric matrix built from $\mathbf{c}$, and suppose $\rho_{ii}>0$ for every $i$. Then the largest scaling $\alpha_{\max}$ that keeps $\rho_{\mathrm{diag}}+\alpha C$ PSD is
\[
\alpha_{\max}=1/\max\bigl[\,0,\lambda_{\max}(-\rho_{\mathrm{diag}}^{-1/2}C\rho_{\mathrm{diag}}^{-1/2})\,\bigr],
\]
where $\alpha_{\max}=\infty$ when the maximum is $0$. Only this one extreme eigenvalue is needed. Since $\rho_{\mathrm{diag}}$ is diagonal, multiplying the matrix $\rho_{\mathrm{diag}}^{-1/2}C\rho_{\mathrm{diag}}^{-1/2}$ by a vector costs $\mathcal O(k^2)$. A Lanczos iteration, or a power iteration with a spectral shift, therefore finds the extreme eigenvalue at $\mathcal O(k^2)$ per iteration. This avoids the $\mathcal O(k^3)$ cost of a full eigendecomposition. The cost is small next to forming $\Gamma^{+}$, which is needed to obtain $\hat{\mathbf{c}}$ in the first place. The procedure of Corollary~\ref{cor:adaptive} pays the cost of $\Gamma^{+}$ whether or not the scaling is needed.
\end{remark}

\begin{remark}[The constraint under any parameterization]\label{rem:param}
The PSD constraint on $\rho_{\mathbf{c}}$ does not come from building $\rho$ entry by entry. It appears under any parameterization that targets a specific coherence value. Suppose we want $\rho$ to realize a chosen pair $(\rho_{\mathrm{diag}},\hat{\mathbf{c}})$. This is possible only if every $2\times2$ principal submatrix of the target is PSD, because every density matrix is PSD. For each pair $i,j$, this means $|\hat c_{ij}|^2\le\rho_{ii}\rho_{jj}$. It is the same condition that bounds $\alpha_{\max}$ after Corollary~\ref{cor:feasibility}, and it does not depend on how $\rho$ is parameterized. So the constraint tracks what we ask $\rho$ to represent, not how we build it.

Two methods avoid the constraint because they never target an explicit off-diagonal value. Under the factor parameterization $\rho(U)=UU^\dagger/\Tr(UU^\dagger)$ of Definition~\ref{def:factorparam}, $\rho(U)$ is PSD for every $U$. So gradient steps taken directly on $U$, as in DMPG, never leave the PSD set. DMPI (Algorithm~\ref{alg:dmpi}) takes a different route. It alternates evaluation with the improvement map of Definition~\ref{def:softmaxchannel}. This map is PSD by construction, and Theorem~\ref{thm:monotonic} guarantees per-state improvement for it. Neither method sets a coherence explicitly.
\end{remark}

\begin{remark}[The return--variance trade-off]\label{rem:pareto}
As in Appendix~\ref{sec:theory}, take $O$ and the coherences to be real. Let $J_{\mathrm{diag}}=\sum_ip_iO_{ii}$ be the return of the diagonal policy. Let $\mathbf{o}_{\mathrm{off}}=(O_{ij})_{i<j}$ collect the off-diagonal entries of $O$, ordered like $\mathbf{c}$. Then $J(\rho_{\mathbf{c}})=J_{\mathrm{diag}}+2\mathbf{c}^\top\mathbf{o}_{\mathrm{off}}$ is linear in $\mathbf{c}$. By Theorem~\ref{thm:optcoherence}, $\Var(\hat J(\mathbf{c}))$ is an exact quadratic in $\mathbf{c}$. For a penalty weight $\nu>0$, consider
\[
\max_{\mathbf{c}\,:\,\rho_{\mathbf{c}}\succeq0}\;J(\rho_{\mathbf{c}})-\nu\Var\bigl(\hat J(\mathbf{c})\bigr).
\]
Since $\Gamma\succeq0$, the objective is a concave quadratic. The feasible set is convex, so this is a convex program. Sweeping $\nu$ traces out the Pareto frontier between return and estimator variance.

Now drop the PSD constraint and suppose $\Gamma\succ0$. Setting the gradient of the objective to zero gives
\[
\mathbf{c}(\nu)=\mathbf{c}^\star+\frac{1}{4\nu}\,\Gamma^{-1}\mathbf{o}_{\mathrm{off}}.
\]
As $\nu\to\infty$, this approaches the variance-optimal $\mathbf{c}^\star$. As $\nu$ shrinks, the return term takes over. If $\rho_{\mathbf{c}(\nu)}$ is PSD, then $\mathbf{c}(\nu)$ also solves the constrained program. Otherwise the constrained program is a convex quadratic program with a semidefinite constraint. It has no closed form in general, although standard solvers handle it. The same holds for the limit $\nu\to\infty$, which minimizes the variance alone over the feasible set. This is the problem left open in Appendix~\ref{sec:theory} when $\alpha_{\max}<1$.

If $\Gamma$ is singular, the return still changes along $\ker\Gamma$ while the variance does not (Remark~\ref{rem:kernel}). Without the PSD constraint, the program is then unbounded unless $\mathbf{o}_{\mathrm{off}}$ is orthogonal to $\ker\Gamma$. The PSD constraint is what keeps it bounded in that case.

In every case, the program reuses the $\boldsymbol{\gamma}$ and $\Gamma$ already needed for $\mathbf{c}^\star$, and it adds only $\mathbf{o}_{\mathrm{off}}$. In practice all three are pilot-batch estimates. So the program is refit once per pilot batch, not on every rollout.
\end{remark}

\begin{remark}[The adaptive estimator in detail]\label{rem:adaptive}
This remark supports the discussion after Corollary~\ref{cor:adaptive}.

\emph{Conditional and total variance.} Equation~\eqref{eq:condvar} is conditional on $\tilde{\mathbf{c}}$, so it holds for any $O$ without taking an expectation over pilot batches. The total variance $\Var(\hat J(\tilde{\mathbf{c}}))$ is unconditional. By the law of total variance,
\[
\Var\bigl(\hat J(\tilde{\mathbf{c}})\bigr)=\E\bigl[\Var(\hat J(\tilde{\mathbf{c}})\mid\tilde{\mathbf{c}})\bigr]+\Var\bigl(\E[\hat J(\tilde{\mathbf{c}})\mid\tilde{\mathbf{c}}]\bigr).
\]
The first term is the expectation of \eqref{eq:condvar}. The second term is the variance, across pilot batches, of the conditional mean $\E[\hat J(\tilde{\mathbf{c}})\mid\tilde{\mathbf{c}}]=J(\rho_{\tilde{\mathbf{c}}})$.

When $O$ is diagonal, $J(\rho_{\mathbf{c}})$ does not depend on $\mathbf{c}$. So $J(\rho_{\tilde{\mathbf{c}}})$ is the fixed constant $J$, and the second term vanishes. Taking the expectation of \eqref{eq:condvar} then gives
\[
\Var\bigl(\hat J(\tilde{\mathbf{c}})\bigr)
=\Var\bigl(\hat J_{\mathrm{diag}}\bigr)-\boldsymbol{\gamma}^\top\Gamma^{+}\boldsymbol{\gamma}
+4\,\E\bigl[(\tilde{\mathbf{c}}-\mathbf{c}^\star)^\top\Gamma(\tilde{\mathbf{c}}-\mathbf{c}^\star)\bigr].
\]
When $O$ is non-diagonal, the second term generally does not vanish. The total variance is then at least the expectation of \eqref{eq:condvar}. It is strictly larger when $J(\rho_{\tilde{\mathbf{c}}})$ varies across pilot batches.

\emph{The dimension of $\Gamma$.} The matrix $\Gamma$ has size $L_\Gamma\times L_\Gamma$, where $L_\Gamma=k(k-1)/2$ is the number of pairs $i<j$. It is smaller if only a linearly independent subset of the statistics is kept. Under Example~\ref{ex:constructionA}, its effective rank is only $k-1$. So the pilot batch only needs to pin down this low-dimensional object, not the full problem.
\end{remark}

\begin{remark}[Combining guarantees across states]\label{rem:acrossstates}
Some results are per-state identities, so they need no combining. The Gibbs optimum (Theorem~\ref{thm:gibbs}), the gradient formula (Theorem~\ref{thm:gradient}), and the variance decompositions and optimal-coherence results (Theorem~\ref{thm:optcoherence}) hold exactly at each state. This is true even when all states share one parameterization $\rho_s=\rho_s(\theta)$.

The trust-region transfer does need combining, and it works. The pinching bound of Theorem~\ref{thm:pinching} holds at every state. Averaging both sides over any distribution of states then bounds the expected classical KL by the expected matrix relative entropy.

Monotonic improvement (Theorem~\ref{thm:monotonic}) also needs combining, but the guarantee does not carry over. With shared parameters, projecting the ideal per-state update onto the model class can degrade some states while improving others. This is what happens in classical deep RL, and DMRL does not remove it.

Across training time, the gain factors $g_{\mathrm{EA}}$ and $g_{\mathrm{CV}}$ are evaluated at the current policy. Training is nonstationary, so both drift as training proceeds. The theory therefore does not show that these per-state gains compound into fewer environment interactions to reach a fixed performance level.
\end{remark}

\begin{remark}[Diagnostics for coherences]\label{rem:diagnostics}
This remark supports Section~\ref{sec:limitations}. Off-diagonal entries are harder to read than action probabilities. It therefore helps to monitor them directly during training. Four diagnostics are useful.

First, plot the phases of the off-diagonal entries over training. Second, analyze the eigenvectors of $\rho$ to reveal latent clusters of related actions. Third, track the entropy and the coherence norms. Fourth, track the purity $\Tr\rho^2$ (Definition~\ref{def:purity}).

Purity is especially useful. Proposition~\ref{prop:purityrank} shows that it lower-bounds the rank needed for a target fidelity. So a low purity certifies that a small rank $r$ cannot reach that fidelity.
\end{remark}

\end{document}